\documentclass[10pt,twocolumn,letterpaper]{article}

\usepackage[pagenumbers]{cvpr} 

\usepackage[accsupp]{axessibility}  

\usepackage{multirow}
\usepackage{tablefootnote}

\usepackage{pifont}

\newcommand{\greenyes}{\textcolor{green}{\ding{51}}}
\newcommand{\redno}{\textcolor{red}{\ding{55}}}

\definecolor{cvprblue}{rgb}{0.21,0.49,0.74}
\usepackage[pagebackref,breaklinks,colorlinks,allcolors=cvprblue]{hyperref}

\def\paperID{11680} 
\def\confName{CVPR}
\def\confYear{2025}

\title{
GME: Improving Universal Multimodal Retrieval by Multimodal LLMs
}

\author{
Xin Zhang\textsuperscript{1}\thanks{
Equal Contribution.
Work done during the internship of XZ, WX and ZD.
DL is the tech lead.
\textsuperscript{\dag}Correspondence: mason.zms@gmail.com
},
Yanzhao Zhang\textsuperscript{2}\footnotemark[1], Wen Xie\textsuperscript{2}\footnotemark[1],
Mingxin Li\textsuperscript{2}, Ziqi Dai\textsuperscript{2}, Dingkun Long\textsuperscript{2} \\
Pengjun Xie\textsuperscript{2}, Meishan Zhang\footnotemark[2], Wenjie Li\textsuperscript{1}, Min Zhang\textsuperscript{3} \\
\textsuperscript{1}The Hong Kong Polytechnic University
\textsuperscript{2}Tongyi Lab, Alibaba Group
\textsuperscript{3}Soochow University \\
{\small \url{https://hf.co/Alibaba-NLP/gme-Qwen2-VL-2B-Instruct}} \\
{\small\tt 
\{linzhang.zx,zhangyanzhao.zyz,dingkun.ldk\}@alibaba-inc.com}
}

\begin{document}
\maketitle

\begin{abstract}
Universal Multimodal Retrieval (UMR) aims to enable search across various modalities using a unified model, where queries and candidates can consist of pure text, images, or a combination of both.
Previous work has attempted to adopt multimodal large language models (MLLMs) to realize UMR using only text data.
However, our preliminary experiments demonstrate that more diverse multimodal training data can further unlock the potential of MLLMs.
Despite its effectiveness, the existing multimodal training data is highly imbalanced in terms of modality, which motivates us to develop a training data synthesis pipeline and construct a large-scale, high-quality fused-modal training dataset.
Based on the synthetic training data, we develop the General Multimodal Embedder (GME), an MLLM-based dense retriever designed for UMR.
Furthermore, we construct a comprehensive UMR Benchmark (UMRB) to evaluate the effectiveness of our approach.
Experimental results show that our method achieves state-of-the-art performance among existing UMR methods. 
Last, we provide in-depth analyses of model scaling and training strategies, and perform ablation studies on both the model and synthetic data.
\end{abstract}

\section{Introduction}

The growth of multimedia applications necessitates retrieval models that extend beyond traditional text-to-text and text-to-image search \cite{zhou-etal-2024-marvel}.
In Universal Multimodal Retrieval (UMR) tasks, both queries and candidates can exist in any modality \cite{liu2023universal}. Compared to addressing this challenge with separate uni-modal and cross-modal retrievers in a divide-and-conquer pipeline \cite{DBLP:conf/cvpr/ChangCNGSB22}, a unified retriever is a more viable option in terms of usability and scalability.
Using the dense retrieval paradigm (also known as embedding-based retrieval) \cite{karpukhin-etal-2020-dense}, a unified model can be trained to project inputs from various modalities into a shared embedding space \cite{zhou-etal-2024-marvel,zhou-etal-2024-vista,jiang2024e5v}. In this space, similarity scores are computed between the embeddings of queries and the retrieval collection, facilitating the efficient ranking of the top-$k$ candidates.
To achieve this, some previous studies have primarily focused on two approaches: (1) designing feature fusion mechanisms for cross-modal retrievers based on the CLIP architecture \cite{liu2023universal,wei2023uniir}, and (2) incorporating visual plugin modules into optimized text embedding models to achieve unified multimodal representations \cite{zhou-etal-2024-marvel,zhou-etal-2024-vista}.

\begin{figure}[t]
\centering
\includegraphics[width=\columnwidth]{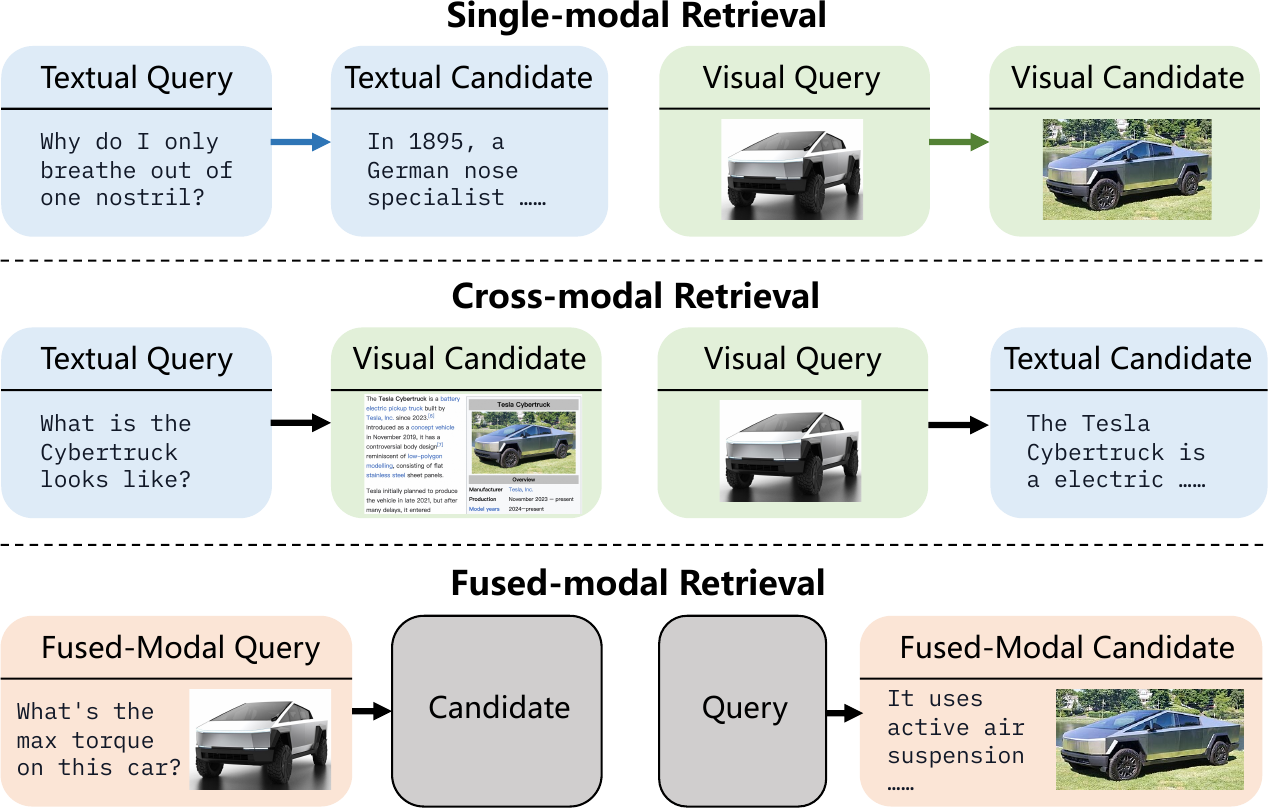}
\caption{
Illustration of different retrieval settings in our universal multimodal retrieval task.
Blocks with black borders represent data in arbitrary modalities, \ie text-only, image-only or fused.
}
\label{fig:umr-setting}
\end{figure}

Recently, researchers have turned to exploring Multimodal Large Language Models (MLLMs) \cite{DBLP:conf/nips/LiuLWL23a,wang2024qwen2vl} in UMR. For example, it is shown that training MLLMs with text data alone can generate universal multimodal embeddings with respectable retrieval performance \cite{jiang2024e5v}.
However, modality-limited training may fail to fully demonstrate the potential of MLLMs in UMR.
We believe that incorporating multimodal data composition (as shown in Figure \ref{fig:umr-setting}) could further enhance the model performance and generalization.
Moreover, visual documents (\ie document screenshots) are increasingly important in UMR tasks, as they not only simplify the pipelines of diverse Retrieval-Augmented Generation (RAG) applications, but also mitigate information loss during modality conversion \cite{ma-etal-2024-unifying,faysse2024colpali}.
However, current UMR models primarily target natural images, neglecting support for this scenario (Table \ref{tab:comparison}).

\begin{table}
\centering
\resizebox{\columnwidth}{!}{
\setlength{\tabcolsep}{2pt}
\begin{tabular}{@{}lccccc@{}}
\toprule
\multirow{2}{*}{\bf Methods} & \multicolumn{2}{c}{\bf Modeling} & \multicolumn{3}{c}{\bf Retrieval Setting}  \\ \cmidrule(lr){2-3} \cmidrule(lr){4-6}
& Approach & Training & S\&C & Fused & VD \\ \midrule
UniVL-DR \cite{liu2023universal} & CLIP Feat. Fusion  & Cross-modal & \greenyes & \redno & \redno\\
\multirow{2}{*}{UniIR \cite{wei2023uniir}} & CLIP Score Fusion  & \multirow{2}{*}{Multimodal} & \multirow{2}{*}{\greenyes} & \multirow{2}{*}{\greenyes} & \multirow{2}{*}{\redno} \\
 & BLIP Feat. Fusion \\
MARVEL \cite{zhou-etal-2024-marvel} & Text Enc.+Plugin & Cross-modal & \greenyes & \redno & \redno \\
VISTA \cite{zhou-etal-2024-vista} & Text Enc.+Plugin & Multimodal & \greenyes & \greenyes & \redno \\
E5-V \cite{jiang2024e5v} & MLLM & Text-only & \greenyes & \greenyes & \redno \\
\midrule
\textbf{GME} (Ours)  & MLLM & Multimodal & \greenyes &\greenyes & \greenyes \\
\bottomrule
\end{tabular}
}
\caption{
Comparison of UMR studies.
\texttt{Feat.} and \texttt{Enc.} are abbreviations for ``Feature'' and ``Encoder''.
\texttt{S\&C}, \texttt{Fused}, and \texttt{VD} denote the retrieval setting of single-modal \& cross-modal, fused-modal, and retrieving visual documents (\eg PDF screenshots), respectively.
The setting explaination is in Figure \ref{fig:umr-setting}.
}
\label{tab:comparison}
\end{table}

To address the aforementioned challenges, we propose the \underline{G}eneral \underline{M}ultimodal \underline{E}mbedder (\textbf{GME}), an instruction-based embedding framework utilizing MLLMs as the backbone. GME enables retrieval across various modalities in the unified paradigm, including text, images, visual documents, and fused-modal\footnote{
We use fuse-modal instead of multimodal to denote the data that contains both text and image to disambiguate from the UMR task.
} (\ie image-text composed) contents. Our framework is underpinned by two key techniques:
(1) A strategically optimized training data composition for UMR. We categorize UMR tasks into three types: single-modal, cross-modal, and fused-modal (Figure \ref{fig:umr-setting}). Through extensive experimentation, we analyze how different compositions affect performance (Figure \ref{fig:data_mix}) and demonstrate that a balanced mixture of all types yields optimal results.
(2) An efficient fused-modal data synthesis pipeline. Recognizing the under-representation of fused-modal data and its potential impact on training effectiveness, we develop a streamlined data synthesis pipeline (\S\ref{sec:data-gen}). This approach has successfully generated a comprehensive dataset of 1.1M fused-modal pairs, significantly enhancing our training and model capabilities.

To evaluate the effectiveness of our framework, we compile a comprehensive \underline{UMR} \underline{B}enchmark, namely \textbf{UMRB}.
This benchmark encompasses tasks from widely recognized retrieval benchmarks in text \cite{thakur2beir}, multimodal \cite{wei2023uniir}, and visual document retrieval \cite{faysse2024colpali}, as well as our newly processed fused-modal retrieval data.
We build our models on top of the strong Qwen2-VL series MLLMs \cite{wang2024qwen2vl} and train them on our constructed dataset.
Experimental results demonstrate that our model achieves state-of-the-art performance on UMRB.
Additionally, we perform in-depth analyses on model scaling, training strategies, and ablation of our synthetic data.
Our key contributions are:
\begin{itemize}
\item We explore strategies to adapt MLLMs into UMR models, and present \texttt{GME}, a powerful embedding model capable of retrieving candidates across different modalities. \texttt{GME} is the first UMR model to deliver visual document retrieval performance on par with specialized models.
\item We propose a novel data synthesis pipeline for constructing large-scale, fused-modal training data to encounter the scarcity of such training data. This pipeline is more efficient than previous approaches and can be easily extended to other domains.
\item We compile the UMR benchmark, \texttt{UMRB}, to evaluate a broader range of retrieval tasks compared to existing benchmarks. \texttt{UMRB} categorizes tasks into three types: single-modal, cross-modal, and fused-modal, and offers a comprehensive  performance evaluation across them.
\end{itemize}

\section{Related Work}
\paragraph{Multimodal Large Language Models} The emergence of Large Language Models (LLMs) has driven significant progress in natural language processing \cite{gpt3,openai2024gpt4technicalreport}, leading to the development of Multimodal LLMs that extend these capabilities to handle multimodal information. Prominent MLLMs such as GPT-4V \cite{2023GPT4VisionSC}, LLaVa~\cite{DBLP:conf/nips/LiuLWL23a,liu2024llavanext}, Qwen-VL~\cite{wang2024qwen2vl}, InternVL~\cite{chen2024far} and MiniCPM-V~\cite{yao2024minicpm} have shown promising advancements in multimodal information understanding and reasoning. Typically, an MLLM consists of an LLM, a vision encoder, and a projector that bridges the two components by transforming raw multimodal inputs into vectors compatible with the LLM~\cite{yin2024surveymultimodallargelanguage}.

\paragraph{Multimodal Retrieval} Early multimodal retrieval tasks focused on single-modal~\cite{Zhao2022DenseTR} or cross-modal retrieval~\cite{wang2016comprehensivesurveycrossmodalretrieval}. Recently, the expansion of multimedia applications and multimodal retrieval-augmented generation (RAG) by MLLMs has created a need for unified multimodal retrieval models for complex scenarios. Existing approaches largely utilize pre-trained models such as CLIP~\cite{DBLP:conf/icml/RadfordKHRGASAM21} or BLIP~\cite{DBLP:conf/icml/0001LXH22} for multimodal embedding. For instance, UniVL-DR~\cite{liu2023universal} and UniIR~\cite{wei2023uniir} initially encode images and texts separately using CLIP or BLIP encoders, followed by fusion strategies like score fusion to integrate features from both modalities. Additionally, VISTA~\cite{zhou-etal-2024-vista} and MARVEL~\cite{zhou-etal-2024-marvel} employ pre-trained text embedding models enhanced with visual plugins to encode composite image-text candidates. However, these methods are typically designed for specific tasks like multimodal document retrieval and lack flexibility to handle diverse multimodal retrieval tasks.

Concurrent with our work, E5-V~\cite{jiang2024e5v} and VLM2VEC~\cite{jiang2024vlm2vec} propose fine-tuning MLLMs on single-text (NLI~\cite{gao-etal-2021-simcse}) or vision-centric relevance data, demonstrating their transferability to multimodal retrieval. In this paper, we are the first to explore the fine-tuning of an MLLM-based universal multimodal retriever that can address both visual retrieval tasks and maintain strong text-to-text retrieval capabilities. Moreover, we are the first to extend a unified retrieval model to handle not only natural image retrieval but also text-rich image retrieval~\cite{faysse2024colpali}.

\paragraph{Embedding Models with Pre-trained Language Models} With the advancement of pre-trained Language Models, research in both pure text and Vision-Language Models has focused on building representation models based on these pre-trained language models. In the text retrieval domain, state-of-the-art text embedding models such as Contriver~\cite{izacard2022unsupervised}, E5~\cite{DBLP:journals/corr/abs-2212-03533}, GTE~\cite{li2023generaltextembeddingsmultistage}, and BGE~\cite{DBLP:conf/sigir/XiaoLZMLN24} are all built upon pre-trained language models and have demonstrated impressive generalization and robust performance in text retrieval tasks. Recently, leveraging LLMs combined with supervised fine-tuning (SFT), researchers have developed unified text representation models that fully utilize the text understanding capabilities of LLMs, resulting in models with enhanced performance and generalization~\cite{DBLP:conf/acl/WangYHYMW24,li2023generaltextembeddingsmultistage,lee2025nvembed}. These models typically process user text inputs through LLMs, using the hidden states from the final transformer layer—either through pooling or by selecting the last token—as the final representation. Inspired by the success of universal text embedding models based on text LLMs, researchers have begun to explore the construction of unified multimodal retrieval models using MLLMs~\cite{jiang2024e5v,jiang2024vlm2vec}. In this paper, we aim to demonstrate through systematic experiments that constructing a truly universal multimodal retrieval model using MLLMs is feasible.

\section{Universal Multimodal Retrieval}
Current UMR sub-tasks can be categorized into three types based on the modalities of the query and the candidate:
\begin{itemize}
\item \textbf{Single-Modal Retrieval}: Both the query and the candidate belong to the same modality, such as text-to-text (T$\rightarrow$T) or image-to-image (I$\rightarrow$I) retrieval scenarios. 
\item \textbf{Cross-Modal Retrieval}: The query and the candidate belong to different modalities, typically text-to-image (T$\rightarrow$I) retrieval. Unlike most prior work that focuses on natural-style image retrieval, we also consider the retrieval of rich-text images (e.g., images converted from scholarly PDFs). We denote this scenario as text-to-visual document (T$\rightarrow$VD) retrieval.
\item \textbf{Fused-Modal Retrieval}: More complicated retrieval tasks involve mixed modalities in queries, candidates, or both. For example, in EVQA~\cite{DBLP:conf/iccv/MensinkUCGCZSAF23}, both queries and candidates combine text and images.
\end{itemize}
The visualization of these settings refers to Figure \ref{fig:umr-setting}.

\begin{table}
\resizebox{1.0\columnwidth}{!}{
\begin{tabular}{c|c|l}
\toprule
\bf Class & \bf Task & \bf Datasets \\ \midrule
\multirow{2}{*}{
\begin{tabular}{c}Single-\\Modal\\(17)\end{tabular}
}         & T$\rightarrow$T (16)  &
\begin{tabular}[c]{@{}l@{}}
ArguAna\cite{DBLP:conf/acl/WachsmuthSS18}
Climate-FEVER\cite{DBLP:journals/corr/abs-2012-00614} \\ 
CQADupStack\cite{DBLP:conf/adcs/HoogeveenVB15} 
DBPedia\cite{DBLP:conf/sigir/HasibiNXBBKC17} 
FEVER\cite{DBLP:conf/naacl/ThorneVCM18} \\ 
FiQA2018\cite{DBLP:conf/www/MaiaHFDMZB18} 
HotpotQA\cite{DBLP:conf/emnlp/Yang0ZBCSM18} \\ 
MSMARCO\cite{DBLP:conf/nips/NguyenRSGTMD16}
NFCorpus\cite{DBLP:conf/ecir/BotevaGSR16} 
NQ\cite{DBLP:journals/tacl/KwiatkowskiPRCP19}  \\ 
Quora\tablefootnote{
More details can be found at \href{https://www.quora.com/q/quoradata/First-Quora-Dataset-Release-Question-Pairs}{Quora Dataset Release: Question Pairs}. 
}
SCIDOCS\cite{cohan-etal-2020-specter} 
SciFact\cite{DBLP:conf/emnlp/WaddenLLWZCH20}  \\ 
Touche2020\cite{DBLP:conf/clef/BondarenkoFBGAP20}
TRECCOVID\cite{DBLP:journals/sigir/VoorheesABDHLRS20} 
WebQA\cite{DBLP:conf/cvpr/ChangCNGSB22}
\end{tabular} \\ \cline{2-3}
& I$\rightarrow$I (1)            & Nights\cite{fu2023dreamsim}  \\ \hline

\multirow{9}{*}{
\begin{tabular}{c}Cross-\\Modal\\(18)\end{tabular}
}    & T$\rightarrow$I (4)  & \begin{tabular}[c]{@{}l@{}}
VisualNews\cite{DBLP:conf/emnlp/LiuWWO21} Fashion200k\cite{DBLP:conf/iccv/HanWHZZLZD17} \\ MSCOCO\cite{lin2014microsoft} Flickr30k\cite{DBLP:conf/iccv/PlummerWCCHL15} \end{tabular}

\\ \cline{2-3} 
& T$\rightarrow$VD (10) & \begin{tabular}[c]{@{}l@{}}
TAT-DQA\cite{DBLP:conf/mm/ZhuLFWZC22}
ArxivQA\cite{DBLP:conf/acl/0039WXWFK024} \\
DocVQA\cite{DBLP:conf/wacv/MathewKJ21} 
InfoVQA\cite{DBLP:conf/wacv/MathewBTKVJ22} \\
Shift Project$^\dagger$ 
Artificial Intelligence$^\dagger$ \\
Government Reports$^\dagger$ 
Healthcare Industry$^\dagger$ \\ 
Energy $^\dagger$
TabFQuad$^\dagger$ \end{tabular}                     

\\ \cline{2-3} 
& I$\rightarrow$T (4)        & \begin{tabular}[c]{@{}l@{}}VisualNews\cite{DBLP:conf/emnlp/LiuWWO21} 
Fashion200K\cite{DBLP:conf/iccv/HanWHZZLZD17} \\ 
MSCOCO\cite{lin2014microsoft} 
Flickr30k\cite{DBLP:conf/iccv/PlummerWCCHL15} \end{tabular}                                                                                                                    \\ \hline
\multirow{5}{*}{
\begin{tabular}{c}Fused-\\Modal\\(12)\end{tabular}
} & T$\rightarrow$IT (2)          & 
WebQA\cite{DBLP:conf/cvpr/ChangCNGSB22} 
EDIS\cite{DBLP:conf/emnlp/LiuFFCW23}   \\ 

\cline{2-3} 
& IT$\rightarrow$T (5)         & \begin{tabular}[c]{@{}l@{}}
OVEN\cite{DBLP:conf/iccv/HuLCKJLTC23} 
INFOSEEK\cite{DBLP:conf/emnlp/ChenHLSCRC23} \\ 
ReMuQ\cite{luo-etal-2023-end} 
OKVQA\cite{DBLP:conf/cvpr/MarinoRFM19} 
LLaVA\cite{lin-etal-2024-preflmr}    
\end{tabular}                                                         

\\ \cline{2-3} 
& IT$\rightarrow$I (2)         & 
FashionIQ\cite{DBLP:conf/cvpr/WuGGARGF21} 
CIRR\cite{Liu2021ImageRO}                                                                 

\\ \cline{2-3} 
& IT$\rightarrow$IT (3)      & 
OVEN\cite{DBLP:conf/iccv/HuLCKJLTC23} 
EVQA\cite{DBLP:conf/iccv/MensinkUCGCZSAF23} 
INFOSEEK\cite{DBLP:conf/emnlp/ChenHLSCRC23}                                               \\ \bottomrule
\end{tabular}}
\caption{
An overview of tasks and datasets in our UMRB. $\dagger$ means that they all originate from \cite{faysse2024colpali}.
}
\label{tab:umrb-datasets}
\end{table}

\subsection{Universal Multimodal Retrieval Benchmark}
Based on the aforementioned classification principles, we introduce a new benchmark to comprehensively assess the performance of UMR models. This benchmark comprises \textbf{47} evaluation datasets that cover a broad spectrum of multimodal retrieval tasks, and we name it the Universal Multimodal Retrieval Benchmark (UMRB). These evaluation datasets primarily originate from previously constructed datasets tailored for each sub-scenario or sub-task.
Specifically, UMRB includes: (1) The BEIR~\cite{thakur2beir} benchmark for text-to-text retrieval scenarios; (2) The M-BEIR~\cite{wei2023uniir} dataset for vision-centric retrieval scenarios; (3) Additional fused-modal datasets that not cover by M-BEIR; and (4) text-to-visual document search datasets, such as ViDoRe \cite{faysse2024colpali}, to extend the coverage of our benchmark and ensure a comprehensive evaluation of model universality.
A detailed list of the UMRB datasets is presented in Table \ref{tab:umrb-datasets}. 

Given the extensive size of UMRB, to expedite our experimental validation and analysis, we have sampled a subset of datasets from each category, constituting a smaller dataset named UMRB-Partial. This subset retains $39\%$ of the total datasets while maintaining evaluation richness. More detailed statistical information about UMRB-Partial can be found in Appendix Table \ref{tab:umr_tasks}.

\section{Method}
\label{sec:method}
In this section, we present the training framework for developing the General Multimodal Embedder (GME) model. We describe the contrastive learning approach used to train the embedding model. Building on this, we conduct detailed experiments to determine the optimal balance of training data type. Specifically, our experiments demonstrate that diverse data type mixtures significantly enhances the model's ability to perform retrieval across various modalities. Lastly, recognizing the scarcity of high-quality fused-modal training data, we propose a novel method for automatically synthesizing large-scale, high-quality training data using MLLM.

\begin{figure}
\centering
\includegraphics[width=0.9\columnwidth]{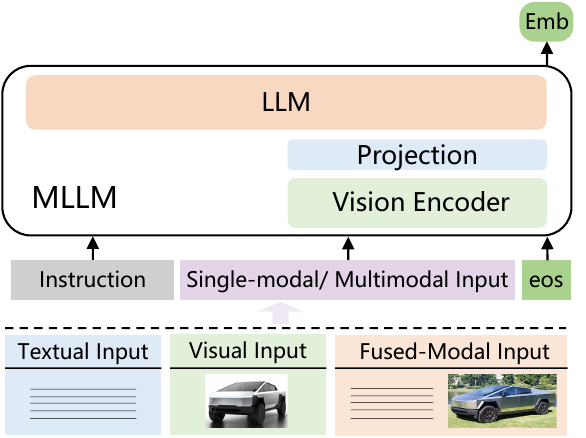}
\caption{The \texttt{GME} model architecture. \texttt{Emb} denotes the embedding of the input content.}
\label{fig:model}
\end{figure}

\subsection{GME: General Multimodal Embedder}

\paragraph{Model Architecture} We employ a MLLM as the foundation for GME. This model can accept images, text, or image-text pairs as input. Inspired by previous research on text embedding~\cite{li2023generaltextembeddingsmultistage,DBLP:conf/acl/WangYHYMW24}, we use the final hidden state of the last token as the representation (or embedding) for the input. Although pre-trained MLLMs possess strong multimodal understanding capabilities, their original training objectives are not optimized for representation learning. Therefore, task-specific fine-tuning (or alignment) is necessary to enhance the model's representational capacity. Contrastive learning has been shown to effectively train LLMs and MLLMs to produce retrieval embeddings~\cite{li2023generaltextembeddingsmultistage,jiang2024e5v}.

\paragraph{Contrastive Learning} In our contrastive learning setup, each training instance comprises a query \( q \), a relevant candidate \( c \), and a set of irrelevant candidates \( \{c_1^-, c_2^-, \ldots, c_K^-\} \). Both \( q \) and \( c \) can be text, images, or image-text pairs, allowing the model to handle diverse data modalities. To tailor the model to various downstream retrieval tasks, we incorporate an instruction tuning method by including a tailored instructional text \( i \) with each retrieval task. For example, for the Visual Question Answering (VQA) task, the instruction could be: ``Retrieve a passage that provides an answer to the given query about the image'' guiding the model on how to process and interpret the query for specific objectives.

During training, we input \( q \) and instruction \( i \) into the model to obtain the query representation \( e_q \). Similarly, each candidate \( c \) is input into the model to obtain its representation \( e_c \). The training objective minimizes the cosine distance between \( e_q \) and \( e_c \) for relevant pairs while maximizing the distance between \( e_q \) and \( e_{c^-} \) for irrelevant pairs. Cosine similarity is employed to measure the directional alignment between embeddings, effectively capturing semantic similarities irrespective of their magnitudes.

The optimization process utilizes the InfoNCE loss function~\cite{oord2019representationlearningcontrastivepredictive}, defined as:
\begin{equation*}
\mathcal{L} = -\log \frac{\exp\left(cos(e_q, e_c^+) / \tau\right)}{\exp\left(cos(e_q, e_c^+) / \tau\right) + \sum\limits_{i=1}^{K} \exp\left(cos(e_q, e_{c_i^-}) / \tau\right)}
\end{equation*}
where \(\tau\) is the temperature parameter that scales the cosine similarities to control the distribution's concentration. This approach ensures that the model effectively learns to distinguish relevant from irrelevant information across different modalities, thereby enhancing its performance in multimodal retrieval tasks.

\paragraph{Hard Negatives}
The quality and diversity of negative samples are essential for improving contrastive learning \cite{Robinson2020ContrastiveLW}. Inspired by ANCE~\cite{DBLP:conf/iclr/XiongXLTLBAO21}, we employ a two-stage training strategy: (1) Initial Training: We first train the model using randomly selected negative candidates, resulting in Model \( M_1 \).
(2) Hard Negative Mining and Continue Training: Using \( M_1 \), we retrieve the top \( K \) candidates for each query and select non-relevant candidates from them as hard negatives. We then use these hard negatives to further train \( M_1 \), refining it into the final model.
This ensures that the model can learn from both easily distinguishable and more challenging examples, thereby enhancing performance.

\label{sec4} \paragraph{Training Data Composition} A critical factor in multimodal representation learning is the composition of training data. Although previous studies like \cite{jiang2024e5v} have demonstrated that MLLMs can develop multimodal representation capabilities after being fine-tuned on single-modal data, the effect of data diversity on model performance remains unclear. Therefore, we compare the performance of models trained with different data combinations across various retrieval scenarios within our classification principle. Specifically, we used four types of training data:
single-modal (including T$\rightarrow$T and I$\rightarrow$I), cross-modal (including T$\rightarrow$VD and T$\rightarrow$I), fused-modal training data (including IT$\rightarrow$IT), and a mixed dataset combining the first three types. These different training data types result in a total of six models.

\begin{figure}
    \centering
    \includegraphics[width=1.0\columnwidth]{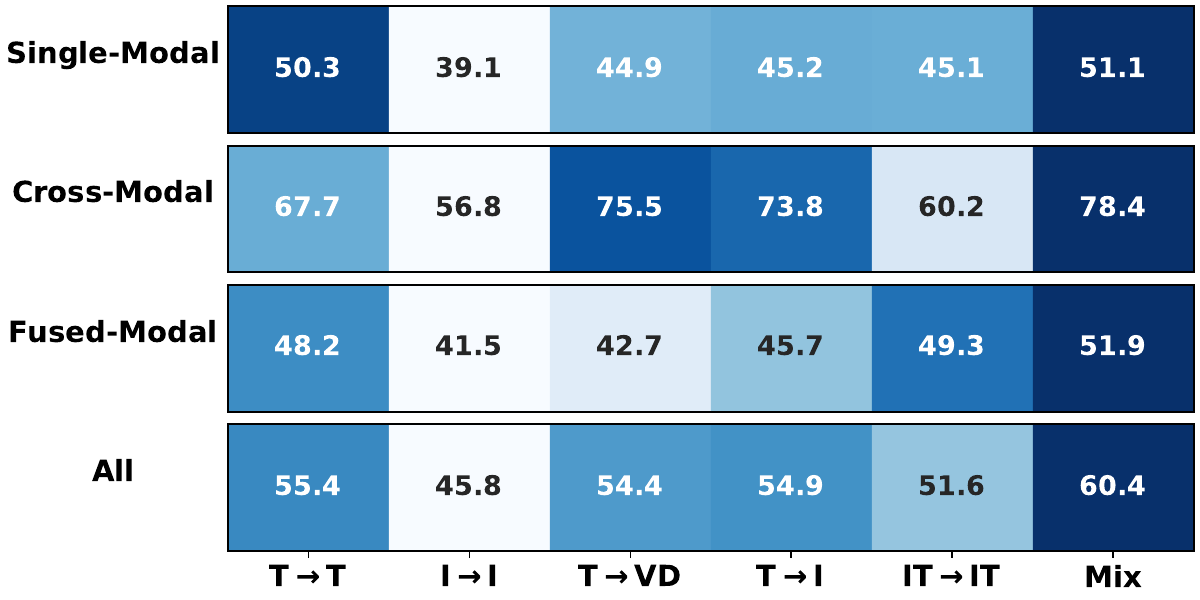}
    \caption{Impact of training data on multimodal retrieval tasks.}
    \label{fig:data_mix}
\end{figure}

For single-modal data, we utilized the T$\rightarrow$T dataset from MSMARCO~\cite{DBLP:conf/nips/NguyenRSGTMD16} and the I$\rightarrow$I dataset from ImageNet~\cite{5206848}, treating images within the same category as positive matches and those from different categories as negatives. For cross-modal data, we employed T$\rightarrow$I pairs from the LAION~\cite{schuhmann2022laion5bopenlargescaledataset} dataset and T$\rightarrow$VD pairs from the Docmatix~\cite{docmatix} dataset. For fused-modal data, we use the EVQA~\cite{DBLP:conf/iccv/MensinkUCGCZSAF23} dataset (IT$\rightarrow$IT). For each subcategory, we randomly sampled 100,000 training instances to train the models independently. For the mixed dataset, we uniformly sampled 20,000 instances from each of the five datasets to train the final model, ensuring fair and reliable comparative experimental results. The performance of these six models on the UMRB-Partial test dataset is presented in Figure ~\ref{fig:data_mix}.

The results indicate that: (1) Models trained on single data types excel in corresponding retrieval tasks. For instance, models trained on T$\rightarrow$T data performed best in text retrieval tasks.\footnote{Detail results are shown in the Appendix Table \ref{tab:umrb_partial_results}.} (2) A balanced mix of different data types enhanced performance across various settings. This suggests that increasing the diversity of training modalities effectively improves the model's overall retrieval capabilities.

The above analysis highlights the importance of adequately representing each data type in training datasets to develop models that meet the requirements of universal multi-modal retrieval. During data collection, we observed that single-modal and cross-modal data are abundant, with over ten million training instances available. In contrast, fused-modal data remains limited. Common fused-modal training datasets such as EVQA\cite{DBLP:conf/iccv/MensinkUCGCZSAF23}, INFOSEEK\cite{DBLP:conf/emnlp/ChenHLSCRC23}, and CIRR~\cite{Liu2021ImageRO} collectively contain fewer than one million instances. Additionally, these existing fused-modal datasets cover only a limited range of domains. Thus, efficiently supplementing high-quality fused-modal training data is essential. To address this challenge, we propose leveraging the generative capabilities of LLMs and MLLMs to synthesize additional training data.

\begin{figure}
\centering
\includegraphics[width=\columnwidth]{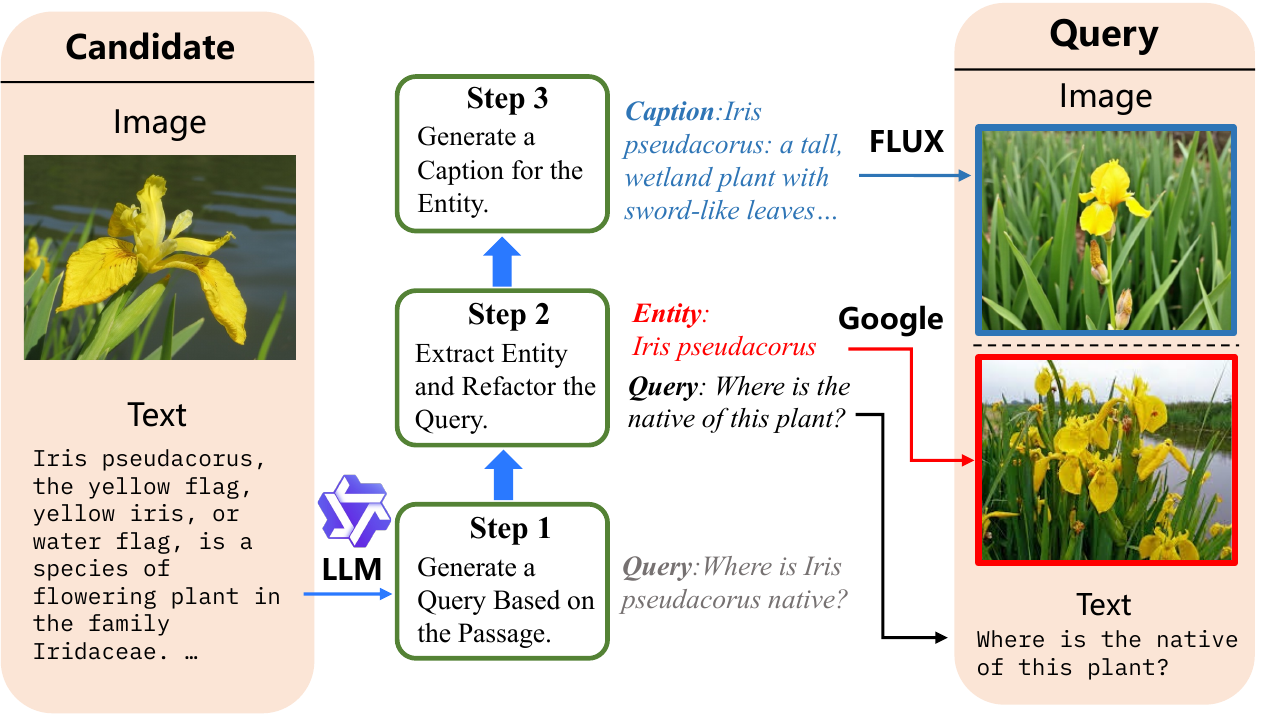}
\caption{Pipeline for synthesizing fused-modal training data.}
\label{fig:pipeline}
\end{figure}

\subsection{Fused-Modal Data Synthesis}\label{sec:data-gen}

To efficiently synthesize high-quality data while minimizing manual intervention, we adopt a strategy similar to Doc2Query~\cite{DBLP:conf/ecir/GospodinovMM23}. However, our approach differs in that we aim to generate fuse-modal candidate-to-query relevance data instead of single-modality, text-based relevance pairs. This requires obtaining high-quality candidates that include both image and text content. We primarily extracted such data from Wikipedia paragraphs\footnote{\href{https://github.com/google-research-datasets/wit/blob/main/wikiweb2m.md}{github.com/google-research-datasets/wit/blob/main/wikiweb2m.md}}. Additionally, to enhance the domain diversity of the candidate data, we employed a domain classification model\footnote{\href{https://hf.co/facebook/bart-large-mnlifacebook}{hf.co/facebook/bart-large-mnlifacebook}} to perform fine-grained classification of Wikipedia data into categories such as animals and plants. We then uniformly sampled from these categories and retained data with classification confidence scores above 0.5. Ultimately, we obtained 313,284 candidate entries, each containing both text and image content.

Based on the prepared data, the overall synthesis pipeline (Figure \ref{fig:pipeline}) could be divided into the following steps: 

\noindent$\bullet$ \textbf{Doc2Query Generation}: The passage content from each candidate is input into an LLM\footnote{In the entire pipeline, we utilize \href{https://huggingface.co/Qwen/Qwen2.5-72B-Instruct}{Qwen2.5-72B-Instruct} as our LLM.} using a prompt to generate a natural query. To ensure the quality of the generated queries, we built a vector index of all passage contents using a text vector retrieval model\footnote{\href{https://huggingface.co/Alibaba-NLP/gte-Qwen2-1.5B-instruct}{hf.co/Alibaba-NLP/gte-Qwen2-1.5B-instruct}}. Each generated query is then used to retrieve the corresponding passage from this collection. If the passage associated with the query is not within the top 20 retrieved items, the query is considered low quality due to low relevance and is discarded. In this step, we discarded 1.2\% of the total generated queries. This process allows us to construct T$\rightarrow$IT training data.

\noindent$\bullet$ \textbf{Entity Extraction and Query Rewrite}: We aim for the synthesized queries to include both texts and images (i.e., IT$\rightarrow$IT type). To achieve this, we leverage entity extraction followed by image retrieval for the extracted entities and caption generation to supplement the image data on the query side. Specifically, for each generated query \( q \) from the first step, we prompt the LLM to extract entities from it with the text passage as reference, and then rewrite the original query into \( q' \). For example, the query ``Where is Iris pseudacorus native?'' is transformed by the model to the rewritten query ''Where is the native habitat of this plant?'' with the entity ''Iris pseudacorus'' extracted. We then seek images that match this entity and combine them with the rewritten query \( q' \) to form the final fuse-modal query.

\noindent$\bullet$ \textbf{Image Retrieval and Generation}: We explore two methods for obtaining images. The first method uses the Google Image Search API\footnote{\href{https://serpapi.com/google-images-api}{https://serpapi.com/google-images-api}} to retrieve images matching the entity terms, retaining the top five results. The second method involves generating images using a text-to-image model\footnote{
\href{https://huggingface.co/black-forest-labs/FLUX.1-dev}{https://hf.co/black-forest-labs/FLUX.1-dev}}. Specifically, we first use the LLM to generate a caption suitable for image generation based on the entity and the passage of the generated query, then input this caption into the text-to-image generation model to create the corresponding image. This approach allows us to quickly and efficiently obtain high-quality, diverse images. The synthesized results can also be assembled into IT$\rightarrow$ IT retrieval type data.

\noindent$\bullet$ \textbf{Data Filtering}: To ensure the quality of the synthesized data, we perform filtering \cite{DBLP:conf/iclr/DaiZMLNLBGHC23} on the final dataset. We observe that images generated by the FLUX model have consistent quality, whereas images retrieved via the Google Image Search API often include noisy data. Therefore, for images obtained through the Google Image Search API, we use the CLIP model\footnote{\href{https://huggingface.co/openai/clip-vit-large-patch14}{https://hf.co/openai/clip-vit-large-patch14}} to assess image-caption relevance. Images with a relevance score below $0.2$ were filtered out.

Through the synthesis pipeline, we produce 1,135,000 high-quality fuse-modal training data entries (including T$\rightarrow$IT and IT$\rightarrow$IT types). After filtering, we retain 1,102,000 entries, resulting in a data loss rate of $2.9\%$. The entire process consumed 600 A100 GPU hours.
Detailed descriptions of all prompts used in the data synthesis pipeline and examples of the synthesized data are provided in the Appendix \S\ref{app:data-synthesis}.

\begin{table*}
\resizebox{\textwidth}{!}{
\setlength{\tabcolsep}{2pt}
\begin{tabular}{c|c|cc|ccc|cccc|c}
\toprule
\texttt{UMRB} &
\multicolumn{1}{c|}{Size}  & \multicolumn{2}{c|}{Single-Modal} & \multicolumn{3}{c|}{Cross-Modal} & \multicolumn{4}{c|}{Fused-Modal} & Avg. \\ \midrule
Task (\#Datasets)  & &
T$\rightarrow$T (16) &
I$\rightarrow$I (1) &
T$\rightarrow$I (4) &
T$\rightarrow$VD (10) &
I$\rightarrow$T (4) &
T$\rightarrow$IT (2) &
IT$\rightarrow$T (5) &
IT$\rightarrow$I (2) &
IT$\rightarrow$IT (3) &
(47) \\ \midrule
VISTA \cite{zhou-etal-2024-vista}  & 0.2B   & 55.15  & \bf 31.98 & 32.88 & 10.12 & 31.23 & 45.81 & 53.32 & 8.97  & 26.26 &  37.32 \\
CLIP-SF \cite{wei2023uniir}    & 0.4B & 39.75  & 31.42 & 59.05 & 24.09 & 62.95 & 66.41 & 53.32 & 34.90 & 55.65 &  43.66 \\
One-Peace \cite{wang2023one-peace}  & 4B & 43.54 & 31.27 & 61.38 &  42.9 & 65.59 & 42.72 & 28.29 & 6.73  & 23.41 & 42.01 \\
DSE \cite{ma-etal-2024-unifying}   & 4.2B     & 48.94 & 27.92 & 40.75  & 78.21 &  52.54  & 49.62  & 35.44  &  8.36 &  40.18 & 50.04 \\
E5-V \cite{jiang2024e5v}  & 8.4B & 52.41 & 27.36 & 46.56 &   41.22    & 47.95 & 54.13 & 32.90 & 23.17 & 7.23  &  42.52 \\ \midrule
\textbf{GME}-Qwen2VL-2B & 2.2B& 55.93 &  29.86  & 57.36 & 87.84 & 61.93  & 76.47  &  64.58  &  37.02  &   66.47 &  64.45 \\
\textbf{GME}-Qwen2VL-7B & 8.2B & \bf 58.19 & 31.89  &  \bf 61.35  & \bf 89.92 & \textbf{65.83}  &  \bf 80.94  & \bf 66.18  &  \bf 42.56  &  \bf 73.62  & \bf 67.44 \\ \bottomrule
\end{tabular}}
\caption{Results of different models on our benchmark. Following previous works~\cite{thakur2beir,wei2023uniir,faysse2024colpali}, we present NDCG@10 scores for T$\rightarrow$T tasks, excluding the WebQA dataset. For T$\rightarrow$VD tasks, we provide NDCG@5 scores. For the Fashion200K, FashionIQ and OKVQA datasets, we report Recall@10 scores, while for all other datasets, we report Recall@5 scores.}
\label{tab:main-results}
\end{table*}

\section{Experiments}

\subsection{Settings}
\label{sec:exp-setting}
\paragraph{Training Data} Building on the findings from \S\ref{sec4}, we train our model using a diverse dataset of \textbf{8} million instances spanning various retrieval modalities. For single-modal retrieval tasks, we utilize datasets including MSMARCO~\cite{DBLP:conf/nips/NguyenRSGTMD16}, NQ~\cite{DBLP:journals/tacl/KwiatkowskiPRCP19}, HotpotQA~\cite{DBLP:conf/emnlp/Yang0ZBCSM18}, TriviaQA~\cite{2017arXivtriviaqa}, SQuAD~\cite{rajpurkar-etal-2016-squad}, FEVER~\cite{DBLP:conf/naacl/ThorneVCM18}, and AllNLI for SimCSE~\cite{gao-etal-2021-simcse}, selecting a total of 1 million entries. From ImageNet~\cite{5206848}, we extract 1 million image-to-image training instances, designating images within the same class as positive samples and others as negative samples. For cross-modal retrieval tasks, we incorporate 2 million entries from the LAION~\cite{schuhmann2022laion5bopenlargescaledataset}, MSCOCO~\cite{lin2014microsoft}, and Docmatix~\cite{docmatix} datasets. Additionally, for fused-modal retrieval tasks, we include a total of 2 million instances: $1.1$ million synthesized by us, and the remaining from the M-BEIR~\cite{wei2023uniir} training data.

\paragraph{Training Configuration} We use Qwen2-VL~\cite{wang2024qwen2vl} model series as the backbone for our MLLM, conducting training on models with both 2 billion (2B) and 7 billion (7B) parameters. Our training utilizes Low-Rank Adaptation (LoRA) \cite{DBLP:conf/iclr/HuSWALWWC22} with a rank of 8, a learning rate of 1e-4, and a temperature setting of 0.03. To manage the varying number of visual tokens required by Qwen2-VL for different image resolutions and maintain training efficiency, we limit the maximum number of visual tokens per image to 1,024.

For data with images, we set the maximum text length to 1,800 tokens, using a batch size of 128 for the 2B model and 32 for the 7B model. For text-only data, the maximum length was set to 512 tokens, with batch size of 512 for the 2B model and 128 for the 7B model. Each training sample included 8 negative examples. To conserve GPU memory, we employ gradient checkpointing~\cite{chen2016training} and train the model using bfloat16 precision.
All training was conducted on eight NVIDIA A100 GPUs, each with 80GB of memory.

\paragraph{Baselines}
We compare our method against four types of retrieval systems:
(1) Previous representative UMR models, for example, VISTA \cite{zhou-etal-2024-vista} for text encoder based, and E5-V \cite{jiang2024e5v} for MLLM based;
(2) Powerful multimodal representation (embedding) models, \ie One-Peace \cite{wang2023one-peace}, which supports modalities beyond text and image and hence could also be tested on our \texttt{UMRB};
(3) Recent visual document retrieval models, namely DSE \cite{ma-etal-2024-unifying};
and (4) the classic cross-modal pipeline, CLIP score-fusion, denoted as CLIP-SF, which provides top-tier cross-modal performance.
We exclude comparisons with state-of-the-art text retrieval models as VISTA demonstrates comparable performance levels.

\subsection{Main Results}

Table \ref{tab:main-results} presents the evaluation results of the baseline systems alongside our proposed \texttt{GME}. Scores are averaged across each sub-task and categorized by retrieval modality type: single-modal, cross-modal, and fused-modal. Additionally, the overall micro-average score on the \texttt{UMRB} is in the last column.
First, focusing on the average scores, our smaller model, \ie \texttt{GME-Qwen2-VL-2B}, already outperforms the previous state-of-the-art UMR model (VISTA \cite{zhou-etal-2024-vista}). The larger model, \ie \texttt{GME-Qwen2-VL-7B}, further enhances this performance, demonstrating the effectiveness of our approach in handling UMR tasks.

Second, our models outperform smaller methods such as VISTA (million-level parameters) and One-Peace (4B parameters). The larger MLLM baseline, E5-V \cite{jiang2024e5v} (8B parameters), performs well in text-dominated tasks (e.g., T$\rightarrow$T) but falls short in other areas. This indicates that training with multimodal data is crucial for achieving superior performance in UMR tasks. Our training data provides a stronger foundation for future advancements.

Next, the cross-modal pipeline CLIP-SF outperforms UMR models like VISTA, E5-V, and One-Peace. For VISTA and E5-V, the performance gap is likely due to limitations in their text-modality bounds: VISTA is constrained by the text embedding space of its fixed backbone, and E5-V is limited by text-only training. One-Peace's modality alignment-centered modeling may not be optimized for fused-modal content. In contrast, our models are specifically designed to handle fused-modal data, resulting in significantly better performance compared to the baselines. Although our training data includes several previously constructed fused-modal datasets, the contribution of our generated fused-modal training data will be discussed in \S\ref{sec:analysis}.

Finally, we compare with the recent visual document retrieval model DSE \cite{ma-etal-2024-unifying}, specialized for the T$\rightarrow$VD task within the Cross-Modal group, which has approximately 4B parameters. 
Our models are competitive with or exceed the performance of this task-specific baseline, demonstrating the feasibility and promise of integrating visual document retrieval into a unified retriever framework.

\subsection{Analyses}
\label{sec:analysis}

\begin{figure}
\centering
\includegraphics[width=0.325\linewidth]{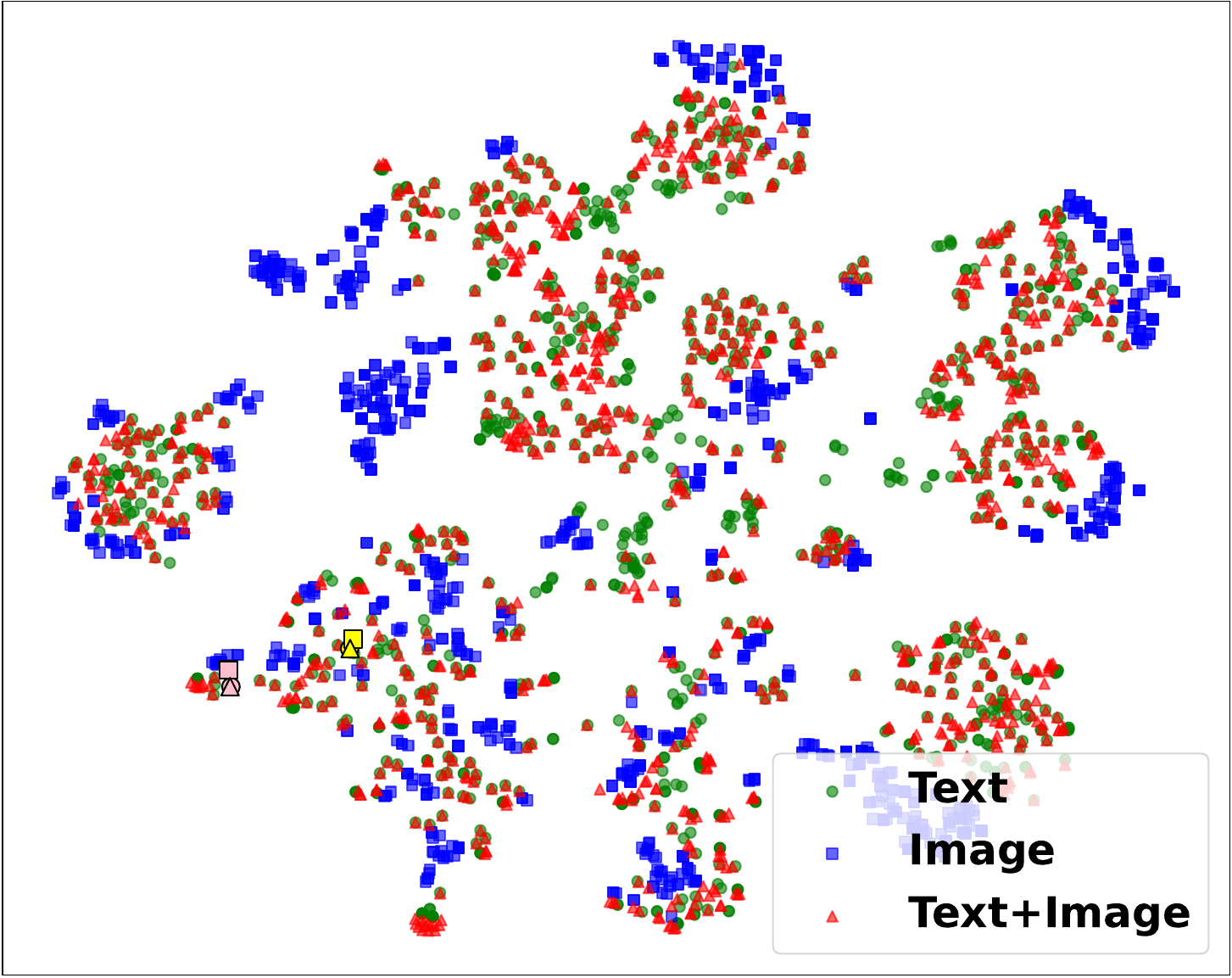}
\includegraphics[width=0.325\linewidth]{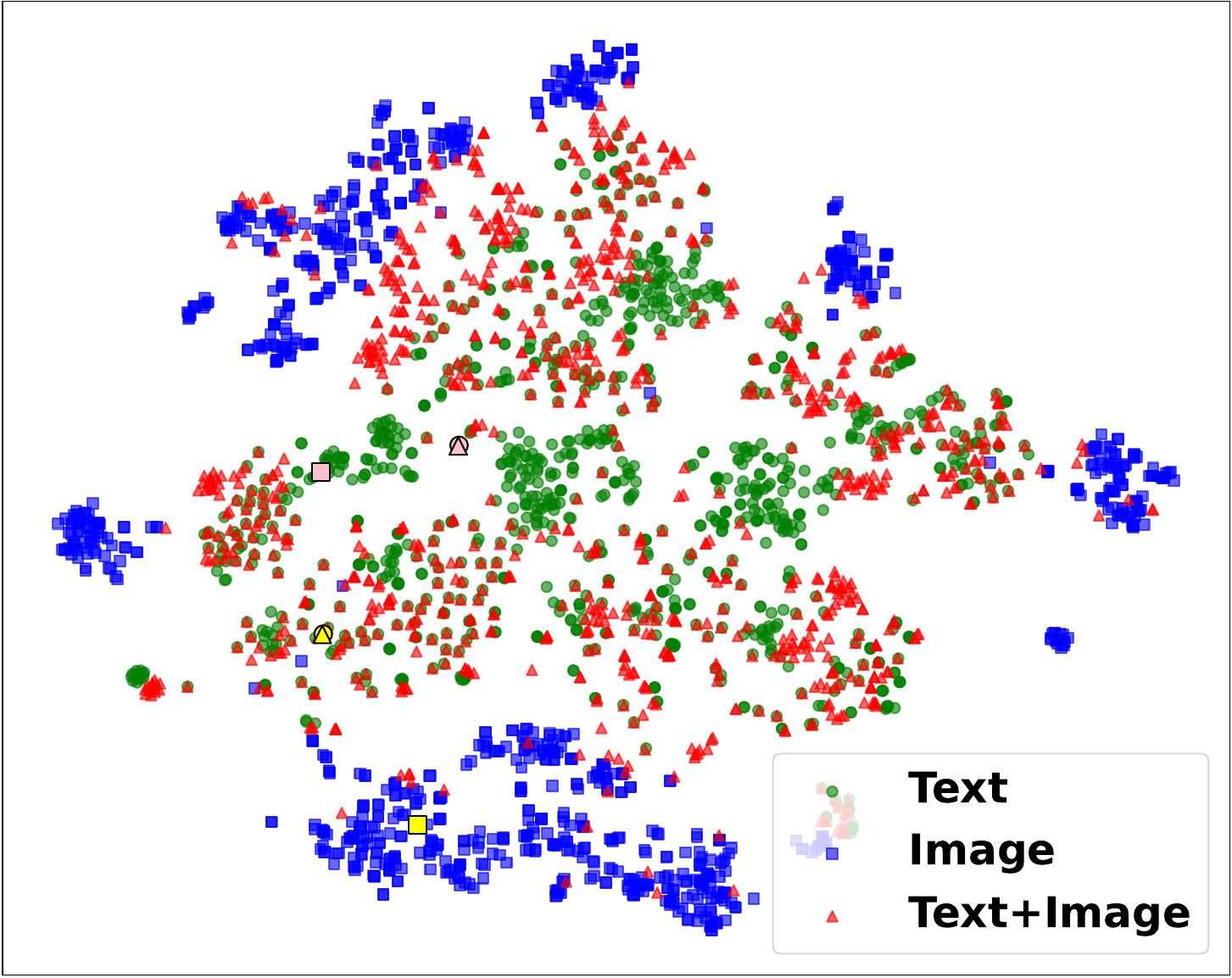}
\includegraphics[width=0.325\linewidth]{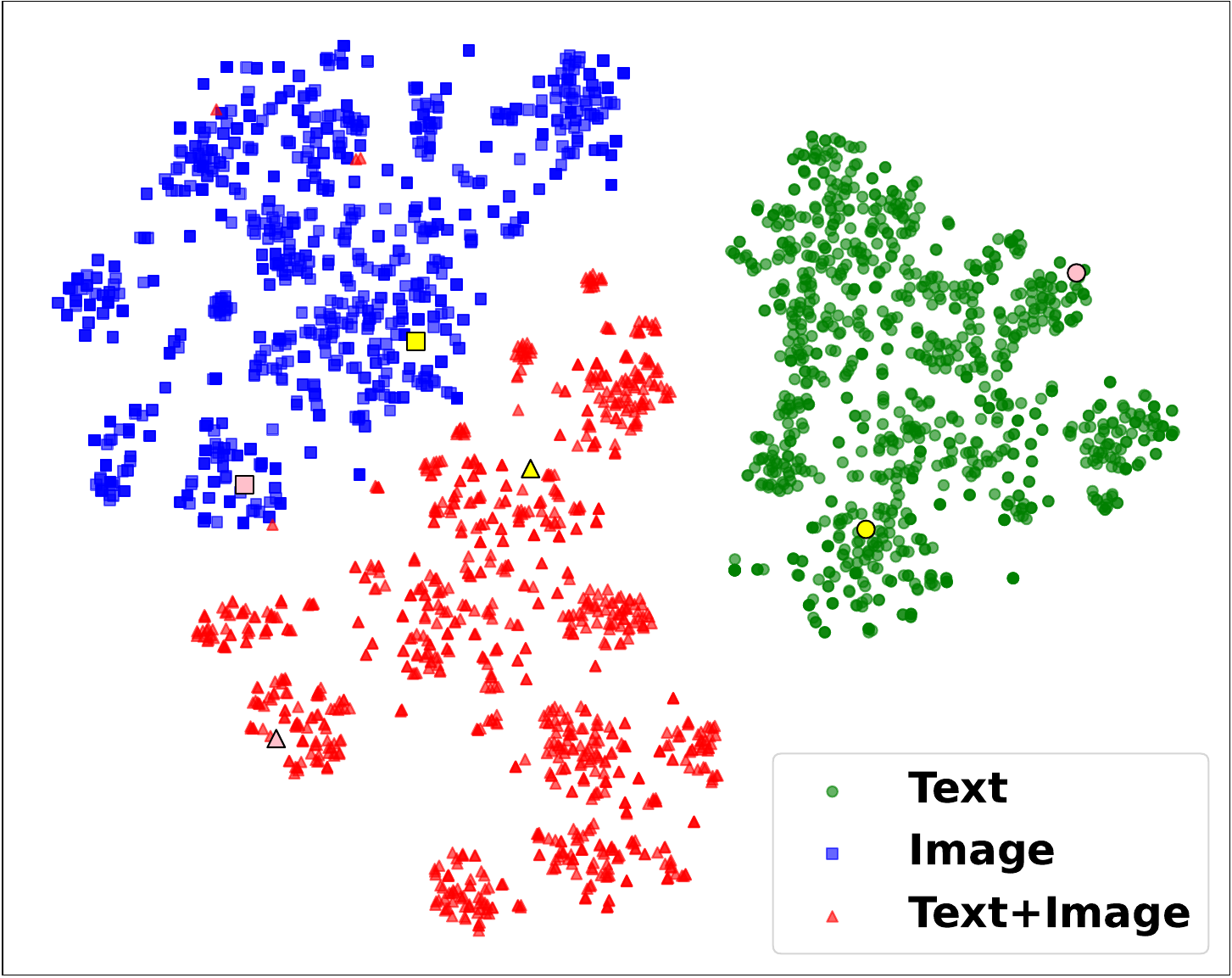}
\caption{
Visualization of the embeddings in a 2D plot by T-SNE.
Left: Our GME, Middle: VISTA, Right: CLIP.
We use instances from Encyclopedia VQA and highlight two semantic groups with yellow and pink labels, respectively.
Please zoom in to view them.
}
\label{fig:dist}
\end{figure}

\paragraph{Are the Produced Embeddings Modality Universal?}
Given our the impressive performance of our model, we assess the quality of its embeddings. Specifically, we investigate whether the embeddings are modality-universal meaning that embeddings representing the same semantic content across different modalities are closely clustered in the embedding space, or if they remain in separate sub-spaces tailored for each modality-specific task. To probe this question, we sample 1000 instances from the EVQA dataset and visualize their embeddings of different modalities by t-SNE, as shown in Figure \ref{fig:dist}.
We also highlight two semantic close groups with yellow and pink labels, respectively.
We can observe that the embeddings from CLIP are distinctly separated by modality, whereas the embeddings from our model are intermingled and organized semantically.
Meanwhile, the points from the same semantic group are closely clustered.
This demonstrates that our model effectively generates modality-universal representations, enhancing its applicability across various UMR tasks.

\begin{table}
\centering
\resizebox{0.8\linewidth}{!}{
\begin{tabular}{c|ccc|c}
\toprule
Setting   & Single & Cross & Fused & Average \\ \midrule
w/ EVQA    &  45.13    &   60.21  & 49.32  &  51.55\\
w/ Gen$_{\text{Flux}}$ & 46.27   & 61.19    &  51.46  & 52.97 \\
w/ Gen$_{\text{Google}}$ & 47.08  &   61.35 & 52.01  & 53.48 \\
\bottomrule
\end{tabular}
}
\caption{Results of GME-Qwen2-VL-2B trained with different generated datasets and evaluated on UMRB-Partial.}
\label{tab:ablation_synthesis}
\end{table}

\paragraph{Ablation Study on Synthetic Fused-Modal Data}
We propose an efficient data synthesis pipeline (\S\ref{sec:data-gen}) and generate large-scale fused-modal pairs to support model training. After witnessing the state-of-the-art performance of our model, it is natural to question the contribution of this synthetic data to the overall performance. To this end, we conduct an ablation study using three parallel training datasets, each comprising 100,000 pairs: original EVQA data, synthetic data with Google-retrieved images (Gen\textsubscript{Google}), and synthetic data with FLUX-generated images (Gen\textsubscript{Flux}). We train three models with identical parameters on these datasets and evaluate their performance on UMRB-Partial, with results shown in Table \ref{tab:ablation_synthesis}. Both synthetic datasets outperform the original EVQA data, indicating the high quality of our synthesized data. Although Google-retrieved images achieved marginally better performance than FLUX-generated images, the difference is minor and acceptable given the potential limitations of the Google Search API for rapid, large-scale dataset generation.

\begin{figure}
\centering
\includegraphics[width=0.9\columnwidth]{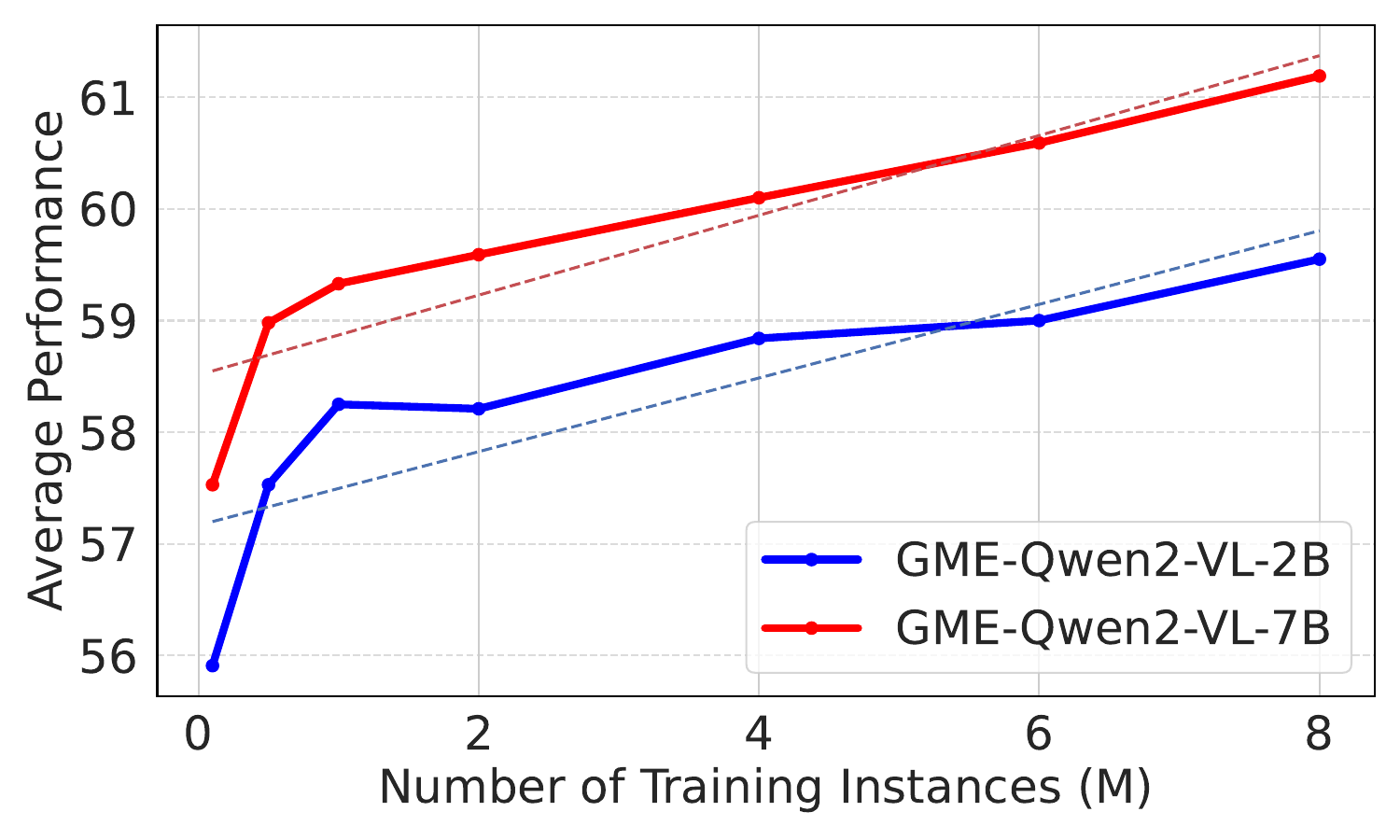}
\caption{Average Performance of GME-Qwen2-VL-2B (Blue) and GME-Qwen2-VL-7B (Red) on UMRB-Partial, trained with varying numbers of training instances.}
\label{fig:scaling-law}
\end{figure}

\paragraph{Training Scaling Law}
Our approach is primarily data-centric, constructing a diverse training dataset of approximately $8$ million samples across various UMR settings (\S\ref{sec:exp-setting}). Training on such a large-scale dataset demands significant computational resources and time.
Therefore, we explored the training scaling law by examining how model performance evolves with increasing training steps. Due to the time-consuming nature of evaluating certain retrieval tasks, we assessed performance on our \texttt{UMRB-Partial} dataset for faster evaluation. Figure \ref{fig:scaling-law} illustrates the performance progression of our 2B and 7B models on \texttt{UMRB-Partial} during training. Both models exhibit linear performance improvements as training continues, suggesting that extended training could yield further benefits. However, due to time constraints, we halted current training. Future work will investigate longer training periods to enhance model performance further.

\paragraph{Ablation Study on Modeling}
We conduct an ablation study to investigate the effectiveness of different design choices of \texttt{GME}.
We consider the following three aspects:
(1) Fine-tuning strategy. Our final models are trained by LoRA with rank 8. We compare with other rank values and full fine-tuning. The results in the first group of Table \ref{tab:ablation} show that LoRA with rank 8 yields the best performance. 
(2) Training data organization. We compare models trained without hard negative mining. The second group of Table \ref{tab:ablation} demonstrates that the removal of hard negatives led to performance declines, indicating that it is essential for effective retrieval model training.
(3) Retrieval instructions. We compare models trained without retrieval instructions. The third group shows that retrieval instructions are crucial for better UMR.
(4) Modeling techniques. Our final models are in the casual attention mode and use the EOS token state as the embedding, hence we compare the performance of the model trained with mean pooling and the bi-directional attention mechanism.
The last group of Table \ref{tab:ablation} shows that these alternative settings negatively impact performance.

\begin{table}
\centering
\resizebox{0.9\linewidth}{!}{
\begin{tabular}{c|ccc|c}
\toprule
Setting   & Single & Cross & Fused & Average \\ \midrule
\multicolumn{5}{c}{Fine-tuning strategy} \\ \midrule
LoRA r=8                    & \bf 48.09        & 78.39       & \bf 51.88       & \bf 59.45   \\
LoRA r=16                   & 47.86        & \bf 78.63       & 51.42       & 59.30    \\
LoRA r=32                   & 47.85        & 78.55       & 50.48       & 58.96   \\
LoRA r=64                   & 47.65        & 78.61       & 51.09       & 59.11   \\
Full training   & 43.16    &  75.79 &   49.28    &  56.07         \\ \midrule
\multicolumn{5}{c}{Training data organization} \\ \midrule
w/o hard-negative          &     47.55     &   78.01 &   50.95          &    58.83     \\ \midrule
\multicolumn{5}{c}{Retrieval Setting} \\ \midrule
w/o Instruction  &      46.82      & 78.10     &   49.09         &   58.00      \\ \midrule
\multicolumn{5}{c}{Model Design} \\ \midrule
w/ mean pooling   &  47.86  &  77.95    &  51.33           &    59.04     \\ \midrule
w/ bi-attention &    46.55   &   76.78  & 49.54         &   57.62 \\ \bottomrule
\end{tabular}
}
\caption{Results of the ablation study on Qwen2-VL-2B. All models are trained using 100,000 instances, consistent with the experimental setup described in Section~\ref{sec4}.}
\label{tab:ablation}
\end{table}

\section{Conclusion}
In this work, we target the universal multimodal retrieval (UMR) problem.
We begin by systematically categorizing current UMR tasks, proposing a comprehensive classification framework.
Based on this, we explore ways to further improve MLLM-based UMR models, suggesting the GME model.
The GME models are trained using contrastive learning loss on a diverse set of multimodal data settings, while also extending support for visual retrieval.
Additionally, to overcome limitations in existing UMR evaluation benchmarks, we compiled a new comprehensive benchmark (i.e., UMRB) by integrating multiple data sources.
This benchmark effectively balances existing UMR tasks with the increasingly important text and visual document retrieval tasks, enabling a more thorough assessment of UMR model performance.
We evaluate existing UMR models and our proposed GME model on UMRB, finding that our model achieves state-of-the-art performance. We also conducted various analyses to validate the effectiveness of our methods and enhance our understanding of them.
Our benchmark, models, and other materials are open-source at \url{https://hf.co/Alibaba-NLP/gme-Qwen2-VL-7B-Instruct}.

\section*{Acknowledgments}
This work receives partial support from the Natural Science Foundation of China (under Grant 624B2048) and Research Grant Council of Hong Kong (PolyU/15209724).

{
    \small
    \bibliographystyle{ieeenat_fullname}
    \bibliography{main}

\begin{thebibliography}{76}
\providecommand{\natexlab}[1]{#1}
\providecommand{\url}[1]{\texttt{#1}}
\expandafter\ifx\csname urlstyle\endcsname\relax
  \providecommand{\doi}[1]{doi: #1}\else
  \providecommand{\doi}{doi: \begingroup \urlstyle{rm}\Url}\fi

\bibitem[Bondarenko et~al.(2020)Bondarenko, Fr{\"{o}}be, Beloucif, Gienapp, Ajjour, Panchenko, Biemann, Stein, Wachsmuth, Potthast, and Hagen]{DBLP:conf/clef/BondarenkoFBGAP20}
Alexander Bondarenko, Maik Fr{\"{o}}be, Meriem Beloucif, Lukas Gienapp, Yamen Ajjour, Alexander Panchenko, Chris Biemann, Benno Stein, Henning Wachsmuth, Martin Potthast, and Matthias Hagen.
\newblock Overview of touch{\'{e}} 2020: Argument retrieval - extended abstract.
\newblock In \emph{Experimental {IR} Meets Multilinguality, Multimodality, and Interaction - {CLEF} 2020}, pages 384--395. Springer, 2020.

\bibitem[Boteva et~al.(2016)Boteva, Ghalandari, Sokolov, and Riezler]{DBLP:conf/ecir/BotevaGSR16}
Vera Boteva, Demian~Gholipour Ghalandari, Artem Sokolov, and Stefan Riezler.
\newblock A full-text learning to rank dataset for medical information retrieval.
\newblock In \emph{Advances in Information Retrieval - 38th European Conference on {IR} Research, {ECIR} 2016, Padua, Italy}, pages 716--722. Springer, 2016.

\bibitem[Brown et~al.(2020)Brown, Mann, Ryder, Subbiah, Kaplan, Dhariwal, Neelakantan, Shyam, Sastry, Askell, Agarwal, Herbert{-}Voss, Krueger, Henighan, Child, Ramesh, Ziegler, Wu, Winter, Hesse, Chen, Sigler, Litwin, Gray, Chess, Clark, Berner, McCandlish, Radford, Sutskever, and Amodei]{gpt3}
Tom~B. Brown, Benjamin Mann, Nick Ryder, Melanie Subbiah, Jared Kaplan, Prafulla Dhariwal, Arvind Neelakantan, Pranav Shyam, Girish Sastry, Amanda Askell, Sandhini Agarwal, Ariel Herbert{-}Voss, Gretchen Krueger, Tom Henighan, Rewon Child, Aditya Ramesh, Daniel~M. Ziegler, Jeffrey Wu, Clemens Winter, Christopher Hesse, Mark Chen, Eric Sigler, Mateusz Litwin, Scott Gray, Benjamin Chess, Jack Clark, Christopher Berner, Sam McCandlish, Alec Radford, Ilya Sutskever, and Dario Amodei.
\newblock Language models are few-shot learners.
\newblock In \emph{Advances in Neural Information Processing Systems 33: Annual Conference on Neural Information Processing Systems 2020, NeurIPS 2020, December 6-12, 2020, virtual}, 2020.

\bibitem[Chang et~al.(2022)Chang, Cao, Narang, Gao, Suzuki, and Bisk]{DBLP:conf/cvpr/ChangCNGSB22}
Yingshan Chang, Guihong Cao, Mridu Narang, Jianfeng Gao, Hisami Suzuki, and Yonatan Bisk.
\newblock Webqa: Multihop and multimodal {QA}.
\newblock In \emph{{IEEE/CVF} Conference on Computer Vision and Pattern Recognition}, pages 16474--16483, 2022.

\bibitem[Chen et~al.(2016)Chen, Xu, Zhang, and Guestrin]{chen2016training}
Tianqi Chen, Bing Xu, Chiyuan Zhang, and Carlos Guestrin.
\newblock Training deep nets with sublinear memory cost.
\newblock \emph{CoRR}, abs/1604.06174, 2016.

\bibitem[Chen et~al.(2023)Chen, Hu, Luan, Sun, Changpinyo, Ritter, and Chang]{DBLP:conf/emnlp/ChenHLSCRC23}
Yang Chen, Hexiang Hu, Yi Luan, Haitian Sun, Soravit Changpinyo, Alan Ritter, and Ming-Wei Chang.
\newblock Can pre-trained vision and language models answer visual information-seeking questions?
\newblock In \emph{Proceedings of the 2023 Conference on Empirical Methods in Natural Language Processing}, pages 14948--14968, Singapore, 2023. Association for Computational Linguistics.

\bibitem[Chen et~al.(2024)Chen, Wang, Tian, Ye, Gao, Cui, Tong, Hu, Luo, Ma, et~al.]{chen2024far}
Zhe Chen, Weiyun Wang, Hao Tian, Shenglong Ye, Zhangwei Gao, Erfei Cui, Wenwen Tong, Kongzhi Hu, Jiapeng Luo, Zheng Ma, et~al.
\newblock How far are we to gpt-4v? closing the gap to commercial multimodal models with open-source suites.
\newblock \emph{Science China Information Sciences}, 67\penalty0 (12):\penalty0 220101, 2024.

\bibitem[Cohan et~al.(2020)Cohan, Feldman, Beltagy, Downey, and Weld]{cohan-etal-2020-specter}
Arman Cohan, Sergey Feldman, Iz Beltagy, Doug Downey, and Daniel Weld.
\newblock {SPECTER}: Document-level representation learning using citation-informed transformers.
\newblock In \emph{Proceedings of the 58th Annual Meeting of the Association for Computational Linguistics}, pages 2270--2282, Online, 2020. Association for Computational Linguistics.

\bibitem[Dai et~al.(2023)Dai, Zhao, Ma, Luan, Ni, Lu, Bakalov, Guu, Hall, and Chang]{DBLP:conf/iclr/DaiZMLNLBGHC23}
Zhuyun Dai, Vincent~Y. Zhao, Ji Ma, Yi Luan, Jianmo Ni, Jing Lu, Anton Bakalov, Kelvin Guu, Keith~B. Hall, and Ming{-}Wei Chang.
\newblock Promptagator: Few-shot dense retrieval from 8 examples.
\newblock In \emph{The Eleventh International Conference on Learning Representations, {ICLR} 2023, Kigali, Rwanda, May 1-5, 2023}. OpenReview.net, 2023.

\bibitem[Deng et~al.(2009)Deng, Dong, Socher, Li, Li, and Fei{-}Fei]{5206848}
Jia Deng, Wei Dong, Richard Socher, Li{-}Jia Li, Kai Li, and Li Fei{-}Fei.
\newblock Imagenet: {A} large-scale hierarchical image database.
\newblock In \emph{2009 {IEEE} Computer Society Conference on Computer Vision and Pattern Recognition, Miami, {USA}}, pages 248--255. {IEEE} Computer Society, 2009.

\bibitem[Diggelmann et~al.(2020)Diggelmann, Boyd{-}Graber, Bulian, Ciaramita, and Leippold]{DBLP:journals/corr/abs-2012-00614}
Thomas Diggelmann, Jordan~L. Boyd{-}Graber, Jannis Bulian, Massimiliano Ciaramita, and Markus Leippold.
\newblock {CLIMATE-FEVER:} {A} dataset for verification of real-world climate claims.
\newblock \emph{CoRR}, abs/2012.00614, 2020.

\bibitem[Faysse et~al.(2025)Faysse, Sibille, Wu, Omrani, Viaud, Hudelot, and Colombo]{faysse2024colpali}
Manuel Faysse, Hugues Sibille, Tony Wu, Bilel Omrani, Gautier Viaud, C{\'{e}}line Hudelot, and Pierre Colombo.
\newblock Colpali: Efficient document retrieval with vision language models.
\newblock In \emph{The Thirteenth International Conference on Learning Representations}, 2025.

\bibitem[Fu et~al.(2023)Fu, Tamir, Sundaram, Chai, Zhang, Dekel, and Isola]{fu2023dreamsim}
Stephanie Fu, Netanel~Yakir Tamir, Shobhita Sundaram, Lucy Chai, Richard Zhang, Tali Dekel, and Phillip Isola.
\newblock Dreamsim: Learning new dimensions of human visual similarity using synthetic data.
\newblock In \emph{Thirty-seventh Conference on Neural Information Processing Systems}, 2023.

\bibitem[Gao et~al.(2021)Gao, Yao, and Chen]{gao-etal-2021-simcse}
Tianyu Gao, Xingcheng Yao, and Danqi Chen.
\newblock {S}im{CSE}: Simple contrastive learning of sentence embeddings.
\newblock In \emph{Proceedings of the 2021 Conference on Empirical Methods in Natural Language Processing}, pages 6894--6910, Online and Punta Cana, Dominican Republic, 2021. Association for Computational Linguistics.

\bibitem[Gospodinov et~al.(2023)Gospodinov, MacAvaney, and Macdonald]{DBLP:conf/ecir/GospodinovMM23}
Mitko Gospodinov, Sean MacAvaney, and Craig Macdonald.
\newblock Doc2query-: When less is more.
\newblock In \emph{Advances in Information Retrieval - 45th European Conference on Information Retrieval, {ECIR} 2023}, pages 414--422, Dublin, Ireland, 2023. Springer.

\bibitem[Han et~al.(2017)Han, Wu, Huang, Zhang, Zhu, Li, Zhao, and Davis]{DBLP:conf/iccv/HanWHZZLZD17}
Xintong Han, Zuxuan Wu, Phoenix~X. Huang, Xiao Zhang, Menglong Zhu, Yuan Li, Yang Zhao, and Larry~S. Davis.
\newblock Automatic spatially-aware fashion concept discovery.
\newblock In \emph{{IEEE} International Conference on Computer Vision, {ICCV} 2017}, pages 1472--1480, Venice, Italy, 2017. {IEEE} Computer Society.

\bibitem[Hasibi et~al.(2017)Hasibi, Nikolaev, Xiong, Balog, Bratsberg, Kotov, and Callan]{DBLP:conf/sigir/HasibiNXBBKC17}
Faegheh Hasibi, Fedor Nikolaev, Chenyan Xiong, Krisztian Balog, Svein~Erik Bratsberg, Alexander Kotov, and Jamie Callan.
\newblock Dbpedia-entity v2: A test collection for entity search.
\newblock In \emph{Proceedings of the 40th International ACM SIGIR Conference on Research and Development in Information Retrieval}, page 1265–1268, New York, NY, USA, 2017. Association for Computing Machinery.

\bibitem[Hoogeveen et~al.(2015)Hoogeveen, Verspoor, and Baldwin]{DBLP:conf/adcs/HoogeveenVB15}
Doris Hoogeveen, Karin~M. Verspoor, and Timothy Baldwin.
\newblock Cqadupstack: A benchmark data set for community question-answering research.
\newblock In \emph{Proceedings of the 20th Australasian Document Computing Symposium}, New York, NY, USA, 2015. Association for Computing Machinery.

\bibitem[Hu et~al.(2022)Hu, yelong shen, Wallis, Allen-Zhu, Li, Wang, Wang, and Chen]{DBLP:conf/iclr/HuSWALWWC22}
Edward~J Hu, yelong shen, Phillip Wallis, Zeyuan Allen-Zhu, Yuanzhi Li, Shean Wang, Lu Wang, and Weizhu Chen.
\newblock Lo{RA}: Low-rank adaptation of large language models.
\newblock In \emph{International Conference on Learning Representations}, 2022.

\bibitem[Hu et~al.(2023)Hu, Luan, Chen, Khandelwal, Joshi, Lee, Toutanova, and Chang]{DBLP:conf/iccv/HuLCKJLTC23}
Hexiang Hu, Yi Luan, Yang Chen, Urvashi Khandelwal, Mandar Joshi, Kenton Lee, Kristina Toutanova, and Ming{-}Wei Chang.
\newblock Open-domain visual entity recognition: Towards recognizing millions of wikipedia entities.
\newblock In \emph{{IEEE/CVF} International Conference on Computer Vision, {ICCV} 2023}, pages 12031--12041, Paris, France, 2023. {IEEE}.

\bibitem[Izacard et~al.(2022)Izacard, Caron, Hosseini, Riedel, Bojanowski, Joulin, and Grave]{izacard2022unsupervised}
Gautier Izacard, Mathilde Caron, Lucas Hosseini, Sebastian Riedel, Piotr Bojanowski, Armand Joulin, and Edouard Grave.
\newblock Unsupervised dense information retrieval with contrastive learning.
\newblock \emph{Transactions on Machine Learning Research}, 2022.

\bibitem[Jiang et~al.(2024)Jiang, Song, Zhang, Huang, Deng, Sun, Zhang, Wang, and Zhuang]{jiang2024e5v}
Ting Jiang, Minghui Song, Zihan Zhang, Haizhen Huang, Weiwei Deng, Feng Sun, Qi Zhang, Deqing Wang, and Fuzhen Zhuang.
\newblock {E5-V:} universal embeddings with multimodal large language models.
\newblock \emph{CoRR}, abs/2407.12580, 2024.

\bibitem[Jiang et~al.(2025)Jiang, Meng, Yang, Yavuz, Zhou, and Chen]{jiang2024vlm2vec}
Ziyan Jiang, Rui Meng, Xinyi Yang, Semih Yavuz, Yingbo Zhou, and Wenhu Chen.
\newblock {VLM}2vec: Training vision-language models for massive multimodal embedding tasks.
\newblock In \emph{The Thirteenth International Conference on Learning Representations}, 2025.

\bibitem[Joshi et~al.(2017)Joshi, Choi, Weld, and Zettlemoyer]{2017arXivtriviaqa}
Mandar Joshi, Eunsol Choi, Daniel Weld, and Luke Zettlemoyer.
\newblock {T}rivia{QA}: A large scale distantly supervised challenge dataset for reading comprehension.
\newblock In \emph{Proceedings of the 55th Annual Meeting of the Association for Computational Linguistics (Volume 1: Long Papers)}, pages 1601--1611, Vancouver, Canada, 2017. Association for Computational Linguistics.

\bibitem[Karpukhin et~al.(2020)Karpukhin, Oguz, Min, Lewis, Wu, Edunov, Chen, and Yih]{karpukhin-etal-2020-dense}
Vladimir Karpukhin, Barlas Oguz, Sewon Min, Patrick Lewis, Ledell Wu, Sergey Edunov, Danqi Chen, and Wen-tau Yih.
\newblock Dense passage retrieval for open-domain question answering.
\newblock In \emph{Proceedings of the 2020 Conference on Empirical Methods in Natural Language Processing (EMNLP)}, pages 6769--6781, Online, 2020. Association for Computational Linguistics.

\bibitem[Kwiatkowski et~al.(2019)Kwiatkowski, Palomaki, Redfield, Collins, Parikh, Alberti, Epstein, Polosukhin, Devlin, Lee, Toutanova, Jones, Kelcey, Chang, Dai, Uszkoreit, Le, and Petrov]{DBLP:journals/tacl/KwiatkowskiPRCP19}
Tom Kwiatkowski, Jennimaria Palomaki, Olivia Redfield, Michael Collins, Ankur Parikh, Chris Alberti, Danielle Epstein, Illia Polosukhin, Jacob Devlin, Kenton Lee, Kristina Toutanova, Llion Jones, Matthew Kelcey, Ming-Wei Chang, Andrew~M. Dai, Jakob Uszkoreit, Quoc Le, and Slav Petrov.
\newblock Natural questions: A benchmark for question answering research.
\newblock \emph{Transactions of the Association for Computational Linguistics}, 7:\penalty0 452--466, 2019.

\bibitem[Lauren{\c{c}}on et~al.(2024)Lauren{\c{c}}on, Marafioti, Sanh, and Tronchon]{docmatix}
Hugo Lauren{\c{c}}on, Andr{\'e}s Marafioti, Victor Sanh, and Leo Tronchon.
\newblock Building and better understanding vision-language models: insights and future directions.
\newblock In \emph{Workshop on Responsibly Building the Next Generation of Multimodal Foundational Models}, 2024.

\bibitem[Lee et~al.(2025)Lee, Roy, Xu, Raiman, Shoeybi, Catanzaro, and Ping]{lee2025nvembed}
Chankyu Lee, Rajarshi Roy, Mengyao Xu, Jonathan Raiman, Mohammad Shoeybi, Bryan Catanzaro, and Wei Ping.
\newblock {NV}-embed: Improved techniques for training {LLM}s as generalist embedding models.
\newblock In \emph{The Thirteenth International Conference on Learning Representations}, 2025.

\bibitem[Li et~al.(2022)Li, Li, Xiong, and Hoi]{DBLP:conf/icml/0001LXH22}
Junnan Li, Dongxu Li, Caiming Xiong, and Steven C.~H. Hoi.
\newblock {BLIP:} bootstrapping language-image pre-training for unified vision-language understanding and generation.
\newblock In \emph{International Conference on Machine Learning, {ICML} 2022}, pages 12888--12900, Baltimore, Maryland, {USA}, 2022. {PMLR}.

\bibitem[Li et~al.(2024)Li, Wang, Xu, Wang, Feng, Kong, and Liu]{DBLP:conf/acl/0039WXWFK024}
Lei Li, Yuqi Wang, Runxin Xu, Peiyi Wang, Xiachong Feng, Lingpeng Kong, and Qi Liu.
\newblock Multimodal {A}r{X}iv: A dataset for improving scientific comprehension of large vision-language models.
\newblock In \emph{Proceedings of the 62nd Annual Meeting of the Association for Computational Linguistics (Volume 1: Long Papers)}, pages 14369--14387, Bangkok, Thailand, 2024. Association for Computational Linguistics.

\bibitem[Li et~al.(2023)Li, Zhang, Zhang, Long, Xie, and Zhang]{li2023generaltextembeddingsmultistage}
Zehan Li, Xin Zhang, Yanzhao Zhang, Dingkun Long, Pengjun Xie, and Meishan Zhang.
\newblock Towards general text embeddings with multi-stage contrastive learning.
\newblock \emph{CoRR}, abs/2308.03281, 2023.

\bibitem[Lin et~al.(2014)Lin, Maire, Belongie, Hays, Perona, Ramanan, Doll{\'{a}}r, and Zitnick]{lin2014microsoft}
Tsung{-}Yi Lin, Michael Maire, Serge~J. Belongie, James Hays, Pietro Perona, Deva Ramanan, Piotr Doll{\'{a}}r, and C.~Lawrence Zitnick.
\newblock Microsoft {COCO:} common objects in context.
\newblock In \emph{13th European Conference on Computer Vision, {ECCV} 2014}, pages 740--755, Zurich, Switzerland, 2014. Springer.

\bibitem[Lin et~al.(2024)Lin, Mei, Chen, and Byrne]{lin-etal-2024-preflmr}
Weizhe Lin, Jingbiao Mei, Jinghong Chen, and Bill Byrne.
\newblock {P}re{FLMR}: Scaling up fine-grained late-interaction multi-modal retrievers.
\newblock In \emph{Proceedings of the 62nd Annual Meeting of the Association for Computational Linguistics (Volume 1: Long Papers)}, pages 5294--5316, Bangkok, Thailand, 2024. Association for Computational Linguistics.

\bibitem[Liu et~al.(2021{\natexlab{a}})Liu, Wang, Wang, and Ordonez]{DBLP:conf/emnlp/LiuWWO21}
Fuxiao Liu, Yinghan Wang, Tianlu Wang, and Vicente Ordonez.
\newblock Visual news: Benchmark and challenges in news image captioning.
\newblock In \emph{Proceedings of the 2021 Conference on Empirical Methods in Natural Language Processing}, pages 6761--6771, Online and Punta Cana, Dominican Republic, 2021{\natexlab{a}}. Association for Computational Linguistics.

\bibitem[Liu et~al.(2023{\natexlab{a}})Liu, Li, Wu, and Lee]{DBLP:conf/nips/LiuLWL23a}
Haotian Liu, Chunyuan Li, Qingyang Wu, and Yong~Jae Lee.
\newblock Visual instruction tuning.
\newblock In \emph{Thirty-seventh Conference on Neural Information Processing Systems}, 2023{\natexlab{a}}.

\bibitem[Liu et~al.(2024)Liu, Li, Li, Li, Zhang, Shen, and Lee]{liu2024llavanext}
Haotian Liu, Chunyuan Li, Yuheng Li, Bo Li, Yuanhan Zhang, Sheng Shen, and Yong~Jae Lee.
\newblock Llava-next: Improved reasoning, ocr, and world knowledge, 2024.

\bibitem[Liu et~al.(2023{\natexlab{b}})Liu, Feng, Fu, Chen, and Wang]{DBLP:conf/emnlp/LiuFFCW23}
Siqi Liu, Weixi Feng, Tsu-Jui Fu, Wenhu Chen, and William Wang.
\newblock {EDIS}: Entity-driven image search over multimodal web content.
\newblock In \emph{Proceedings of the 2023 Conference on Empirical Methods in Natural Language Processing}, pages 4877--4894, Singapore, 2023{\natexlab{b}}. Association for Computational Linguistics.

\bibitem[Liu et~al.(2021{\natexlab{b}})Liu, Opazo, Teney, and Gould]{Liu2021ImageRO}
Zheyuan Liu, Cristian~Rodriguez Opazo, Damien Teney, and Stephen Gould.
\newblock Image retrieval on real-life images with pre-trained vision-and-language models.
\newblock In \emph{2021 {IEEE/CVF} International Conference on Computer Vision, {ICCV} 2021}, pages 2105--2114, Montreal, Canada, 2021{\natexlab{b}}. {IEEE}.

\bibitem[Liu et~al.(2023{\natexlab{c}})Liu, Xiong, Lv, Liu, and Yu]{liu2023universal}
Zhenghao Liu, Chenyan Xiong, Yuanhuiyi Lv, Zhiyuan Liu, and Ge Yu.
\newblock Universal vision-language dense retrieval: Learning a unified representation space for multi-modal retrieval.
\newblock In \emph{The Eleventh International Conference on Learning Representations}, 2023{\natexlab{c}}.

\bibitem[Luo et~al.(2023)Luo, Fang, Gokhale, Yang, and Baral]{luo-etal-2023-end}
Man Luo, Zhiyuan Fang, Tejas Gokhale, Yezhou Yang, and Chitta Baral.
\newblock End-to-end knowledge retrieval with multi-modal queries.
\newblock In \emph{Proceedings of the 61st Annual Meeting of the Association for Computational Linguistics (Volume 1: Long Papers)}, pages 8573--8589, Toronto, Canada, 2023. Association for Computational Linguistics.

\bibitem[Ma et~al.(2024)Ma, Lin, Li, Chen, and Lin]{ma-etal-2024-unifying}
Xueguang Ma, Sheng-Chieh Lin, Minghan Li, Wenhu Chen, and Jimmy Lin.
\newblock Unifying multimodal retrieval via document screenshot embedding.
\newblock In \emph{Proceedings of the 2024 Conference on Empirical Methods in Natural Language Processing}, pages 6492--6505, Miami, Florida, USA, 2024. Association for Computational Linguistics.

\bibitem[Maia et~al.(2018)Maia, Handschuh, Freitas, Davis, McDermott, Zarrouk, and Balahur]{DBLP:conf/www/MaiaHFDMZB18}
Macedo Maia, Siegfried Handschuh, Andr{\'{e}} Freitas, Brian Davis, Ross McDermott, Manel Zarrouk, and Alexandra Balahur.
\newblock Www'18 open challenge: Financial opinion mining and question answering.
\newblock In \emph{Companion of the The Web Conference 2018 on The Web Conference 2018}, pages 1941--1942, Lyon, France, 2018. {ACM}.

\bibitem[Marino et~al.(2019)Marino, Rastegari, Farhadi, and Mottaghi]{DBLP:conf/cvpr/MarinoRFM19}
Kenneth Marino, Mohammad Rastegari, Ali Farhadi, and Roozbeh Mottaghi.
\newblock {OK-VQA:} {A} visual question answering benchmark requiring external knowledge.
\newblock In \emph{{IEEE} Conference on Computer Vision and Pattern Recognition, {CVPR} 2019}, pages 3195--3204, Long Beach, CA, USA, 2019. Computer Vision Foundation / {IEEE}.

\bibitem[Mathew et~al.(2021)Mathew, Karatzas, and Jawahar]{DBLP:conf/wacv/MathewKJ21}
Minesh Mathew, Dimosthenis Karatzas, and C.~V. Jawahar.
\newblock Docvqa: {A} dataset for {VQA} on document images.
\newblock In \emph{{IEEE} Winter Conference on Applications of Computer Vision, {WACV} 2021}, pages 2199--2208, Waikoloa, HI, USA, 2021. {IEEE}.

\bibitem[Mathew et~al.(2022)Mathew, Bagal, Tito, Karatzas, Valveny, and Jawahar]{DBLP:conf/wacv/MathewBTKVJ22}
Minesh Mathew, Viraj Bagal, Rub{\`{e}}n Tito, Dimosthenis Karatzas, Ernest Valveny, and C.~V. Jawahar.
\newblock Infographicvqa.
\newblock In \emph{{IEEE/CVF} Winter Conference on Applications of Computer Vision, {WACV} 2022, Waikoloa, HI, USA, January 3-8, 2022}, pages 2582--2591. {IEEE}, 2022.

\bibitem[Mensink et~al.(2023)Mensink, Uijlings, Castrej{\'{o}}n, Goel, Cadar, Zhou, Sha, Ara{\'{u}}jo, and Ferrari]{DBLP:conf/iccv/MensinkUCGCZSAF23}
Thomas Mensink, Jasper R.~R. Uijlings, Llu{\'{\i}}s Castrej{\'{o}}n, Arushi Goel, Felipe Cadar, Howard Zhou, Fei Sha, Andr{\'{e}} Ara{\'{u}}jo, and Vittorio Ferrari.
\newblock Encyclopedic {VQA:} visual questions about detailed properties of fine-grained categories.
\newblock In \emph{{IEEE/CVF} International Conference on Computer Vision, {ICCV} 2023}, pages 3090--3101, Paris, France, 2023. {IEEE}.

\bibitem[Nguyen et~al.(2016)Nguyen, Rosenberg, Song, Gao, Tiwary, Majumder, and Deng]{DBLP:conf/nips/NguyenRSGTMD16}
Tri Nguyen, Mir Rosenberg, Xia Song, Jianfeng Gao, Saurabh Tiwary, Rangan Majumder, and Li Deng.
\newblock {MS} {MARCO:} {A} human generated machine reading comprehension dataset.
\newblock In \emph{Proceedings of the Workshop on Cognitive Computation: Integrating neural and symbolic approaches 2016}, Barcelona, Spain, 2016. CEUR-WS.org.

\bibitem[OpenAI(2023{\natexlab{a}})]{2023GPT4VisionSC}
OpenAI.
\newblock Gpt-4v(ision) system card, 2023{\natexlab{a}}.

\bibitem[OpenAI(2023{\natexlab{b}})]{openai2024gpt4technicalreport}
OpenAI.
\newblock {GPT-4} technical report.
\newblock \emph{CoRR}, abs/2303.08774, 2023{\natexlab{b}}.

\bibitem[Plummer et~al.(2015)Plummer, Wang, Cervantes, Caicedo, Hockenmaier, and Lazebnik]{DBLP:conf/iccv/PlummerWCCHL15}
Bryan~A. Plummer, Liwei Wang, Chris~M. Cervantes, Juan~C. Caicedo, Julia Hockenmaier, and Svetlana Lazebnik.
\newblock Flickr30k entities: Collecting region-to-phrase correspondences for richer image-to-sentence models.
\newblock In \emph{2015 {IEEE} International Conference on Computer Vision, {ICCV} 2015}, pages 2641--2649, Santiago, Chile, 2015. {IEEE} Computer Society.

\bibitem[Radford et~al.(2021)Radford, Kim, Hallacy, Ramesh, Goh, Agarwal, Sastry, Askell, Mishkin, Clark, Krueger, and Sutskever]{DBLP:conf/icml/RadfordKHRGASAM21}
Alec Radford, Jong~Wook Kim, Chris Hallacy, Aditya Ramesh, Gabriel Goh, Sandhini Agarwal, Girish Sastry, Amanda Askell, Pamela Mishkin, Jack Clark, Gretchen Krueger, and Ilya Sutskever.
\newblock Learning transferable visual models from natural language supervision.
\newblock In \emph{Proceedings of the 38th International Conference on Machine Learning, {ICML} 2021}, pages 8748--8763. {PMLR}, 2021.

\bibitem[Rajpurkar et~al.(2016)Rajpurkar, Zhang, Lopyrev, and Liang]{rajpurkar-etal-2016-squad}
Pranav Rajpurkar, Jian Zhang, Konstantin Lopyrev, and Percy Liang.
\newblock {SQ}u{AD}: 100,000+ questions for machine comprehension of text.
\newblock In \emph{Proceedings of the 2016 Conference on Empirical Methods in Natural Language Processing}, pages 2383--2392, Austin, Texas, 2016. Association for Computational Linguistics.

\bibitem[Robinson et~al.(2021)Robinson, Chuang, Sra, and Jegelka]{Robinson2020ContrastiveLW}
Joshua~David Robinson, Ching-Yao Chuang, Suvrit Sra, and Stefanie Jegelka.
\newblock Contrastive learning with hard negative samples.
\newblock In \emph{International Conference on Learning Representations}, 2021.

\bibitem[Schuhmann et~al.(2022)Schuhmann, Beaumont, Vencu, Gordon, Wightman, Cherti, Coombes, Katta, Mullis, Wortsman, Schramowski, Kundurthy, Crowson, Schmidt, Kaczmarczyk, and Jitsev]{schuhmann2022laion5bopenlargescaledataset}
Christoph Schuhmann, Romain Beaumont, Richard Vencu, Cade~W Gordon, Ross Wightman, Mehdi Cherti, Theo Coombes, Aarush Katta, Clayton Mullis, Mitchell Wortsman, Patrick Schramowski, Srivatsa~R Kundurthy, Katherine Crowson, Ludwig Schmidt, Robert Kaczmarczyk, and Jenia Jitsev.
\newblock {LAION}-5b: An open large-scale dataset for training next generation image-text models.
\newblock In \emph{Thirty-sixth Conference on Neural Information Processing Systems: Datasets and Benchmarks Track}, 2022.

\bibitem[Thakur et~al.(2021)Thakur, Reimers, R{\"u}ckl{\'e}, Srivastava, and Gurevych]{thakur2beir}
Nandan Thakur, Nils Reimers, Andreas R{\"u}ckl{\'e}, Abhishek Srivastava, and Iryna Gurevych.
\newblock {BEIR}: A heterogeneous benchmark for zero-shot evaluation of information retrieval models.
\newblock In \emph{Thirty-fifth Conference on Neural Information Processing Systems Datasets and Benchmarks Track (Round 2)}, 2021.

\bibitem[Thorne et~al.(2018)Thorne, Vlachos, Christodoulopoulos, and Mittal]{DBLP:conf/naacl/ThorneVCM18}
James Thorne, Andreas Vlachos, Christos Christodoulopoulos, and Arpit Mittal.
\newblock {FEVER}: a large-scale dataset for fact extraction and {VER}ification.
\newblock In \emph{Proceedings of the 2018 Conference of the North {A}merican Chapter of the Association for Computational Linguistics: Human Language Technologies, Volume 1 (Long Papers)}, pages 809--819, New Orleans, Louisiana, 2018. Association for Computational Linguistics.

\bibitem[van~den Oord et~al.(2018)van~den Oord, Li, and Vinyals]{oord2019representationlearningcontrastivepredictive}
A{\"{a}}ron van~den Oord, Yazhe Li, and Oriol Vinyals.
\newblock Representation learning with contrastive predictive coding.
\newblock \emph{CoRR}, abs/1807.03748, 2018.

\bibitem[Voorhees et~al.(2021)Voorhees, Alam, Bedrick, Demner-Fushman, Hersh, Lo, Roberts, Soboroff, and Wang]{DBLP:journals/sigir/VoorheesABDHLRS20}
Ellen Voorhees, Tasmeer Alam, Steven Bedrick, Dina Demner-Fushman, William~R. Hersh, Kyle Lo, Kirk Roberts, Ian Soboroff, and Lucy~Lu Wang.
\newblock Trec-covid: constructing a pandemic information retrieval test collection.
\newblock \emph{SIGIR Forum}, 54\penalty0 (1), 2021.

\bibitem[Wachsmuth et~al.(2018)Wachsmuth, Syed, and Stein]{DBLP:conf/acl/WachsmuthSS18}
Henning Wachsmuth, Shahbaz Syed, and Benno Stein.
\newblock Retrieval of the best counterargument without prior topic knowledge.
\newblock In \emph{Proceedings of the 56th Annual Meeting of the Association for Computational Linguistics (Volume 1: Long Papers)}, pages 241--251, Melbourne, Australia, 2018. Association for Computational Linguistics.

\bibitem[Wadden et~al.(2020)Wadden, Lin, Lo, Wang, van Zuylen, Cohan, and Hajishirzi]{DBLP:conf/emnlp/WaddenLLWZCH20}
David Wadden, Shanchuan Lin, Kyle Lo, Lucy~Lu Wang, Madeleine van Zuylen, Arman Cohan, and Hannaneh Hajishirzi.
\newblock Fact or fiction: Verifying scientific claims.
\newblock In \emph{Proceedings of the 2020 Conference on Empirical Methods in Natural Language Processing (EMNLP)}, pages 7534--7550, Online, 2020. Association for Computational Linguistics.

\bibitem[Wang et~al.(2016)Wang, Yin, Wang, Wu, and Wang]{wang2016comprehensivesurveycrossmodalretrieval}
Kaiye Wang, Qiyue Yin, Wei Wang, Shu Wu, and Liang Wang.
\newblock A comprehensive survey on cross-modal retrieval.
\newblock \emph{CoRR}, abs/1607.06215, 2016.

\bibitem[Wang et~al.(2022)Wang, Yang, Huang, Jiao, Yang, Jiang, Majumder, and Wei]{DBLP:journals/corr/abs-2212-03533}
Liang Wang, Nan Yang, Xiaolong Huang, Binxing Jiao, Linjun Yang, Daxin Jiang, Rangan Majumder, and Furu Wei.
\newblock Text embeddings by weakly-supervised contrastive pre-training.
\newblock \emph{CoRR}, abs/2212.03533, 2022.

\bibitem[Wang et~al.(2024{\natexlab{a}})Wang, Yang, Huang, Yang, Majumder, and Wei]{DBLP:conf/acl/WangYHYMW24}
Liang Wang, Nan Yang, Xiaolong Huang, Linjun Yang, Rangan Majumder, and Furu Wei.
\newblock Improving text embeddings with large language models.
\newblock In \emph{Proceedings of the 62nd Annual Meeting of the Association for Computational Linguistics (Volume 1: Long Papers)}, pages 11897--11916, Bangkok, Thailand, 2024{\natexlab{a}}. Association for Computational Linguistics.

\bibitem[Wang et~al.(2023)Wang, Wang, Lin, Bai, Zhou, Zhou, Wang, and Zhou]{wang2023one-peace}
Peng Wang, Shijie Wang, Junyang Lin, Shuai Bai, Xiaohuan Zhou, Jingren Zhou, Xinggang Wang, and Chang Zhou.
\newblock {ONE-PEACE:} exploring one general representation model toward unlimited modalities.
\newblock \emph{CoRR}, abs/2305.11172, 2023.

\bibitem[Wang et~al.(2024{\natexlab{b}})Wang, Bai, Tan, Wang, Fan, Bai, Chen, Liu, Wang, Ge, Fan, Dang, Du, Ren, Men, Liu, Zhou, Zhou, and Lin]{wang2024qwen2vl}
Peng Wang, Shuai Bai, Sinan Tan, Shijie Wang, Zhihao Fan, Jinze Bai, Keqin Chen, Xuejing Liu, Jialin Wang, Wenbin Ge, Yang Fan, Kai Dang, Mengfei Du, Xuancheng Ren, Rui Men, Dayiheng Liu, Chang Zhou, Jingren Zhou, and Junyang Lin.
\newblock Qwen2-vl: Enhancing vision-language model's perception of the world at any resolution.
\newblock \emph{CoRR}, abs/2409.12191, 2024{\natexlab{b}}.

\bibitem[Wei et~al.(2024)Wei, Chen, Chen, Hu, Zhang, Fu, Ritter, and Chen]{wei2023uniir}
Cong Wei, Yang Chen, Haonan Chen, Hexiang Hu, Ge Zhang, Jie Fu, Alan Ritter, and Wenhu Chen.
\newblock Uniir: Training and benchmarking universal multimodal information retrievers.
\newblock In \emph{18th European Conference on Computer Vision}, page 387–404, Milan, Italy, 2024. Springer-Verlag.

\bibitem[Wu et~al.(2021)Wu, Gao, Guo, Al{-}Halah, Rennie, Grauman, and Feris]{DBLP:conf/cvpr/WuGGARGF21}
Hui Wu, Yupeng Gao, Xiaoxiao Guo, Ziad Al{-}Halah, Steven Rennie, Kristen Grauman, and Rog{\'{e}}rio Feris.
\newblock Fashion {IQ:} {A} new dataset towards retrieving images by natural language feedback.
\newblock In \emph{{IEEE} Conference on Computer Vision and Pattern Recognition, {CVPR} 2021}, pages 11307--11317. Computer Vision Foundation / {IEEE}, 2021.

\bibitem[Xiao et~al.(2024)Xiao, Liu, Zhang, Muennighoff, Lian, and Nie]{DBLP:conf/sigir/XiaoLZMLN24}
Shitao Xiao, Zheng Liu, Peitian Zhang, Niklas Muennighoff, Defu Lian, and Jian-Yun Nie.
\newblock C-pack: Packed resources for general chinese embeddings.
\newblock In \emph{Proceedings of the 47th International ACM SIGIR Conference on Research and Development in Information Retrieval}, page 641–649, New York, NY, USA, 2024. Association for Computing Machinery.

\bibitem[Xiong et~al.(2021)Xiong, Xiong, Li, Tang, Liu, Bennett, Ahmed, and Overwijk]{DBLP:conf/iclr/XiongXLTLBAO21}
Lee Xiong, Chenyan Xiong, Ye Li, Kwok-Fung Tang, Jialin Liu, Paul~N. Bennett, Junaid Ahmed, and Arnold Overwijk.
\newblock Approximate nearest neighbor negative contrastive learning for dense text retrieval.
\newblock In \emph{9th International Conference on Learning Representations}, 2021.

\bibitem[Yang et~al.(2018)Yang, Qi, Zhang, Bengio, Cohen, Salakhutdinov, and Manning]{DBLP:conf/emnlp/Yang0ZBCSM18}
Zhilin Yang, Peng Qi, Saizheng Zhang, Yoshua Bengio, William Cohen, Ruslan Salakhutdinov, and Christopher~D. Manning.
\newblock {H}otpot{QA}: A dataset for diverse, explainable multi-hop question answering.
\newblock In \emph{Proceedings of the 2018 Conference on Empirical Methods in Natural Language Processing}, pages 2369--2380, Brussels, Belgium, 2018. Association for Computational Linguistics.

\bibitem[Yao et~al.(2024)Yao, Yu, Zhang, Wang, Cui, Zhu, Cai, Li, Zhao, He, Chen, Zhou, Zou, Zhang, Hu, Zheng, Zhou, Cai, Han, Zeng, Li, Liu, and Sun]{yao2024minicpm}
Yuan Yao, Tianyu Yu, Ao Zhang, Chongyi Wang, Junbo Cui, Hongji Zhu, Tianchi Cai, Haoyu Li, Weilin Zhao, Zhihui He, Qianyu Chen, Huarong Zhou, Zhensheng Zou, Haoye Zhang, Shengding Hu, Zhi Zheng, Jie Zhou, Jie Cai, Xu Han, Guoyang Zeng, Dahai Li, Zhiyuan Liu, and Maosong Sun.
\newblock Minicpm-v: {A} {GPT-4V} level {MLLM} on your phone.
\newblock \emph{CoRR}, abs/2408.01800, 2024.

\bibitem[Yin et~al.(2024)Yin, Fu, Zhao, Li, Sun, Xu, and Chen]{yin2024surveymultimodallargelanguage}
Shukang Yin, Chaoyou Fu, Sirui Zhao, Ke Li, Xing Sun, Tong Xu, and Enhong Chen.
\newblock A survey on multimodal large language models.
\newblock \emph{National Science Review}, 11\penalty0 (12), 2024.

\bibitem[Zhao et~al.(2024)Zhao, Liu, Ren, and Wen]{Zhao2022DenseTR}
Wayne~Xin Zhao, Jing Liu, Ruiyang Ren, and Ji{-}Rong Wen.
\newblock Dense text retrieval based on pretrained language models: {A} survey.
\newblock \emph{{ACM} Trans. Inf. Syst.}, 42\penalty0 (4):\penalty0 89:1--89:60, 2024.

\bibitem[Zhou et~al.(2024{\natexlab{a}})Zhou, Liu, Xiao, Zhao, and Xiong]{zhou-etal-2024-vista}
Junjie Zhou, Zheng Liu, Shitao Xiao, Bo Zhao, and Yongping Xiong.
\newblock {VISTA}: Visualized text embedding for universal multi-modal retrieval.
\newblock In \emph{Proceedings of the 62nd Annual Meeting of the Association for Computational Linguistics (Volume 1: Long Papers)}, pages 3185--3200, Bangkok, Thailand, 2024{\natexlab{a}}. Association for Computational Linguistics.

\bibitem[Zhou et~al.(2024{\natexlab{b}})Zhou, Mei, Li, Liu, Xiong, Liu, Gu, and Yu]{zhou-etal-2024-marvel}
Tianshuo Zhou, Sen Mei, Xinze Li, Zhenghao Liu, Chenyan Xiong, Zhiyuan Liu, Yu Gu, and Ge Yu.
\newblock {MARVEL}: Unlocking the multi-modal capability of dense retrieval via visual module plugin.
\newblock In \emph{Proceedings of the 62nd Annual Meeting of the Association for Computational Linguistics (Volume 1: Long Papers)}, pages 14608--14624, Bangkok, Thailand, 2024{\natexlab{b}}. Association for Computational Linguistics.

\bibitem[Zhu et~al.(2022)Zhu, Lei, Feng, Wang, Zhang, and Chua]{DBLP:conf/mm/ZhuLFWZC22}
Fengbin Zhu, Wenqiang Lei, Fuli Feng, Chao Wang, Haozhou Zhang, and Tat{-}Seng Chua.
\newblock Towards complex document understanding by discrete reasoning.
\newblock In \emph{{MM} '22: The 30th {ACM} International Conference on Multimedia, Lisboa, Portugal, October 10 - 14, 2022}, pages 4857--4866. {ACM}, 2022.

\end{thebibliography}
}


\section*{Appendix}

\section{UMRB Details}

\begin{table*}[!t]
\centering
\scriptsize
\resizebox{\textwidth}{!}{\begin{tabular}{l|ccccccc}
\toprule
\bf Name & \bf Type & \bf Categ. & \bf Eval & \bf Candidates & \bf Eval Query & \bf Eval Candidate & \bf In partial \\
& & & \bf Samples & \bf Nums & \bf avg. chars & \bf avg. chars & \\
\midrule
\midrule
ArguAna & Single-Modal & T$\rightarrow$T & 10,080 & 1,406 & 192.98 & 166.80 &  True \\
Climate-FEVER & Single-Modal & T$\rightarrow$T & 1,535 & 5,416,593 & 20.13 &  84.76 & False \\
CQADupStack & Single-Modal & T$\rightarrow$T & 13,145 & 457,199 & 8.59 & 129.09 &   False \\
DBPedia & Single-Modal & T$\rightarrow$T & 400 & 4,635,922 & 5.39 & 49.68 &  False \\
FEVER & Single-Modal & T$\rightarrow$T & 6,666 & 5,416,568 & 8.13 & 84.76 &  False \\
FiQA2018 & Single-Modal & T$\rightarrow$T & 648 & 57,638 & 10.77 & 132.32 & False \\
HotpotQA & Single-Modal & T$\rightarrow$T & 7,405 & 5,233,329 & 17.61 & 46.30 &  False \\
MSMARCO & Single-Modal & T$\rightarrow$T & 6,980 & 8,841,823 & 5.96 &  55.98 &  False \\
NFCorpus & Single-Modal & T$\rightarrow$T & 323 & 3,633 & 3.30 & 232.26 & True \\
NQ & Single-Modal & T$\rightarrow$T & 3,452 & 2,681,468 & 9.16 & 78.88 &   False \\
Quora & Single-Modal & T$\rightarrow$T & 10,000 & 522,931 & 9.53 & 11.44 & True \\
SCIDOCS & Single-Modal & T$\rightarrow$T & 1,000 & 25,657 & 9.38 & 176.19 &  True \\
SciFact & Single-Modal & T$\rightarrow$T & 300 & 5,183 & 12.37 & 213.63 &  False \\
Touche2020 & Single-Modal & T$\rightarrow$T & 49 & 382,545 & 6.55 & 292.37 & False  \\
TRECCOVID & Single-Modal & T$\rightarrow$T & 50 & 171,332 & 10.60 & 160.77 &  True \\
WebQA & Single-Modal & T$\rightarrow$T & 2,455  & 544,457 & 18.58 & 37.67  & False \\
Nights & Single-Modal & I$\rightarrow$I  & 2,120 & 40,038 & - & -& True  \\
\midrule
VisualNews & Cross-Modal & T$\rightarrow$I  & 19,995 & 542,246 & 18.78 & - & False  \\
Fashion200k & Cross-Modal & T$\rightarrow$I  & 1,719 & 201,824 & 4.89 & - & False \\
MSCOCO & Cross-Modal & T$\rightarrow$I  & 24,809 & 5,000 & 10.43 & - & True \\
Flickr30k & Cross-Modal & T$\rightarrow$I  & 5,000 & 1,000 & 12.33 & - & True \\
TAT-DQA & Cross-Modal & T$\rightarrow$VD  & 1,646 & 277 & 12.44 & -  & False \\
ArxivQA & Cross-Modal & T$\rightarrow$VD  & 500 & 500 & 17.12 & - & False \\
DocVQA & Cross-Modal & T$\rightarrow$VD  & 451 & 500 & 8.23 & -  & True \\
InfoVQA & Cross-Modal & T$\rightarrow$VD  & 494 & 500 & 11.29 & - & False \\
Shift Project & Cross-Modal & T$\rightarrow$VD  & 100 & 1,000 & 16.01 & - & True \\
Artificial Intelligence & Cross-Modal & T$\rightarrow$VD  & 100 & 968 & 12.3 & - & False \\
Government Reports & Cross-Modal & T$\rightarrow$VD  & 100 & 972 & 12.62 & - & False \\
Healthcare Industry & Cross-Modal & T$\rightarrow$VD  & 100 & 965 & 12.56 & - & False \\
Energy & Cross-Modal & T$\rightarrow$VD  & 100 & 977 & 13.49 & - & False \\
TabFQuad & Cross-Modal & T$\rightarrow$VD  & 280 & 70 & 16.49 & - & False \\
VisualNews & Cross-Modal & I$\rightarrow$T  & 20,000 & 537,568 & - & 18.53 & False \\
Fashion200k & Cross-Modal & I$\rightarrow$T  & 4,889 & 61,707 & - & 4.95  & False \\
MSCOCO & Cross-Modal & I$\rightarrow$T  & 5,000 & 24,809 & - & 10.43  & True \\
Flickr30k & Cross-Modal & I$\rightarrow$T  & 1,000 & 5,000 & - & 12.33  & True \\
\midrule

WebQA & Fused-Modal & T$\rightarrow$IT & 2,511 & 403,196 & 16.43 & 12.83 & False \\
EDIS & Fused-Modal & T$\rightarrow$IT & 3,241 & 1,047,067 & 20.07 & 15.53 & False \\
OVEN & Fused-Modal & IT$\rightarrow$T &  50,004 & 676,667 & 6.52 & 82.13  & False \\
INFOSEEK & Fused-Modal & IT$\rightarrow$T & 11,323 & 611,651 & 8.76 & 91.49 & False \\
ReMuQ & Fused-Modal & IT$\rightarrow$T & 3,609 & 138,794 & 13.82 & 34.26  & True \\
OKVQA & Fused-Modal & IT$\rightarrow$T & 5,046 & 114,516 & 8.09 & 102.55 & True  \\
LLaVA & Fused-Modal & IT$\rightarrow$T & 5,120 & 5,994 & 10.70 & 90.65 & True \\

FashionIQ & Fused-Modal & IT$\rightarrow$I & 6,003 & 74,381 & 11.70 & - & True \\
CIRR & Fused-Modal & IT$\rightarrow$I & 4,170 & 21,551 & 11.01 & -  & True \\

OVEN & Fused-Modal & IT$\rightarrow$IT & 14,741 & 335,135 & 5.91 & 94.76 & True \\
EVQA & Fused-Modal & IT$\rightarrow$IT & 3,743 & 68,313 & 9.38 & 211.12 & False \\
INFOSEEK & Fused-Modal & IT$\rightarrow$IT & 17,593 & 481,782 & 7.94 & 96.00 & False \\
\bottomrule
\end{tabular}}
\caption{Tasks in UMRB. We counted the number of datasets under each task type and the number of evaluation instances in the dataset, the size of the candidate set, and the average length of the text.}
\label{tab:umr_tasks}
\end{table*}

Table \ref{tab:umr_tasks} summarizes all UMRB tasks along with their statistics. Table \ref{tab:data examples} provides examples of different task types. Below is a brief description of each dataset included in the UMRB.
\subsection{Single-Modal Tasks}
\paragraph{WebQA \cite{DBLP:conf/cvpr/ChangCNGSB22}} This dataset is derived from Wikipedia. In the T$\rightarrow$T setup, both the query and candidate are text. The objective is to find a Wikipedia paragraph that answers the question. We have used 2,455 samples as the test set.
\paragraph{Nights \cite{fu2023dreamsim}} This dataset contains human judgments on the similarity of various image pairs, where both the query and candidate are images. The task is to identify an image that resembles the provided query image. We included 2,120 samples in our UMRB.

\paragraph{ArguAna, ClimateFEVER, CQADupstack, DBPedia, FEVER, FiQA2018, HotpotQA, MSMARCO, NFCorpus, NQ, Quora, SCIDOCS, SciFact, Touche2020 and TRECCOVID} For these datasets, we use the processed versions from BEIR \cite{thakur2beir}.

\subsection{Cross-Modal Tasks}
\paragraph{VisualNews \cite{DBLP:conf/emnlp/LiuWWO21}} This dataset focuses on the news domain and consists of pairs of news headlines and associated images. In UMRB, this dataset can be transformed into two tasks: retrieving the corresponding image based on the news headline (T$\rightarrow$I) and retrieving the corresponding news headline based on the image (I$\rightarrow$T). We utilized 19,995 and 20,000 samples to construct the test set.
\paragraph{Fashion200k \cite{DBLP:conf/iccv/HanWHZZLZD17}} This dataset includes pairs of images and product descriptions. In total, we have 1,719 instances for the task T$\rightarrow$I and 4,889 instances for the task I$\rightarrow$T for evaluation. 
\paragraph{MSCOCO \cite{lin2014microsoft}} This dataset is a well-known image caption dataset. Similar to VisualNews, it is converted into two tasks: ``I$\rightarrow$T'', which retrieves the caption given an image and ``T$\rightarrow$I'', which retrieves the image given a caption.
\paragraph{Flickr30k\cite{DBLP:conf/iccv/PlummerWCCHL15}} This dataset consists of images paired with detailed textual descriptions. We have a total of 1,000 instances for the I$\rightarrow$T task and 5,000 instances for the T$\rightarrow$I task available for evaluation.
\paragraph{TAT-DQA, ArxivQA, DocVQA, InfoVQA, Shift Project, Artificial Intelligence, Government Reports, Healthcare Industry, Energy, TabFQuad} These datasets constitute the retrieval task of T$\rightarrow$VD. Their queries are standard questions, and the candidates are document screenshots. For these datasets, we used the processed versions from ViDoRe \cite{faysse2024colpali}.

\subsection{Fused-Modal Tasks}
\paragraph{WebQA \cite{DBLP:conf/cvpr/ChangCNGSB22}} Similar to WebQA in the Single-Modal setting, this dataset is also derived from Wikipedia, but in the T$\rightarrow$IT setup, the candidates consist of images and text. The task is to find a Wikipedia paragraph with accompanying text and images to answer a specific question. There are 2,511 samples in the evaluation set.
\paragraph{EDIS \cite{DBLP:conf/emnlp/LiuFFCW23}} This dataset involves the cross-modal image search within the news domain. The queries are texts containing entities and events, with candidates consisting of news images and their accompanying headlines. The task requires the model to comprehend both entities and events from the text queries and retrieve the corresponding image and headline.
\paragraph{OVEN \cite{DBLP:conf/iccv/HuLCKJLTC23}} The dataset is sourced from Wikipedia, where a query consists of an image and a question related to the image. The candidates are the Wikipedia title along with the first 100 tokens of its summary. If the associated Wikipedia content includes images, it constitutes an IT$\rightarrow$IT task; otherwise, it forms an IT$\rightarrow$T task. In the evaluation, we have 14,741 samples for the IT$\rightarrow$IT task and 50,004 samples for the IT$\rightarrow$T task.
\paragraph{INFOSEEK \cite{DBLP:conf/emnlp/ChenHLSCRC23}} This dataset is similar to OVEN, with queries consisting of images alongside text questions. The candidates are Wikipedia snippets of 100 tokens containing the exact answers. This dataset also encompasses two tasks: for the IT$\rightarrow$IT and IT$\rightarrow$T tasks, we used 17,593 and 11,323 samples, respectively.
\paragraph{ReMuQ \cite{luo-etal-2023-end}} The dataset is augmented from the WebQA questions by adding images to create new multimodal queries along with a large text corpus. For evaluation, we used 3,609 instances from this dataset.
\paragraph{OKVQA \cite{DBLP:conf/cvpr/MarinoRFM19}} This dataset includes visual questions that require external knowledge to answer. It is structured as an IT$\rightarrow$T retrieval task, where queries consist of visual questions containing images and text, with candidates being external knowledge sources that can assist in answering the questions.
\paragraph{LLaVA \cite{lin-etal-2024-preflmr}} This dataset contains high-quality conversations about an image generated by GPT-3.5, involving exchanges between a human and an AI assistant. The queries comprise questions and instructions sent by humans to the AI assistant, which include both images and text, while the candidates are the AI assistant's replies. We utilized 5,120 samples from this dataset in the UMRB evaluation.
\paragraph{FashionIQ \cite{DBLP:conf/cvpr/WuGGARGF21}} This dataset features images of fashion products along with crowd-sourced descriptions that highlight the differences between these products. Each query consists of an image and a modification sentence that describes changes to the given image, with the retrieval target being the specified image. In the UMRB evaluation, we used 6,003 samples from this dataset.
\paragraph{CIRR \cite{Liu2021ImageRO}} Similar to FashionIQ, CIRR can also be used for composed image retrieval. It involves pairs of real-life reference and target images in each test case, along with a modification sentence detailing the differences between the two images. For the UMRB evaluation, we utilized 4,170 samples from this dataset.
\paragraph{EVQA \cite{DBLP:conf/iccv/MensinkUCGCZSAF23}} This dataset is akin to INFOSEEK, with the key distinction being that the retrieval target of EVQA is a complete Wikipedia paragraph with a maximum length of several thousand tokens. We used 3,743 samples for evaluation, eliminating multi-hop issues present in the original test set. We selected Wikipedia paragraphs from the original dataset as candidates and supplemented them with images. Images native to each paragraph were included when available; otherwise, the first image from the article was utilized due to its typically representative nature.

\subsection{UMRB-Partial}
The full UMRB dataset consists of 47 subtasks, approximately 200,000 evaluation instances, and 40 million candidates, resulting in a significant overhead when testing the model. During our experiments with the GME-7B model, a full evaluation required approximately 400 A100*80G GPU hours. To facilitate development and verification, we created a smaller benchmark by condensing the complete UMRB, which we refer to as \textbf{UMRB-Partial}. Column 8 of Table \ref{tab:umr_tasks} indicates whether a dataset is included in \textbf{UMRB-Partial}. Testing the GME-7B model on UMRB-Partial reduced the evaluation time from 400 A100*80G GPU hours to 80 A100*80G GPU hours.

\section{Results Details}
In this section, we present the detailed scores achieved by our GME and the baseline models on various tasks. Additionally, we provide results from other benchmarks, including BEIR, M-BEIR, and ViDoRe.

\begin{table*}
\centering
\resizebox{0.87\linewidth}{!}{
\setlength{\tabcolsep}{4pt}
\begin{tabular}{cll *{7}{c}}
\toprule
\rowcolor{gray!20}
Type & Task& Dataset & VISTA & CLIP-SF & One-Peace & DSE & E5-V & GME-2B & GME-7B \\
\midrule

\multirow{17}{*}{\begin{tabular}{c}Single-\\Modal\end{tabular}}
&\multirow{16}{*}{T$\rightarrow$T}
&ArguAna & 63.61 & 52.45 & 32.93 & 53.46 & 54.28 & 63.18 & 72.11 \\
&&Climate-FEVER & 31.17 & 20.00 & 20.27 & 19.79 & 21.64 & 41.08 & 48.36 \\
&&CQADupStack & 42.35 & 30.61 & 41.32 & 36.51 & 41.69 & 39.06 & 42.16  \\
&&DBPedia & 40.77 & 26.37 & 32.43 & 40.75 & 38.78 & 41.00 & 46.30  \\
&&FEVER & 86.29 & 50.58 & 51.91 & 80.12 & 78.99 & 92.06 & 93.81  \\
&&FiQA2018 & 40.65 & 22.14 & 36.79 & 36.2 & 45.41 & 43.8 & 63.23  \\
&&HotpotQA & 72.6 & 41.33 & 46.51 & 70.79 & 60.88 & 65.3 & 68.18 \\
&&MSMARCO & 41.35 & 22.15 & 36.55 & 37.73 & 41.23 & 40.61 & 42.93 \\
&&NFCorpus & 37.39 & 27.05 & 31.6 & 32.82 & 36.97 & 38.84 & 36.95 \\
&&NQ & 54.15 & 25.45 & 42.87 & 52.97 & 51.58 & 54.52 & 56.08  \\
&&Quora & 88.90 & 81.63 & 87.46 & 85.84 & 87.6 & 88.12 & 89.67 \\
&&SCIDOCS & 21.73 & 14.75 & 21.64 & 15.66 & 22.36 & 22.94 & 26.35 \\
&&SciFact & 74.04 & 55.98 & 64.51 & 68.97 & 72.75 & 74.19 & 82.43  \\
&&Touche2020 & 25.7 & 17.47 & 16.90 & 14.50 & 21.61 & 26.57 & 22.55 \\
&&TRECCOVID & 77.90 & 63.61 & 69.28 & 52.98 & 72.85 & 71.73 & 77.49  \\
&&WebQA & 83.80 & 84.44 & 63.67 & 83.95 & 89.94 & 94.34 & 94.34  \\
\cmidrule{2-10}

&I$\rightarrow$I 
&Nights & 24.43 & 31.42 & 31.27 & 27.36 &  27.92 & 30.61 & 30.57 \\
\midrule

\multirow{19}{*}{\begin{tabular}{c}Cross-\\Modal\end{tabular}}
&\multirow{4}{*}{T$\rightarrow$I}
&VisualNews & 5.77 & 42.80 & 48.95 & 14.12 & 29.46 & 39.20 & 46.27 \\
&&Fashion200k & 3.08 & 18.38 & 32.34 & 3.08  & 3.78 & 23.50 & 27.64  \\
&&MSCOCO & 47.97 & 80.75 & 71.45 & 74.62 & 52.38 &  76.22 & 79.77  \\
&&Flickr30k & 74.68 &  94.28 &  92.78 & 94.42 & 77.38 & 94.5 & 97.38 \\
\cmidrule{2-10}

&\multirow{10}{*}{T$\rightarrow$VD}
&TAT-DQA & 2.05 & 5.49 & 14.44 & 49.01 & 9.08 & 57.88 & 64.06 \\
&&ArxivQA & 10.30 & 24.10 & 43.94 & 78.17 & 41.16 & 81.41 & 82.55 \\
&&DocVQA & 8.01 & 11.80 & 23.48 & 45.83 & 24.37 & 46.86 & 49.34 \\
&&InfoVQA & 30.02 & 48.78 & 59.97 & 82.06 & 49.5 & 84.97 &  88.79 \\
&&Shift Project & 3.26 & 6.06 & 17.02 & 69.84 & 13.16 & 77.94 & 83.5  \\
&&Artificial Intelligence & 7.34 & 28.64 & 45.41 & 96.88 & 46.18  & 95.75 & 98.02  \\
&&Government Reports & 6.90 & 34.67 & 55.98 & 92.04 & 53.05 & 92.05 & 94.05 \\
&&Healthcare Industry & 9.39 & 32.64 & 59.55 & 96.35 & 59.61 & 96.08 & 97.29 \\
&&Energy & 11.05 & 27.19 & 53.21 & 92.62 & 56.77 & 89.17 & 93.09 \\
&&TabFQuad & 13.08 & 21.53 & 57.05 & 79.29 & 58.22 & 91.79 & 94.92  \\
\cmidrule{2-10}

&\multirow{4}{*}{I$\rightarrow$T}
&VisualNews & 2.79 & 42.67 &  47.27 &  8.74 &  29.54 & 38.21 & 47.16 \\
&&Fashion200k & 4.72 & 18.10 & 30.89 & 3.91 & 4.62 & 26.61 & 31.05 \\
&&MSCOCO & 48.92 &  91.94 & 85.6 & 82.06 & 86.4 &  85.18 &  85.92 \\
&&Flickr30k & 68.50 &  99.11 & 98.60 &  97.11 &  89.62 & 99.00 & 98.9  \\
\midrule

\multirow{13}{*}{\begin{tabular}{c}Fused-\\Modal\end{tabular}}
&\multirow{2}{*}{T$\rightarrow$IT}
&WebQA & 54.84 & 78.42 & 32.42 & 66.99 & 49.62 & 82.24 & 84.11 \\
&&EDIS & 36.78 & 54.09 & 53.01  &  41.26 & 49.62 & 68.10 & 77.40 \\
\cmidrule{2-10}

&\multirow{5}{*}{IT$\rightarrow$T}
&OVEN & 22.32 & 45.98 & 23.69 & 0.38 &  14.4 & 59.67 & 64.13 \\
&&INFOSEEK & 18.53 & 27.58 & 20.05 &  3.06 & 12.69 & 39.22 & 34.67  \\
&&ReMuQ & 76.20 & 83.71 & 26.41 & 94.60 & 52.15 &  96.73 & 95.48  \\
&&OKVQA & 17.14 & 17.44 & 9.67 & 13.28 & 16.71 & 30.08 & 32.61 \\
&&LLaVA & 72.81 & 91.91 & 51.64 & 53.18 &  77.48 & 98.93 & 98.18 \\
\cmidrule{2-10}

&\multirow{2}{*}{IT$\rightarrow$I}
&FashionIQ & 3.28 & 24.54 & 2.93 & 9.81 & 3.73 & 26.34 & 29.89 \\
&&CIRR & 14.65 & 45.25 & 10.53 & 36.52 & 13.19 & 47.70 &  51.79 \\
\cmidrule{2-10}

&\multirow{3}{*}{IT$\rightarrow$IT}
&OVEN & 27.77 & 68.83 & 30.56 & 0.39 & 54.46 &  78.96 &  83.05 \\
&&EVQA & 28.75 & 40.08 & 16.64 &  15.34 &  26.39 & 77.32 & 79.88  \\
&&INFOSEEK & 22.27 &  49.05 & 23.32 & 5.96 & 39.69 & 41.14 & 31.58 \\ 
\midrule

Avg. &&  & 37.32 & 43.66 & 42.01 & 50.04 & 42.52 & 63.42 & 65.87  \\

\bottomrule
\end{tabular}}
\caption{The detailed results of the baselines and our GME on UMRB. Following previous works~\cite{thakur2beir, wei2023uniir, faysse2024colpali}, we present NDCG@10 scores for T$\rightarrow$T tasks, excluding the WebQA dataset. For T$\rightarrow$VD tasks, we provide NDCG@5 scores. For the Fashion200K, FashionIQ and OKVQA datasets, we report Recall@10 scores, while for all other datasets, we report Recall@5 scores.}
\label{tab:umrb_results}
\end{table*}
\subsection{Detailed Results on UMRB}
Table \ref{tab:umrb_results} presents the detailed evaluation results of the baseline systems alongside our GME on UMRB tasks. First, focusing on the average scores, our smaller model, \ie \texttt{GME-Qwen2-VL-2B}, already outperforms the previous state-of-the-art UMR model (VISTA). The larger model, \ie \texttt{GME-Qwen2-VL-7B}, further enhances this performance. In addition, focusing on specific scores on different datasets, our GME achieves state-of-the-art performance on each dataset except the Nights dataset. VISTA and CLIP-SF scored highly on the Nights dataset, likely due to their use of independent image and text encoders for cross-modal retrieval. In the I$\rightarrow$I task, these models relied solely on the image encoder for encoding without cross-modal alignment, which may explain their superior performance on the Nights dataset.

\begin{table}
\centering
\scriptsize
\resizebox{\columnwidth}{!}{
\setlength{\tabcolsep}{4pt}
\begin{tabular}{cll *{6}{c}}
\toprule
\rowcolor{gray!20}
Type & Task& Dataset & T$\rightarrow$T & I$\rightarrow$I & T$\rightarrow$VD & T$\rightarrow$I & IT$\rightarrow$IT & Mix \\
\midrule

\multirow{6}{*}{\begin{tabular}{c}Single-\\Modal\end{tabular}}
&\multirow{5}{*}{T$\rightarrow$T}
&Arguan & 56.25 & 43.51 & 56.73 & 33.53 & 53.22 & 56.22 \\ 
&&NFCorpus & 35.23 & 28.89 & 33.23 & 33.18 & 30.48 & 35.76  \\
&&Quora & 87.82 & 74.37 & 86.32 & 86.43 & 85.2 & 87.4 \\
&&SCIDOCS & 19.07 & 11.82 & 17.51 & 17.2 & 16.93 & 19.88 \\
&&TRECCOVID & 75.57 & 47.89 & 50.89 & 72.37 & 58.92 & 76.38 \\
\cmidrule{2-9}

&I$\rightarrow$I 
&Nights & 27.97 & 28.11 & 24.9 & 28.53 & 26.04 & 30.85 \\
\midrule

\multirow{6}{*}{\begin{tabular}{c}Cross-\\Modal\end{tabular}}
&\multirow{2}{*}{T$\rightarrow$I}
&MSCOCO & 59.7 & 59.41 & 63.67 & 76.91 & 44.97 & 75.3  \\
&&Flickr30k & 83.92 & 65.52 & 87.32 & 93.18 & 74.52 & 93.06 \\
\cmidrule{2-9}

&\multirow{2}{*}{T$\rightarrow$VD}
&DocVQA & 35.8 & 24.24 & 48.38 & 40.58 & 28.05 & 45.62 \\
&&Shift Project & 57.86 & 45.47 & 77.08 & 50.36 & 53.12 & 74.84 \\
\cmidrule{2-9}

&\multirow{2}{*}{I$\rightarrow$T}
&MSCOCO & 74.72 & 63.82 & 80.46 & 84.64 & 70.48 & 84.24 \\
&&Flickr30k & 94.1 & 82.5 & 96.3 & 97.2 & 90.1 & 97.5 \\
\midrule

\multirow{6}{*}{\begin{tabular}{c}Fused-\\Modal\end{tabular}}
&\multirow{3}{*}{IT$\rightarrow$T}
&LLaVA & 92.75 & 89.05 & 86.02 & 89.24 & 88.73 & 95.02 \\
&&ReMuQ & 89.61 & 85.47 & 76.45 & 85.12 & 86.73 & 89.75 \\
&&OKVQA & 24.55 & 16.6 & 15.78 & 16.92 & 18.57 & 20.23 \\
\cmidrule{2-9}

&\multirow{2}{*}{IT$\rightarrow$I}
&FashionIQ & 5.53 & 4.2 & 5.43 & 8.86 & 11.08 & 11.89 \\
&&CIRR & 17.24 & 15.04 & 15.42 & 17.5 & 25.71 & 29.86 \\
\cmidrule{2-9}

&\multirow{1}{*}{IT$\rightarrow$IT}
&OVEN & 59.81 & 38.42 & 57.31 & 56.69 & 65.08 & 63.04 \\
\midrule

Avg. &&& 55.42 & 45.80 & 54.50 & 54.91 & 51.55 & 60.38 \\

\bottomrule
\end{tabular}}
\caption{Performance of models trained on different data types on UMRB-partial. We present NDCG@10 scores for T$\rightarrow$T tasks. For T$\rightarrow$VD tasks,  we provide NDCG@5 scores. For the FashionIQ dataset, we report Recall@10 scores,  while for all other datasets, we report Recall@5 scores.}
\label{tab:umrb_partial_results}
\end{table}

\subsection{Detailed Results on UMRB-Partial}
Figure \ref{fig:data_mix} of main paper
illustrates our exploration of the training data, as discussed in Section 4.2, with specific results presented in Table \ref{tab:umrb_partial_results}. This table details the scores of our models trained on six data types: T$\rightarrow$T, I$\rightarrow$I, T$\rightarrow$VD, T$\rightarrow$I, IT$\rightarrow$IT, and Mix across various tasks. We find that the model trained on mixed data performs the best.

\begin{table*}
\centering
\setlength{\tabcolsep}{4pt}
\resizebox{\textwidth}{!}{
\begin{tabular}{lc|cccccccccccccccc}
\toprule
BEIR &
Avg. &
\begin{tabular}{c} Argu- \\ Ana \end{tabular} &
\begin{tabular}{c} Cli-\\mate- \\ Fever \end{tabular} &
\begin{tabular}{c} CQA-\\Dup- \\ Stack \end{tabular} &
\begin{tabular}{c} DB- \\ Pedia \end{tabular} &
Fever&
FiQA &
\begin{tabular}{c} Hotpot- \\ QA \end{tabular} &
\begin{tabular}{c} MS \\ MAR-\\CO \end{tabular} & 
\begin{tabular}{c} NF- \\ Corpus \end{tabular} & 
NQ &
Quora &
\begin{tabular}{c} Sci- \\ docs \end{tabular} &
\begin{tabular}{c} Sci- \\ fact \end{tabular} &
\begin{tabular}{c} Touche- \\ 2020 \end{tabular} &
\begin{tabular}{c} Trec- \\ Covid \end{tabular} &
\\ \midrule
\midrule
\multicolumn{18}{c}{\bf Text Embedder} \\
\midrule
gte-Qwen2-7B-instruct & 60.25 & 64.27 & 45.88 & 46.43 & 52.42 & 95.11 & 62.03 & 73.08 & 45.98 & 40.6 & 67 & 90.09 & 28.91 & 79.06 & 30.57 & 82.26 \\
NV-Embed-v1 & 59.36 & 68.2 & 34.72 & 50.51 & 48.29 & 87.77 & 63.1 & 79.92 & 46.49 & 38.04 & 71.22 & 89.21 & 20.19 & 78.43 & 28.38 & 85.88 \\
gte-Qwen2-1.5B-instruct & 58.29 & 69.72 & 42.91 & 44.76 & 48.69 & 91.57 & 54.7 & 68.95 & 43.36 & 39.34 & 64 & 89.64 & 24.98 & 78.44 & 27.89 & 85.38 \\
voyage-large-2-instruct & 58.28 & 64.06 & 32.65 & 46.6 & 46.03 & 91.47 & 59.76 & 70.86 & 40.6 & 40.32 & 65.92 & 87.4 & 24.32 & 79.99 & 39.16 & 85.07 \\
neural-embedding-v1 & 58.12 & 67.21 & 32.3 & 49.11 & 48.05 & 89.46 & 58.94 & 78.87 & 42 & 42.6 & 68.36 & 89.02 & 27.69 & 78.82 & 24.06 & 75.33 \\
GritLM-7B & 57.41 & 63.24 & 30.91 & 49.42 & 46.6 & 82.74 & 59.95 & 79.4 & 41.96 & 40.89 & 70.3 & 89.47 & 24.41 & 79.17 & 27.93 & 74.8 \\
e5-mistral-7b-instruct & 56.89 & 61.88 & 38.35 & 42.97 & 48.89 & 87.84 & 56.59 & 75.72 & 43.06 & 38.62 & 63.53 & 89.61 & 16.3 & 76.41 & 26.39 & 87.25 \\
google-gecko & 55.7 & 62.18 & 33.21 & 48.89 & 47.12 & 86.96 & 59.24 & 71.33 & 32.58 & 40.33 & 61.28 & 88.18 & 20.34 & 75.42 & 25.86 & 82.62 \\
text-embedding-3-large & 55.44 & 58.05 & 30.27 & 47.54 & 44.76 & 87.94 & 55 & 71.58 & 40.24 & 42.07 & 61.27 & 89.05 & 23.11 & 77.77 & 23.35 & 79.56 \\
gte-en-large-v1.5 & 57.91 & 72.11 & 48.36 & 42.16 & 46.3 & 93.81 & 63.23 & 68.18 & 42.93 & 36.95 & 56.08 & 89.67 & 26.35 & 82.43 & 22.55 & 77.49 \\
gte-en-base-v1.5 & 54.09 & 63.49 & 40.36 & 39.52 & 39.9 & 94.81 & 48.65 & 67.75 & 42.62 & 35.88 & 52.96 & 88.42 & 21.92 & 76.77 & 25.22 & 73.13 \\
\midrule 
\multicolumn{18}{c}{\bf Multimodal Embedder} \\
\midrule
VISTA & 53.24 & 63.61 & 31.17 & 42.35 & 40.77 & 86.29 & 40.65 & 72.6 & 41.35 & 37.39 & 54.15 & 88.9 & 21.73 & 74.04 & 25.7 & 77.9  \\
CLIP-SF & 36.77 & 52.45 & 20 & 30.61 & 26.37 & 50.58 & 22.14 & 41.33 & 22.15 & 27.05 & 25.45 & 81.63 & 14.75 & 55.98 & 17.47 & 63.60  \\
One-Peace & 42.19 & 32.93 & 20.27 & 41.32 & 32.43 & 51.91 & 36.79 & 46.51 & 36.55 & 31.6 & 42.87 & 87.46 & 21.64 & 64.51 & 16.9 & 69.28  \\
DSE & 46.60 & 53.46 & 19.79 & 36.51 & 40.75 & 80.12 & 36.2 & 70.79 & 37.73 & 32.82 & 52.97 & 85.84 & 15.66 & 68.97 & 14.50 & 52.98  \\
E5-V & 49.91 & 54.28 & 21.64 & 41.69 & 38.78 & 78.99 & 45.41 & 60.88 & 41.23 & 36.97 & 51.58 & 87.6 & 22.36 & 72.75 & 21.61 & 72.85  \\
\bf GME-Qwen2-VL-2B & 53.31 & 61.52 & 42.3 & 38.13 & 46.31	& 92.6 & 45.3 & 72.93 & 40.88 & 37.2	& 60.01	& 87.24	& 23.17 & 63.82	 & 29.06 & 59.24 \\
\bf GME-Qwen2-VL-7B & 55.68 & 64.60 & 45.38 & 41.66 & 50.78 & 94.27	& 57.14 & 79.21 & 42.38 & 38.40 & 67.74 & 88.05 & 27.38 & 62.31 & 23.26 & 52.6	  \\
\bottomrule
\end{tabular}
}
\caption{
BEIR benchmark~\citep{thakur2beir} nDCG@10 scores. We include top models from MTEB Retrieval English leaderboard.
}
\label{tab:beir_results}
\end{table*}
\subsection{Detailed Results on BEIR}
BEIR is a heterogeneous benchmark containing diverse text IR tasks. We utilize BEIR to compare the performance of our GME with other text embedders on T$\rightarrow$T tasks. Table \ref{tab:beir_results} presents the detailed evaluation nDCG@10 scores for pure text embedders and multimodal embedders on T$\rightarrow$T tasks. Except for our GME, other multimodal embedders do not match the performance of pure text embedders on text retrieval tasks, including those like \texttt{E5-V} that are fine-tuned exclusively on text data.  

Naturally, pure text embedding models of the same model size still outperform multimodal embedding models in pure text retrieval tasks. For example, the score of the \texttt{gte-Qwen2-7B-instruct} model is 60.25, while the \texttt{GME-Qwen2-VL-7B} model, with the same model scale, scores 55.63. Although both models share the same text LLM, incorporating or extending multimodal capabilities leads to additional compromises in pure text performance. Minimizing this kind of loss remains an important research question.

\begin{table*}
\centering
\setlength{\tabcolsep}{4pt}
\resizebox{\textwidth}{!}{
\begin{tabular}{lc|ccccccccccccccccc}
\toprule
\multirow{2}{*}[-1.5ex]{MBEIR} & \multirow{2}{*}[-1.5ex]{Avg.} & \multicolumn{3}{c}{q$_t$$\rightarrow$$c_i$} & $q_t$$\rightarrow$$c_t$ & \multicolumn{2}{c}{$q_t$$\rightarrow$($c_i$,$c_t$)} & \multicolumn{3}{c}{$q_i$$\rightarrow$$c_t$} & $q_i$$\rightarrow$$c_i$ &\multicolumn{2}{c}{($q_i$,$q_t$)$\rightarrow$$c_t$} &\multicolumn{2}{c}{($q_i$,$q_t$)$\rightarrow$$c_i$} &\multicolumn{2}{c}{($q_i$,$q_t$)$\rightarrow$($c_i$,$c_t$)} \\
\cmidrule(lr){3-5}
\cmidrule(lr){6-6}
\cmidrule(lr){7-8}
\cmidrule(lr){9-11}
\cmidrule(lr){12-12}
\cmidrule(lr){13-14}
\cmidrule(lr){15-16}
\cmidrule(lr){15-16}
\cmidrule(lr){17-18}
&&\begin{tabular}{c} Visual-\\News \end{tabular} &
\begin{tabular}{c} MS-\\COCO \end{tabular} &
\begin{tabular}{c} Fashion-\\200K \end{tabular} &
\begin{tabular}{c} Web-\\QA \end{tabular} &
EDIS&
\begin{tabular}{c} Web-\\QA \end{tabular} &
\begin{tabular}{c} Visual-\\News \end{tabular} &
\begin{tabular}{c} MS-\\COCO \end{tabular} &
\begin{tabular}{c} Fashion-\\200K \end{tabular} &
NIGHTS &
OVEN &
\begin{tabular}{c} Info-\\Seek \end{tabular} &
\begin{tabular}{c} Fashion-\\IQ \end{tabular} &
CIRR&
OVEN &
\begin{tabular}{c} Info-\\Seek \end{tabular} &
\\ 
\midrule
\midrule
CLIP & 32.5 & 43.3 & 61.1 & 6.6 & 36.2 & 43.3 & 45.1 & 41.3 & 79.0 & 7.7 & 26.1 & 24.2 & 20.5 & 7.0 & 13.2 & 38.8 & 26.4  \\
SigLIP & 37.2 & 30.1 & 75.7 & 36.5 & 39.8 & 27.0 & 43.5 & 30.8 & 88.2 & 34.2 & 28.9 & 29.7 &  25.1 & 14.4 & 22.7 &  41.7 & 27.4  \\
BLIP &  26.8 & 16.4 & 74.4 & 15.9 & 44.9 & 26.8 & 20.3 & 17.2 & 83.2 & 19.9 & 27.4 & 16.1 & 10.2 & 2.3 & 10.6 & 27.4 & 16.6\\
BLIP2 & 24.8 & 16.7 & 63.8 & 14.0  & 38.6 & 26.9 & 24.5 & 15.0 & 80.0 &  14.2 & 25.4 & 12.2 & 5.5 & 4.4 & 11.8 & 27.3 & 15.8 \\
VISTA & 26.37 & 5.77 & 47.97 & 3.08 & 83.80 & 36.78 & 54.84 & 2.79 & 48.92 & 4.72 & 24.43 & 22.32 & 18.53 & 3.28 & 14.65 & 27.77 & 22.27  \\
CLIP-SF & 50.26 & 42.80 & 80.75 & 18.38 & 84.44 & 54.09 & 78.42 & 42.67 & 91.94 & 18.10 & 31.42 & 45.98 & 27.58 & 24.53 & 45.25 & 68.83 & 49.05  \\
One-Peace & 38.00 & 48.95 & 71.45 & 32.34 & 63.67 & 53.01 & 32.42 & 47.27 & 85.60 & 30.89 & 31.27 & 23.69 & 20.05 & 2.93 & 10.53 & 30.56 & 23.32 \\
DSE & 28.89 & 14.12  & 74.62 & 3.08 & 83.95 & 41.26 & 66.99 & 8.74 & 82.06 & 3.91 & 27.36 & 0.38 & 3.06 & 9.81 & 36.52 & 0.39 & 5.96 \\
E5-V & 35.09 & 29.46 & 52.38 & 3.78 & 89.94 & 49.62 & 49.62 & 29.54 & 86.40 & 4.62 & 27.92 & 14.40 & 12.69 & 3.73 & 13.19 & 54.46 & 39.69 \\
\bf GME-Qwen2-VL-2B & 53.54 & 38.85 & 71.82 & 25.83 & 95.19 & 70.32 & 83.15 & 38.32 & 84.12 & 27.57 & 29.86 & 58.17 & 39.06 & 27.5 & 46.83 & 75.98 & 44.21 \\
\bf GME-Qwen2-VL-7B & 54.50 & 46.54 & 75.14 & 31.82 & 95.85 & 77.29 & 84.59 & 45.54 & 64.90 & 34.20 & 31.89 & 63.41 & 43.14 & 31.43 & 53.69 & 80.30 & 58.80 \\
\bottomrule
\end{tabular}
}
\caption{
Results of M-BEIR benchmark \cite{wei2023uniir}. For the Fashion200K and FashionIQ datasets, we report Recall@10 scores, while for all other datasets, we report Recall@5 scores.
}
\label{tab:mbeir_results}
\end{table*}

\subsection{Detailed Results on M-BEIR}
M-BEIR, a multimodal benchmark for IR, serves as a comprehensive large-scale retrieval benchmark designed to evaluate multimodal retrieval models. As shown in Table \ref{tab:mbeir_results}, we report Recall@10 scores for the Fashion200K and FashionIQ datasets, while Recall@5 scores are provided for all other datasets. In M-BEIR, our GME continues to demonstrate state-of-the-art performance, underscoring the effectiveness of our approach.

\begin{table*}[ht]
\centering
\setlength{\tabcolsep}{4pt}
\resizebox{0.8\textwidth}{!}{
\begin{tabular}{lcccccccccc|cc}

\toprule
 & ArxivQ & DocQ & InfoQ & TabF & TATQ & Shift & AI & Energy & Gov. & Health. & Avg. \\

\midrule
\midrule
BM25\tiny Text + Captioning & 40.1 & 38.4 & 70.0 & 35.4 & 61.5 & 60.9 & 88.0 & 84.7 & 82.7 & 89.2 
& 65.1\\
BGE-M3\tiny Text + Captioning & 35.7 & 32.9 & 71.9 & 69.1 & 43.8 & 73.1 & 88.8 & 83.3 & 80.4 & 91.3 & 67.0\\
\midrule
Jina-CLIP & 25.4 & 11.9 & 35.5 & 20.2 & 3.3 & 3.8 & 15.2 & 19.7 & 21.4 & 20.8 & 17.7 \\
Nomic-vision & 17.1 & 10.7 & 30.1 & 16.3 & 2.7 & 1.1 & 12.9 & 10.9 & 11.4 & 15.7 & 12.9\\
SigLIP \tiny(Vanilla)& 43.2 & 30.3 & 64.1 & 58.1 & 26.2 & 18.7 & 62.5 & 65.7 & 66.1 & 79.1 & 51.4 \\
ColPali & 79.1 & 54.4 & 81.8 & 83.9 & 65.8 & 73.2 & 96.2 & 91.0 & 92.7 & 94.4 & 81.3\\
VISTA & 10.3 & 8.01 & 30.02 & 13.08 & 2.05 & 3.26 & 7.14 & 11.05 & 6.9 & 9.39 & 10.12\\
CLIP-SF & 24.1 & 11.8 & 48.78 & 21.53 & 5.49 & 6.06 & 28.64 & 27.19 & 34.67 & 32.64 & 24.09  \\
One-Peace & 43.94 & 23.48 & 59.97 & 57.05 & 13.44 & 17.02 & 45.41 & 53.21 & 55.98 & 59.5 &42.9  \\
DSE & 78.17 & 45.83 & 82.06 & 79.29 & 49.01 & 69.84 & 96.89 & 92.62 & 92.04 & 96.35 &78.21  \\
E5-V & 41.16 & 24.37 & 49.5 & 58.22 & 9.08 & 13.26 & 46.18 & 57.77 & 53.05 & 59.61 & 41.22  \\
\bf GME-Qwen2-VL-2B & 83.91	& 54.57	& 91.11 & 94.61	 & 71.05 & 94.29 & 99.02 & 93.15 & 97.89 & 98.89 & 87.84 \\
\bf GME-Qwen2-VL-7B &  87.58 & 56.63 & 92.39 & 94.58 & 76.12 & 97.26 & 99.63 & 95.89 & 99.5 & 99.63 & 89.92 \\
\bottomrule

\end{tabular}}

\caption{Comprehensive evaluation of baseline models and our GME on \textit{ViDoRe}. Results are presented using NDCG@5 metrics.} 
\label{table:vidore_results}
\end{table*}
\subsection{Detailed Results on ViDoRe}
ViDoRe represents the Visual Document Retrieval Benchmark, encompassing various page-level screenshot retrieval tasks. This benchmark includes the T$\rightarrow$VD tasks within our UMRB. Table \ref{table:vidore_results} presents the detailed nDCG@5 scores for our GME and other models. Our smaller model, \ie \texttt{GME-Qwen2-VL-2B}, surpasses the previous state-of-the-art model (ColPali), which was exclusively trained on this dataset for this specific task. The larger model, \ie \texttt{GME-Qwen2-VL-7B}, further improves upon this performance.

\section{Experiment Details}
\begin{table}
\centering
\resizebox{1.0\columnwidth}{!}{
\begin{tabular}[b]{lcc}
\toprule
Hyper-param & GME-Qwen2-VL-2B &GME-Qwen2-VL-7B  \\
\midrule 
Number of Params & 2B & 8.2B \\
Number of Layers & 28 & 28 \\
Hidden Size & 1536 & 3584 \\
FFN Inner Size & \multicolumn{2}{c}{3072} \\
Number of Attention Heads & 12 & 28 \\
Vision Depth & \multicolumn{2}{c}{32}  \\
Vision Embed\_dim & \multicolumn{2}{c}{1280}  \\
Vision Patch\_size & \multicolumn{2}{c}{14}  \\
Temperature & \multicolumn{2}{c}{0.03}  \\
Learning Rate Decay & \multicolumn{2}{c}{Linear}  \\
Adam $\epsilon$ & \multicolumn{2}{c}{1e-4}  \\
Adam $\beta_1$  & \multicolumn{2}{c}{0.9}  \\
Adam $\beta_2$  & \multicolumn{2}{c}{0.98} \\
Gradient Clipping & \multicolumn{2}{c}{0.0} \\
Precision &  \multicolumn{2}{c}{PyTorch BF16 AMP}   \\
Max Length & 1800 & 1800  \\
Batch Size & 128 & 32  \\
Warm-up Ratio & \multicolumn{2}{c}{0.06}  \\
\bottomrule
\end{tabular}
}
\caption{
GME training hyper-parameters.
}
\label{tab:hparam_mlm}
\end{table}

\subsection{Training Details}
Our GME models (both 2B and 7B) are initialized using the Qwen2-VL~\citep{wang2024qwen2vl} model series. We employ the \texttt{transformers} library for training in BF16 precision. The training utilizes Low-Rank Adaptation (LoRA) \citep{DBLP:conf/iclr/HuSWALWWC22} with a rank of 8. We apply a decoupled AdamW optimizer with a learning rate and a weight decay of 1e-4. Additional hyperparameters are detailed in Table \ref{tab:hparam_mlm}.

In our contrastive learning approach, we develop dense multimodal representation models (embedders) that utilize the [EOS] hidden state as the embedding of the input. The temperature for contrastive learning is set to 0.03. For each query, we include one positive candidate along with eight hard negative candidates.

\begin{table*}
\centering
\scriptsize
\resizebox{0.95\linewidth}{!}{
\setlength{\tabcolsep}{4pt}
\begin{tabular}{lll}
\toprule
\bf Task& \bf Dataset & \multicolumn{1}{c}{\bf Query Instruction} \\
\midrule

\multirow{16}{*}{T$\rightarrow$T}
&ArguAna &  Given a claim, find documents that refute the claim.  \\ \cmidrule(lr){2-3}
&Climate-FEVER & Given a claim about climate change, retrieve documents that support orrefute the claim. \\ \cmidrule(lr){2-3}
&CQADupStack &  Given a question, retrieve detailed question descriptions from Stackexchange that are duplicates to the given question. \\ \cmidrule(lr){2-3}
&DBPedia & Given a query, retrieve relevant entity descriptions from DBPedia.  \\ \cmidrule(lr){2-3}
&FEVER & Given a claim, retrieve documents that support or refute the claim. \\ \cmidrule(lr){2-3}
&FiQA2018 & Given a financial question, retrieve user replies that best answer the question.  \\ \cmidrule(lr){2-3}
&HotpotQA & Given a multi-hop question, retrieve documents that can help answer the question.   \\ \cmidrule(lr){2-3}
&MSMARCO & Given a web search query, retrieve relevant passages that answer the query. \\ \cmidrule(lr){2-3}
&NFCorpus &  Given a question, retrieve relevant documents that best answer the question. \\ \cmidrule(lr){2-3}
&NQ &  Given a question, retrieve Wikipedia passages that answer the question.  \\ \cmidrule(lr){2-3}
&Quora &  Given a question, retrieve questions that are semantically equivalentto the given question. \\ \cmidrule(lr){2-3}
&SCIDOCS & Given a scientific paper title, retrieve paper abstracts that are cited bythe given paper.  \\ \cmidrule(lr){2-3}
&SciFact & Given a scientific claim, retrieve documents that support or refute theclaim. \\ \cmidrule(lr){2-3}
&Touche2020 & Given a question, retrieve detailed and persuasive arguments that answer the question.  \\ \cmidrule(lr){2-3}
&TRECCOVID &  Given a query on COVID-19, retrieve documents that answer the query. \\ \cmidrule(lr){2-3}
&WebQA & Retrieve passages from Wikipedia that provide answers to the following question.  \\
\midrule

I$\rightarrow$I 
&Nights & Find a day-to-day image that looks similar to the provided image. \\
\midrule

\multirow{4}{*}{T$\rightarrow$I}
&VisualNews & Identify the news-related image in line with the described event.  \\ \cmidrule(lr){2-3}
&Fashion200k & Based on the following fashion description, retrieve the best matching image.  \\ \cmidrule(lr){2-3}
&MSCOCO & Identify the image showcasing the described everyday scene.  \\ \cmidrule(lr){2-3}
&Flickr30k & Find an image that matches the given caption.  \\ 
\midrule

\multirow{10}{*}{T$\rightarrow$VD}
&TAT-DQA & \multirow{10}{*}{Find a screenshot that relevant to the user's question.} \\ 
&ArxivQA \\
&DocVQA  \\ 
&InfoVQA \\ 
&Shift Project\\
&Artificial Intelligence &  \\ 
&Government Reports &   \\ 
&Healthcare Industry &   \\ 
&Energy  \\
&TabFQuad \\
\midrule

\multirow{4}{*}{I$\rightarrow$T}
&VisualNews & Find a caption for the news in the given photo.  \\ \cmidrule(lr){2-3}
&Fashion200k &  Find a product description for the fashion item in the image.\\ \cmidrule(lr){2-3}
&MSCOCO &  Find an image caption describing the following everyday image.  \\ \cmidrule(lr){2-3}
&Flickr30k & Find an image caption describing the following image.  \\
\midrule 

\multirow{2}{*}{T$\rightarrow$IT}
&WebQA & Find a Wikipedia image that answers this question.  \\ \cmidrule(lr){2-3}
&EDIS & Identify the news photo for the given caption.  \\
\midrule

\multirow{5}{*}{IT$\rightarrow$T}
&OVEN & \multirow{2}{*}{Retrieve a Wikipedia paragraph that provides an answer to the given query about the image.}   \\
&INFOSEEK &    \\ \cmidrule(lr){2-3}
&ReMuQ & Retrieve a fact-based paragraph that provides an answer to the given query about the image. \\ \cmidrule(lr){2-3}
&OKVQA & Retrieve documents that provide an answer to the question alongside the image.  \\ \cmidrule(lr){2-3}
&LLaVA & Provide a specific decription of the image along with the following question. \\ 
\midrule

\multirow{2}{*}{IT$\rightarrow$I}
&FashionIQ & Find a fashion image that aligns with the reference image and style note.  \\ \cmidrule(lr){2-3}
&CIRR &  Retrieve a day-to-day image that aligns with the modification instructions of the provided image. \\ 
\midrule

\multirow{3}{*}{IT$\rightarrow$IT}
&OVEN & \multirow{2}{*}{Retrieve a Wikipedia image-description pair that provides evidence for the question of this image.} \\
& INFOSEEK \\ \cmidrule(lr){2-3}
&EVQA & Obtain illustrated documents that correspond to the inquiry alongside the provided image.   \\

\bottomrule
\end{tabular}}
\caption{The instructions for different tasks, we only use the instructions for query encoding.}
\label{tab:task_instruction}
\end{table*}

\subsection{Instructions}
The complete UMRB consists of 47 tasks, each with distinct retrieval candidates and varying domains. Even within the same dataset, retrieval candidates can differ based on task types. For example, the WebQA dataset aims to retrieve textual candidates for T$\rightarrow$T tasks, which is different from retrieving a combination of image and text candidates for T$\rightarrow$IT tasks.

We have designed specific instructions tailored for each task to guide the model in effectively completing the retrieval process. The detailed instructions are provided in Table \ref{tab:task_instruction}.

\section{Fused-Modal Data Synthesis Details}\label{app:data-synthesis}
We utilize doc2query to synthesize data. However, our goal is to generate fused-modal candidate-to-query relevance data rather than single-modality, text-based relevance pairs.
\subsection{Prompts}
\begin{figure}
    \centering
    \includegraphics[width=0.8\columnwidth]{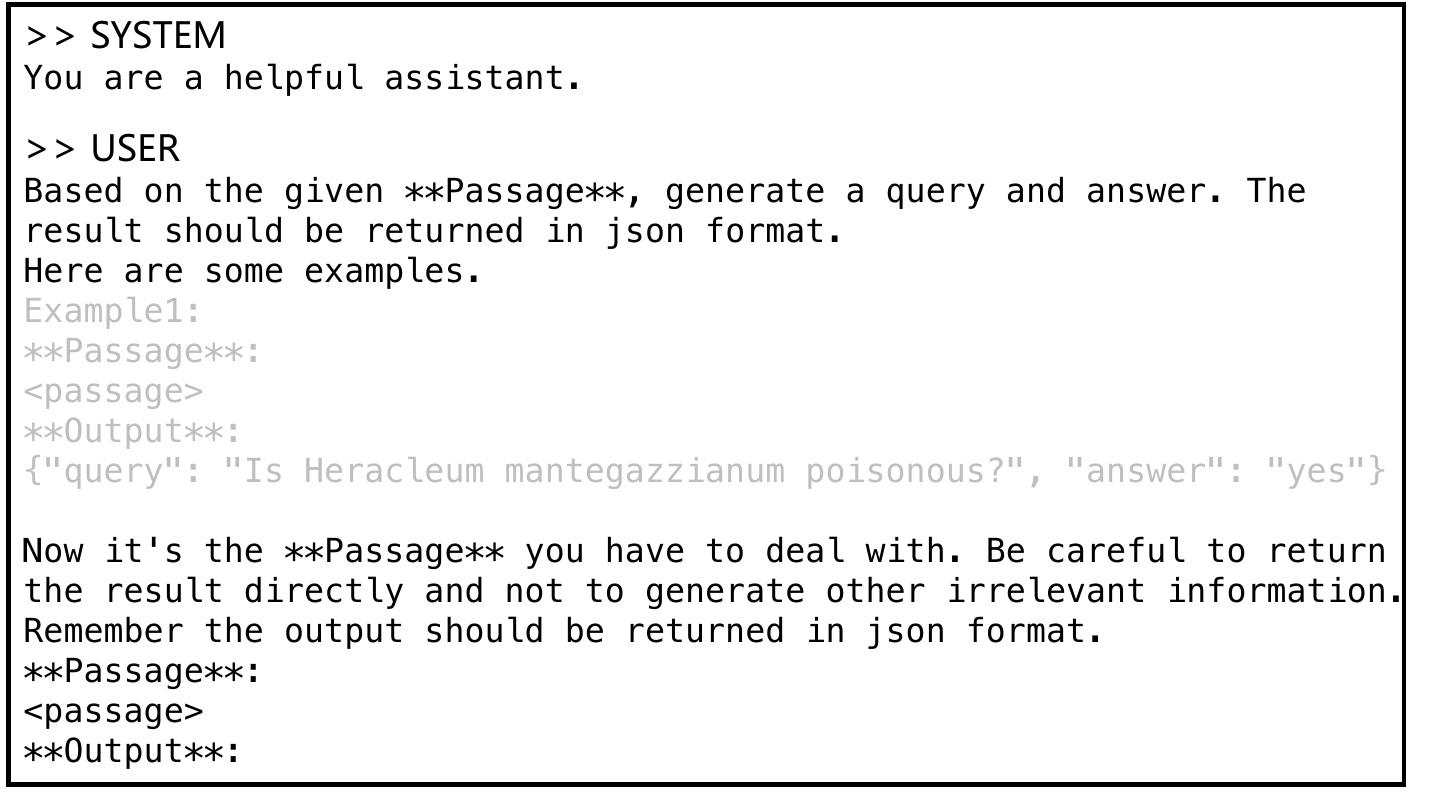}
    \caption{Fused-Modal Data Synthesis Step 1 Prompt.}
    \label{fig:prompt1}
\end{figure}

\begin{figure}
    \centering
    \includegraphics[width=0.8\columnwidth]{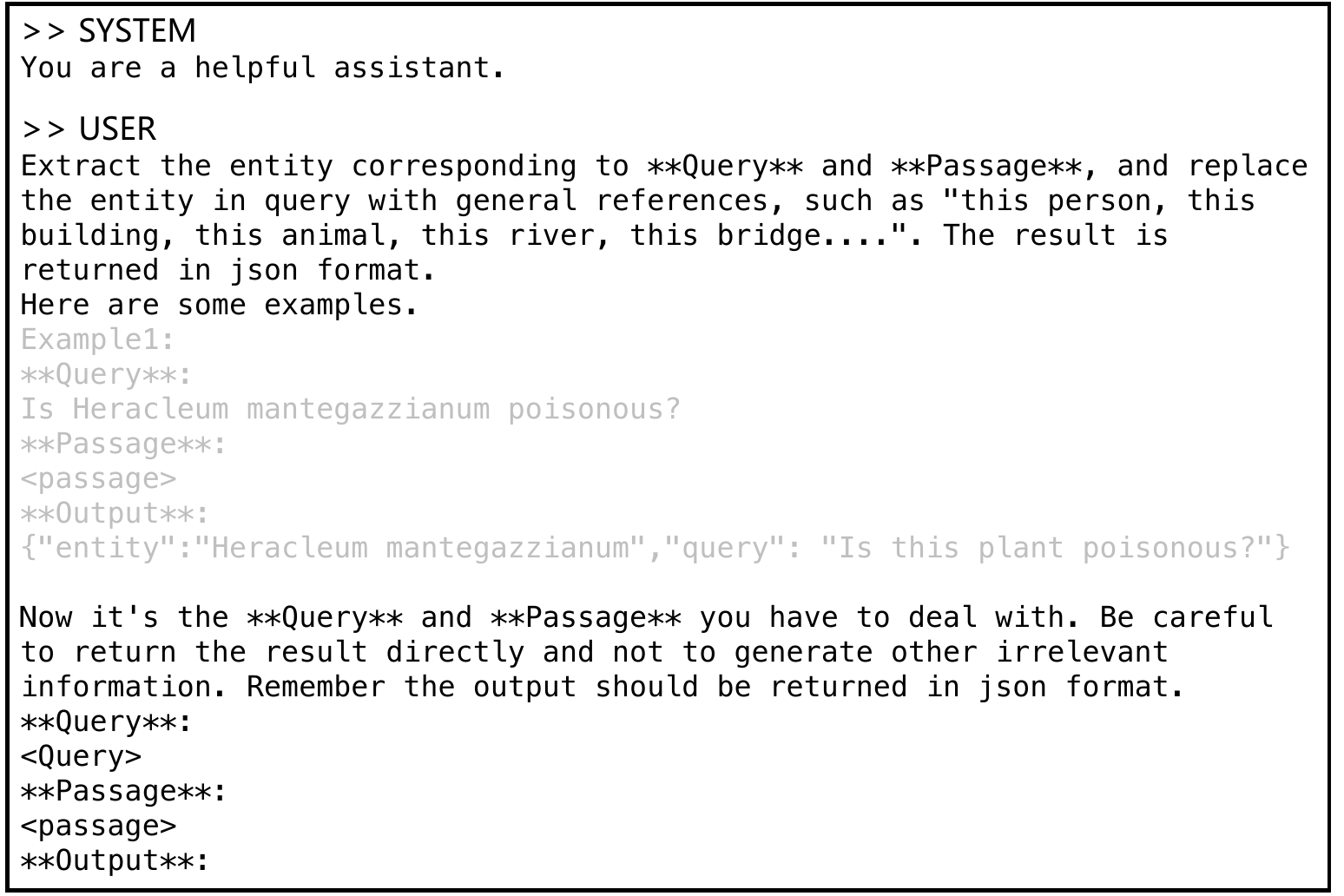}
    \caption{Fused-Modal Data Synthesis Step 2 Prompt.}
    \label{fig:prompt2}
\end{figure}

\begin{figure}
    \centering
    \includegraphics[width=0.8\columnwidth]{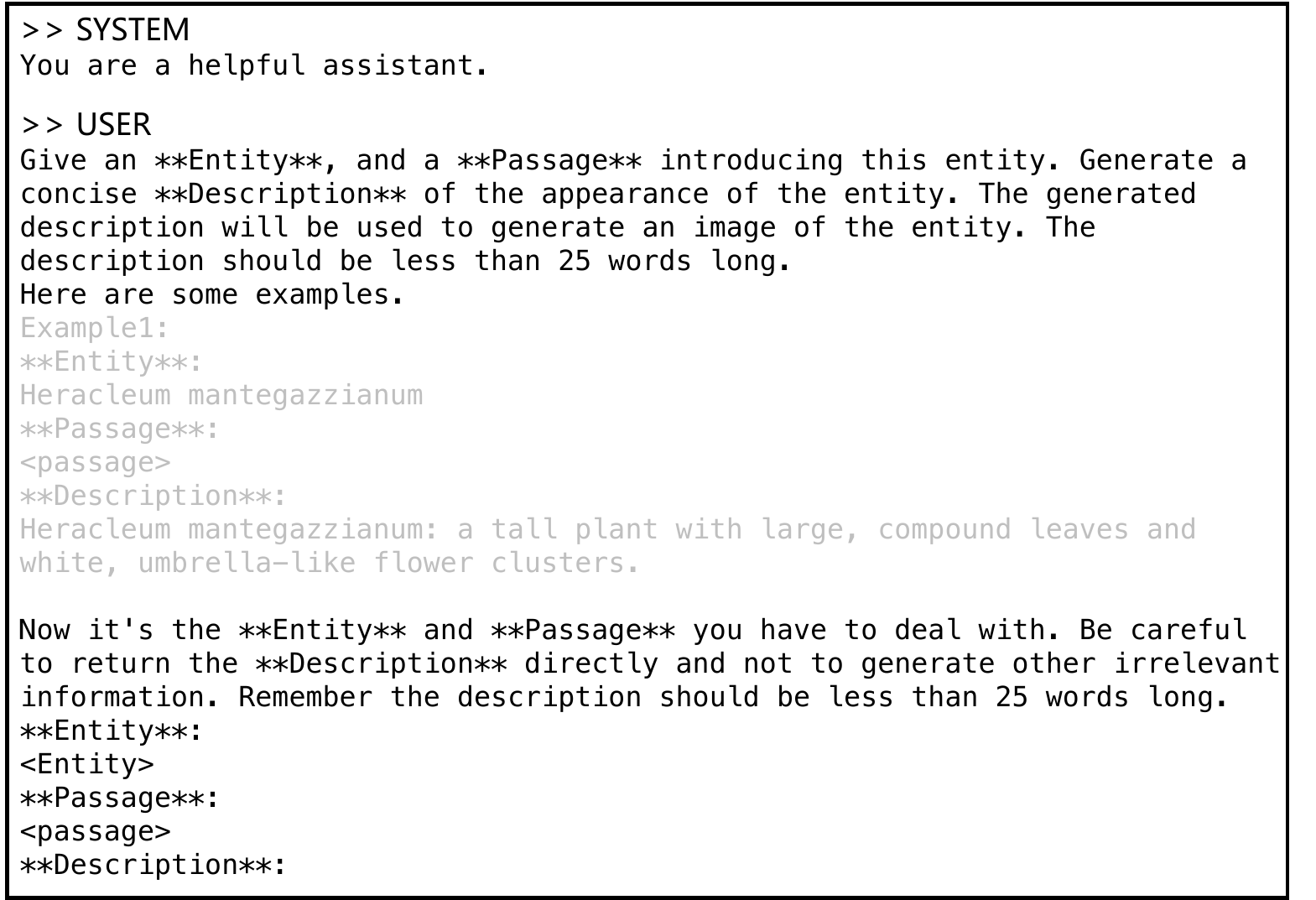}
    \caption{Fused-Modal Data Synthesis Step 3 Prompt.}
    \label{fig:prompt3}
\end{figure}

\paragraph{Step 1:} In the first step of data synthesis, we prompt the large language model (LLM) to generate a natural question and answer based on a selected passage. The specific prompt is illustrated in Figure \ref{fig:prompt1}. This process leverages in-context learning (ICL) to guide the LLM in producing outputs that align with our requirements.
\paragraph{Step 2:} In step 2, we provide the LLM with the passage and the natural question generated in step 1. The LLM is then prompted to extract the main entity from the question and refactor the question accordingly. Figure \ref{fig:prompt2} presents the prompt used in this step. In subsequent steps, the extracted entity will be replaced by the corresponding image, which, when combined with the reconstructed question, will form a fused-modal query.
\paragraph{Step 3:} In step 3, we replace the entity with an image, which can be sourced in two ways. The first method involves prompting the LLM to generate a caption for the entity based on the provided entity and passage, after which the caption is fed into FLUX to generate images. The second method retrieves the entity by utilizing the Google Image Retrieval API. Figure \ref{fig:prompt3} illustrates the caption generation prompt for this step.

\begin{figure}
    \centering
    \includegraphics[width=0.8\columnwidth]{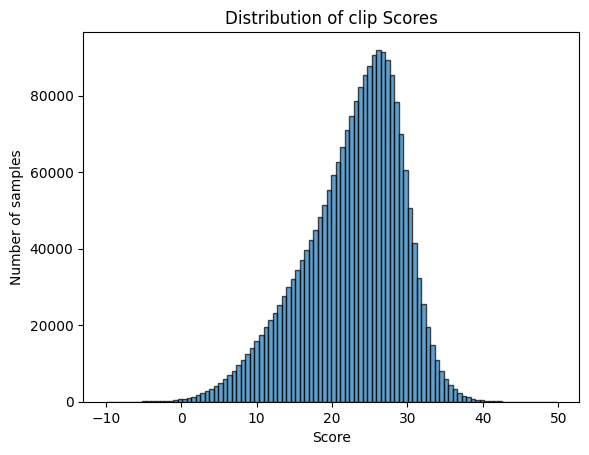}
    \caption{The distribution of relevance scores for all the images searched by Google and captions.}
    \label{fig:clip_score}
\end{figure}

\subsection{Filter}
Two filtering methods are implemented to ensure the quality of the synthesized data. First, a text retrieval model is utilized to evaluate unreconstructed queries and their corresponding passages. We follow the framework of Promptagator \cite{DBLP:conf/iclr/DaiZMLNLBGHC23}; a query is deemed unqualified if the passage that generated it does not appear within the top 20 search results. Second, for images obtained through the Google Image Search API, we employ the CLIP model to assess image-caption relevance. Images with a relevance score below 0.2 are filtered out. 

Why is the threshold score set to 0.2? The relevance scores of all images searched via Google and the corresponding captions we have collected are presented in Figure \ref{fig:clip_score}. We select the median score of 0.2 to ensure image quality while also ensuring that most text queries have sufficient images to pair with.

\begin{table*}
\centering
\scriptsize
\resizebox{\textwidth}{!}{
\setlength{\tabcolsep}{4pt}
\begin{tabular}{lllclc}
\toprule
\bf Type& \bf Task & \multicolumn{1}{c}{\bf Query Text} & \bf Query Image & \multicolumn{1}{c}{\bf Target Text} & \bf Target Image\\
\midrule

\multirow{2}{*}[-8ex]{\begin{tabular}{c}Single-Modal\end{tabular}}
&T$\rightarrow$T & 
\begin{tabular}{l} 
where is whitemarsh\\
island?
\end{tabular}& - & 
\begin{tabular}{l} 
Whitemarsh Island, Georgia Whitemarsh Island, \\
Georgia. Whitemarsh Island (pronounced WIT-marsh) \\
is a census-designated place (CDP) in Chatham County, \\
Georgia, United States. The population was 6,792 at \\
the 2010 census. It is part of the Savannah Metropolitan \\
Statistical Area. The communities of Whitemarsh Island \\
are a relatively affluent suburb of Savannah.
\end{tabular}
& -   \\ \cmidrule(lr){2-6}
&I$\rightarrow$I & \multicolumn{1}{c}{-} &
\begin{tabular}{c} 
\includegraphics[width=15mm,height=15mm]{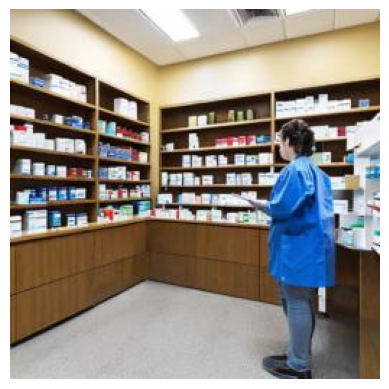}
\end{tabular}& \multicolumn{1}{c}{-} &
\begin{tabular}{c} 
\includegraphics[width=15mm,height=15mm]{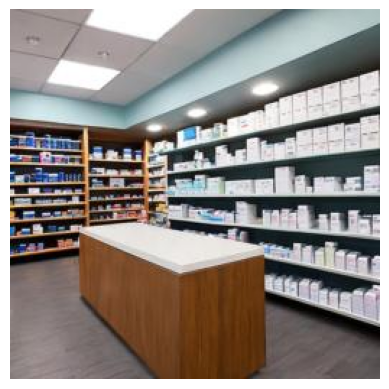}
\end{tabular}
\\ 
\midrule

\multirow{3}{*}[-15ex]{\begin{tabular}{c}Cross-Modal\end{tabular}}
&T$\rightarrow$I &
\begin{tabular}{l} 
Multicolor boutique amy black\\
leather look biker jacket.
\end{tabular} & - & \multicolumn{1}{c}{-} &
\begin{tabular}{c} 
\includegraphics[width=15mm,height=15mm]{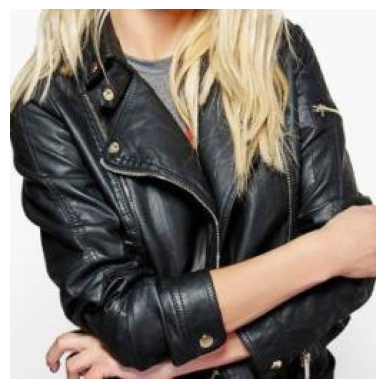}
\end{tabular}
\\ \cmidrule(lr){2-6}
&T$\rightarrow$VD &
\begin{tabular}{l} 
Based on the graph, what is the \\
impact of correcting for fspec not \\
equal to 1 on the surface density trend?
\end{tabular} & - & \multicolumn{1}{c}{-} &
\begin{tabular}{c} 
\includegraphics[width=15mm,height=15mm]{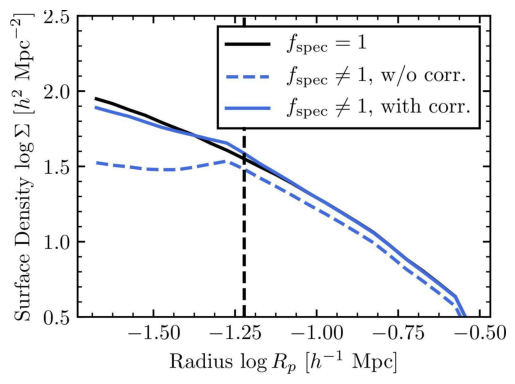}
\end{tabular}
\\ \cmidrule(lr){2-6}
&I$\rightarrow$T &
\begin{tabular}{c} 
\includegraphics[width=15mm,height=15mm]{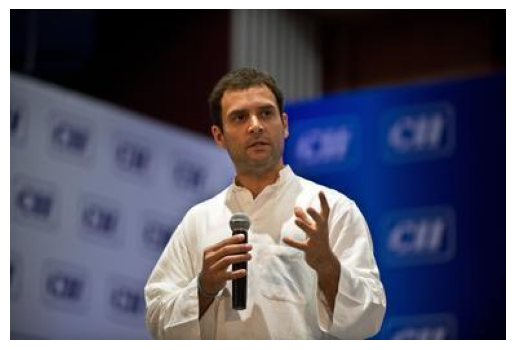}
\end{tabular} & - &
\begin{tabular}{l} 
Indian National Congress Vice President Rahul \\
Gandhi addresses the special plenary session of\\
Confederation of Indian Industr in New Delhi\\
on April 4 2013.
\end{tabular} & -
\\
\midrule

\multirow{4}{*}[-20ex]{\begin{tabular}{c} Fused-Modal\end{tabular}}
&T$\rightarrow$IT &
\begin{tabular}{l} 
Does a Minnetonka Rhododendron flower\\
have petals in a cup shape?
\end{tabular} & - &
\begin{tabular}{l} 
2020-05-08 15 17 05 Minnetonka Rhododendron flower\\
along Tranquility Court in the Franklin Farm section\\
of Oak Hill, Fairfax County, Virginia Minnetonka \\
Rhododendron flower along Tranquility Court in the \\
Franklin Farm section of Oak Hill, Fairfax County, Virginia.
\end{tabular} &
\begin{tabular}{c} 
\includegraphics[width=15mm,height=15mm]{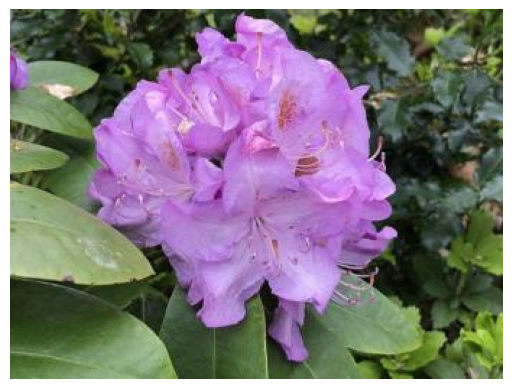}
\end{tabular}

\\ \cmidrule(lr){2-6}
&IT$\rightarrow$T &
\begin{tabular}{l} 
What is this plant named after?
\end{tabular} &
\begin{tabular}{c} 
\includegraphics[width=15mm,height=15mm]{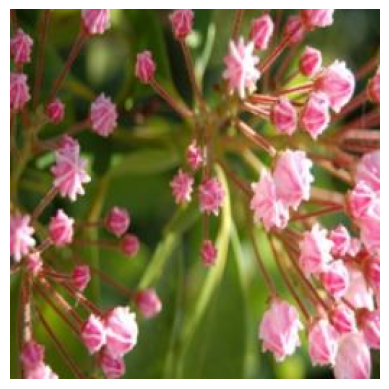}
\end{tabular} &
\begin{tabular}{l} 
Kalmia. Kalmia is a genus of about ten species \\
of evergreen shrubs from 0.2–5 m tall, in the \\
family Ericaceae. They are native to North America\\
... saw it during his travels in Carolina, and \\
after his return to England in.
\end{tabular}& -

\\ \cmidrule(lr){2-6}
&IT$\rightarrow$I &
\begin{tabular}{l} 
Is shiny and silver with shorter sleeves\\
and fit and flare.
\end{tabular} &
\begin{tabular}{c} 
\includegraphics[width=15mm,height=15mm]{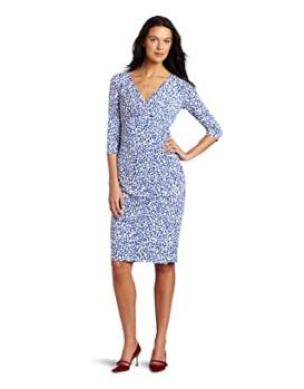}
\end{tabular} & \multicolumn{1}{c}{-} &
\begin{tabular}{c} 
\includegraphics[width=15mm,height=15mm]{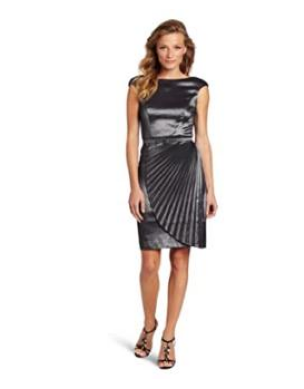}
\end{tabular}
\\ \cmidrule(lr){2-6}
&IT$\rightarrow$IT & 
\begin{tabular}{l} 
Is this plant poisonous?
\end{tabular}&
\begin{tabular}{c} 
\includegraphics[width=15mm,height=15mm]{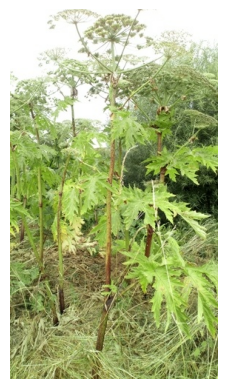}
\end{tabular}&
\begin{tabular}{l} 
Heracleum mantegazzianum, commonly known \\
as giant hogweed, is a monocarpic perennial \\
herbaceous plant in the carrot family Apiaceae\\
...These serious reactions are due to the \\
furanocoumarin derivatives in the leaves, roots, \\
stems, flowers, and seeds of the plant. Consequently, \\
it is considered to be a noxious weed in many jurisdictions.
\end{tabular}&
\begin{tabular}{c} 
\includegraphics[width=15mm,height=15mm]{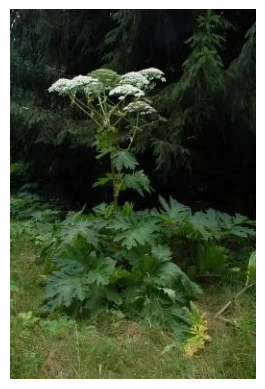}
\end{tabular}
 
\\

\bottomrule
\end{tabular}}
\caption{Data examples in diffierent task type. Due to the limitations of the table, we have cropped the displayed text.}
\label{tab:data examples}
\end{table*}

\subsection{Examples of synthetic data}
Table \ref{tab:synthetic_data_example} illustrates passages from 15 domains and the fused modal queries generated by applying the synthesis flow. ``FLUX image'' refers to images generated by the Vincennes diagram model FLUX.1-dev, whereas ``Google image'' indicates images from Google Image retrieval.
\begin{table*}
\centering
\scriptsize
\resizebox{0.93\linewidth}{!}{
\setlength{\tabcolsep}{4pt}
\begin{tabular}{cclcclc}
\toprule
\bf Domain& \bf Candidate Image & \multicolumn{1}{c}{\bf Candidate Text} & \bf FLUX Image & \bf Google Image & \multicolumn{1}{c}{\bf Query Text} \\
\midrule

animal & 
\begin{tabular}{c} 
\includegraphics[width=15mm,height=15mm]{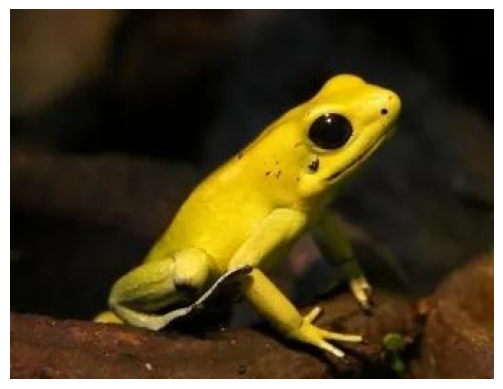}
\end{tabular}&
\begin{tabular}{l} 
The golden poison frog is the most poisonous animal \\
on the planet; these frogs produce deadly alkaloid \\
batrachotoxins in their skin glands as a defense against\\
predators. To become poisoned a predator generally \\
must attempt to consume the frog, ...\\
has modified sodium channels unaffected by batrachotoxin.
\end{tabular}&
\begin{tabular}{c} 
\includegraphics[width=15mm,height=15mm]{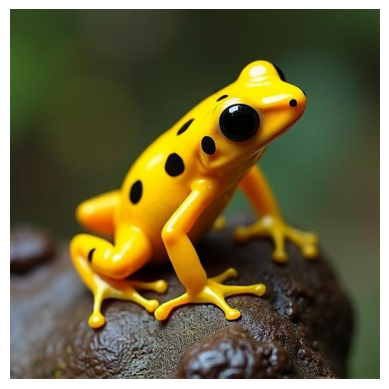}
\end{tabular}&
\begin{tabular}{c} 
\includegraphics[width=15mm,height=15mm]{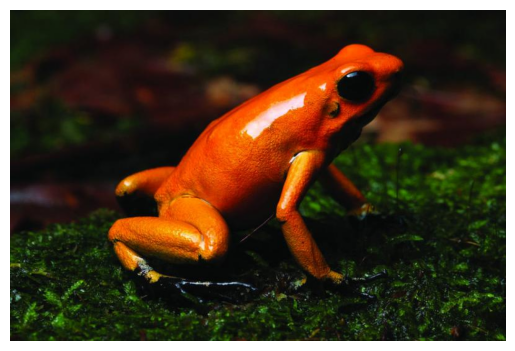}
\end{tabular}&
\begin{tabular}{l} 
What is the primary defense \\
mechanism of this animal?
\end{tabular}\\
\midrule

architecture & 
\begin{tabular}{c} 
\includegraphics[width=15mm,height=15mm]{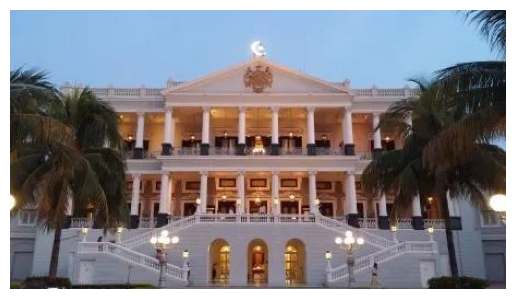}
\end{tabular}&
\begin{tabular}{l} 
Neoclassical buildings are characterized by their \\
magnificence of scale, the prominent use of columns,\\
the use of geometric forms and symmetry, ...Samriddhi \\
Bhavan,...National library of India, Kolkata
\end{tabular}&
\begin{tabular}{c} 
\includegraphics[width=15mm,height=15mm]{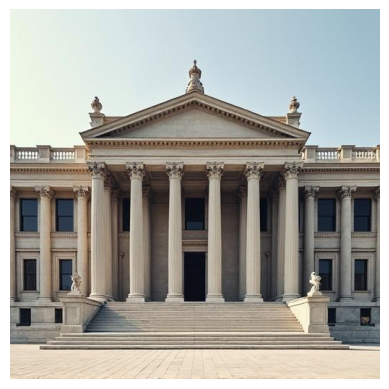}
\end{tabular}&
\begin{tabular}{c} 
\includegraphics[width=15mm,height=15mm]{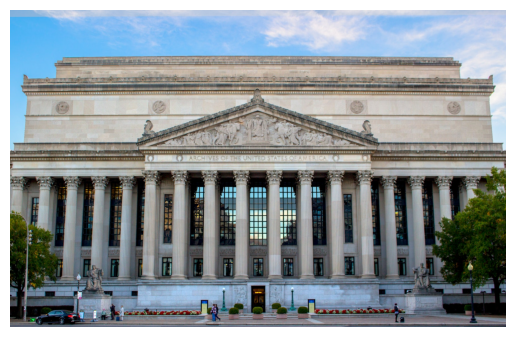}
\end{tabular}&
\begin{tabular}{l} 
What are some examples of this \\
style in Indian public buildings?
\end{tabular}\\
\midrule

artwork & 
\begin{tabular}{c} 
\includegraphics[width=15mm,height=15mm]{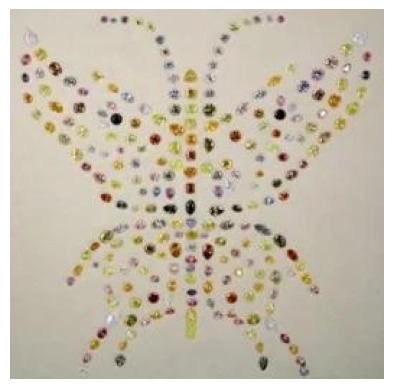}
\end{tabular}&
\begin{tabular}{l} 
"Finding Peace Under Pressure: A Close Look at the \\
new Butterfly of Peace". The Houston Museum of Natural\\
Science. Retrieved 2021-07-05."Aurora Butterfly of Peace\\
on Display at Smithsonian". The Gemmological Association\\
of Great Britain. Retrieved 2021-07-05.
\end{tabular}&
\begin{tabular}{c} 
\includegraphics[width=15mm,height=15mm]{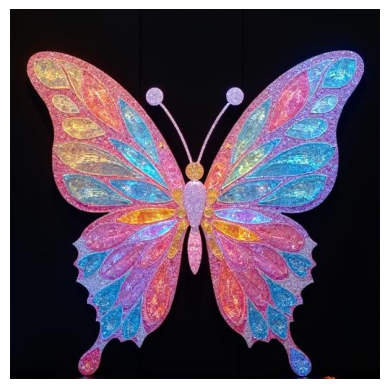}
\end{tabular}&
\begin{tabular}{c} 
\includegraphics[width=15mm,height=15mm]{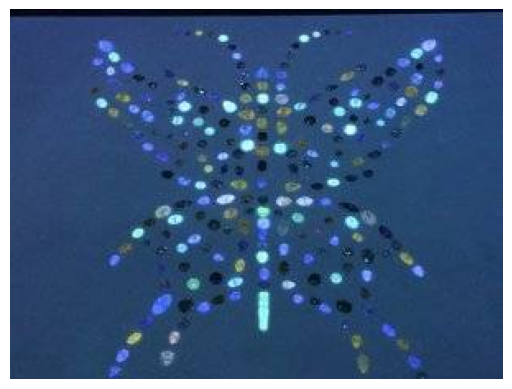}
\end{tabular}&
\begin{tabular}{l} 
Where was this display shown?
\end{tabular}\\
\midrule

currency & 
\begin{tabular}{c} 
\includegraphics[width=15mm,height=15mm]{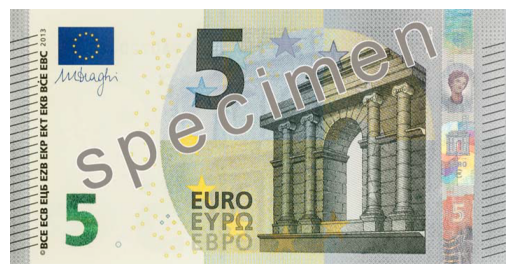}
\end{tabular}&
\begin{tabular}{l} 
The euro was founded on 1 January 1999, when it became \\
the currency of over 300 million people in Europe. \\
For the first three years of its existence it was an... \\
Slovenia joined the Eurozone in 2007, Cyprus and Malta in 2008,\\ Slovakia in 2009, Estonia in 2011 and Latvia on 1 January 2014.
\end{tabular}&
\begin{tabular}{c} 
\includegraphics[width=15mm,height=15mm]{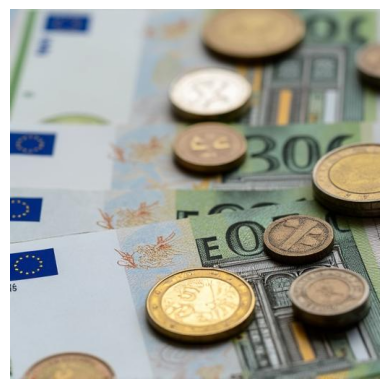}
\end{tabular}&
\begin{tabular}{c} 
\includegraphics[width=15mm,height=15mm]{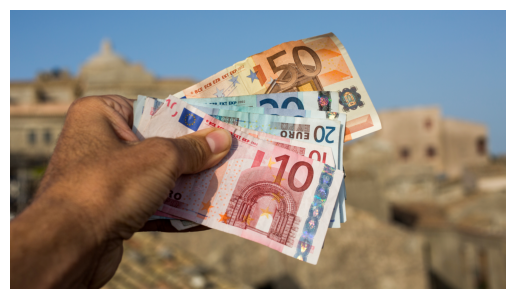}
\end{tabular}&
\begin{tabular}{l} 
When did this currency become \\
available?
\end{tabular}\\
\midrule

entertainment & 
\begin{tabular}{c} 
\includegraphics[width=15mm,height=15mm]{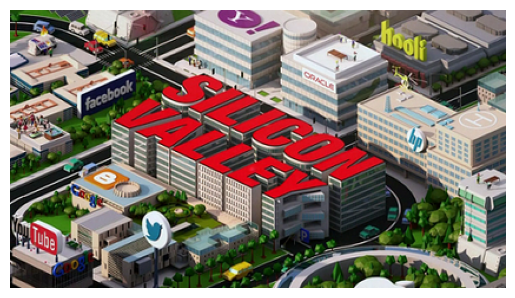}
\end{tabular}&
\begin{tabular}{l} 
Thomas Middleditch as Richard Hendricks, a coder and\\ founder/CEO of Pied Piper.T.J. Miller as Erlich Bachman\\ (seasons 1–4), an Chris Diamantopoulos as Russ Hanneman\\
...a brash, loud and fiery billionaire investor who \\provides Pied Piper with their Series A.
\end{tabular}&
\begin{tabular}{c} 
\includegraphics[width=15mm,height=15mm]{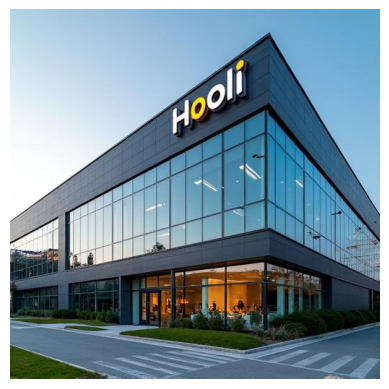}
\end{tabular}&
\begin{tabular}{c} 
\includegraphics[width=15mm,height=15mm]{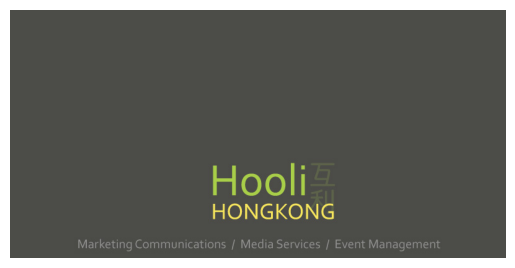}
\end{tabular}&
\begin{tabular}{l} 
Who is the CEO of this company in the\\
TV series Silicon Valley?
\end{tabular}\\
\midrule

food & 
\begin{tabular}{c} 
\includegraphics[width=15mm,height=15mm]{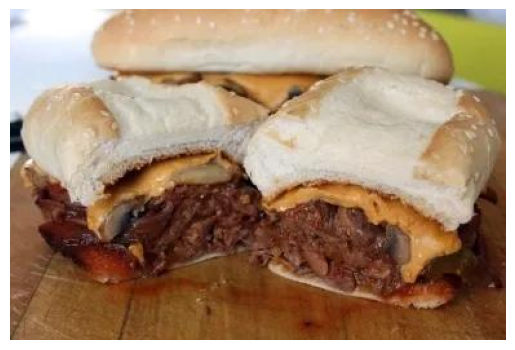}
\end{tabular}&
\begin{tabular}{l} 
An Italian beef sandwich features thin slices of \\
seasoned roast beef, dripping with meat juices, \\
on a dense, long Italian-style roll, believed to \\
have originated in Chicago, where its history ...\\
Despite the name, it is almost completely unknown in Italy.
\end{tabular}&
\begin{tabular}{c} 
\includegraphics[width=15mm,height=15mm]{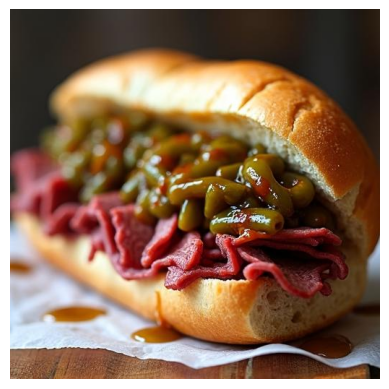}
\end{tabular}&
\begin{tabular}{c} 
\includegraphics[width=15mm,height=15mm]{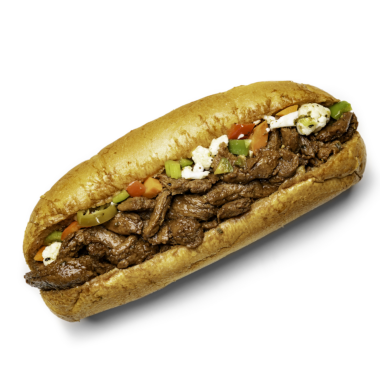}
\end{tabular}&
\begin{tabular}{l} 
What city is this sandwich believed\\
to have originated in?
\end{tabular}\\
\midrule

language & 
\begin{tabular}{c} 
\includegraphics[width=15mm,height=15mm]{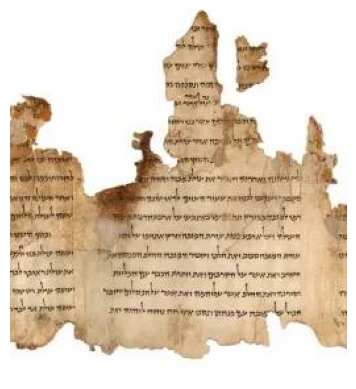}
\end{tabular}&
\begin{tabular}{l} 
In the early 6th century BCE, the Neo-Babylonian \\
Empire conquered the ancient Kingdom of Judah, \\
destroying much of Jerusalem and exiling its \\
population far to the East in Babylon. During \\
...details on Hebrew and Aramaic in the gospels.)
\end{tabular}&
\begin{tabular}{c} 
\includegraphics[width=15mm,height=15mm]{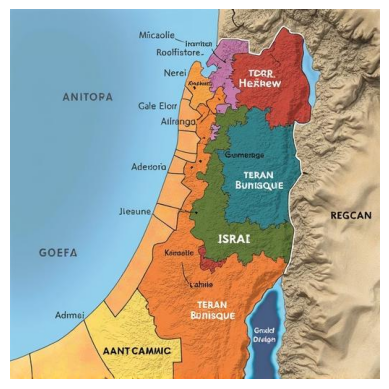}
\end{tabular}&
\begin{tabular}{c} 
\includegraphics[width=15mm,height=15mm]{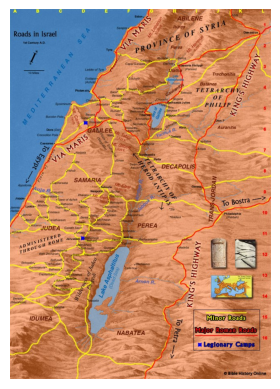}
\end{tabular}&
\begin{tabular}{l} 
What languages were spoken in this \\
region during the Roman period?
\end{tabular}\\
\midrule

literature & 
\begin{tabular}{c} 
\includegraphics[width=15mm,height=15mm]{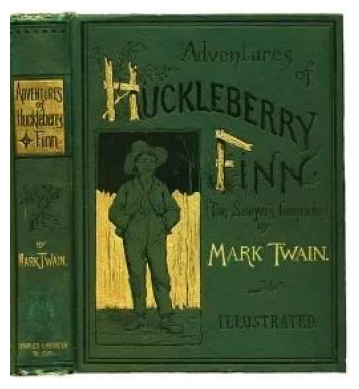}
\end{tabular}&
\begin{tabular}{l} 
The Adventures of Huckleberry Finn (1973), by Robert James\\
Dixson – a simplified version\\
Big River: The Adventures of Huckleberry Finn, a 1985\\
... Classics imprint was released in November 2017.
\end{tabular}&
\begin{tabular}{c} 
\includegraphics[width=15mm,height=15mm]{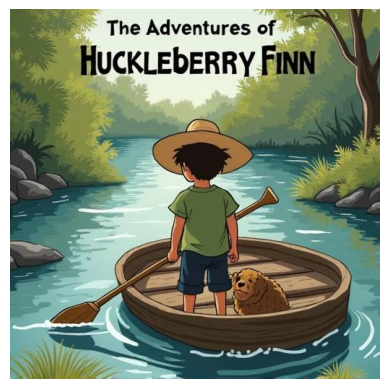}
\end{tabular}&
\begin{tabular}{c} 
\includegraphics[width=15mm,height=15mm]{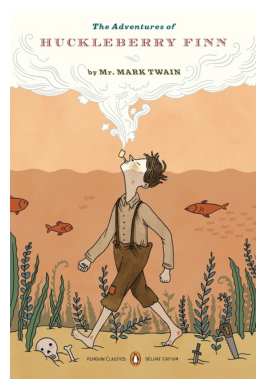}
\end{tabular}&
\begin{tabular}{l} 
What form of media was this book\\
adapted into in 1985?
\end{tabular}\\
\midrule

mythology & 
\begin{tabular}{c} 
\includegraphics[width=15mm,height=15mm]{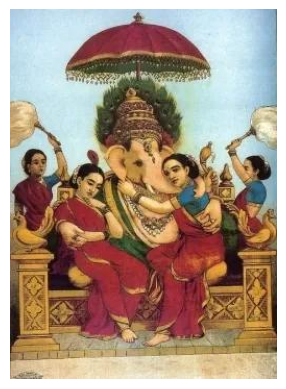}
\end{tabular}&
\begin{tabular}{l} 
Throughout India, on contemporary poster art, \\
Ganesha is portrayed with Sarasvati (goddess of \\
culture and art) or Lakshmi (goddess of luck and \\
prosperity) or both. Ganesha, Lakshmi and Sarswati \\
... to be the brother of Sarasvati and Lakshmi.
\end{tabular}&
\begin{tabular}{c} 
\includegraphics[width=15mm,height=15mm]{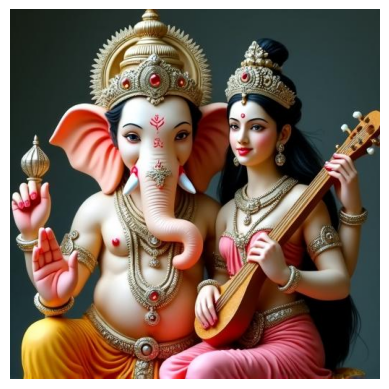}
\end{tabular}&
\begin{tabular}{c} 
\includegraphics[width=15mm,height=15mm]{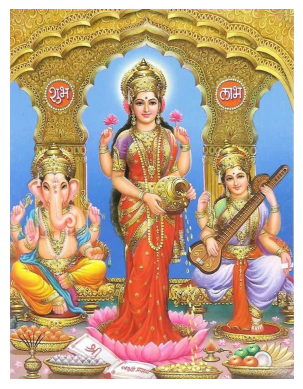}
\end{tabular}&
\begin{tabular}{l} 
What is the relationship between this deity \\
and Sarasvati in Maharashtra?
\end{tabular}\\
\midrule

organization & 
\begin{tabular}{c} 
\includegraphics[width=15mm,height=15mm]{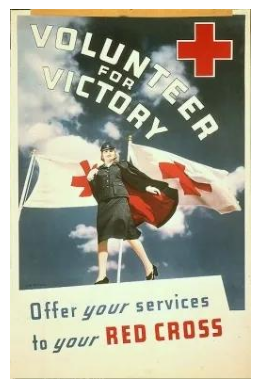}
\end{tabular}&
\begin{tabular}{l} 
During World War II, ARC operated the American Red\\
Cross Clubmobile Service to provide servicemen with\\
food, entertainment and "a connection home." In a \\
...During the Vietnam War 627 American women served\\
in the ARC Supplemental Recreation Overseas Program.\\
At the invitation 
\end{tabular}&
\begin{tabular}{c} 
\includegraphics[width=15mm,height=15mm]{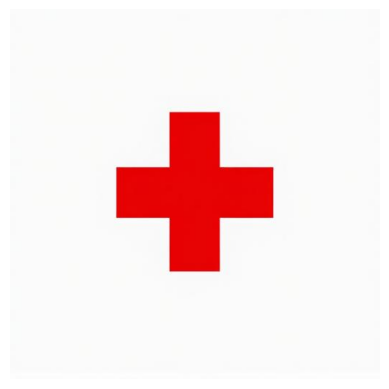}
\end{tabular}&
\begin{tabular}{c} 
\includegraphics[width=15mm,height=15mm]{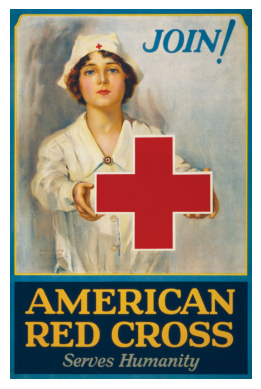}
\end{tabular}&
\begin{tabular}{l} 
What service did this organization provide\\
to boost soldier morale during the Vietnam War?
\end{tabular}\\
\midrule

person & 
\begin{tabular}{c} 
\includegraphics[width=15mm,height=15mm]{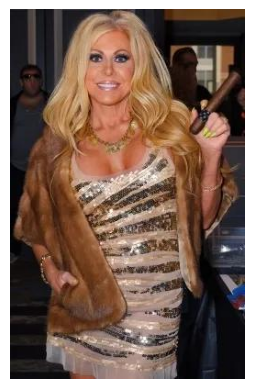}
\end{tabular}&
\begin{tabular}{l} 
Runnels later re-emerged in 1998, under her real name,\\
as the on-screen girlfriend of Val Venis. When Runnels\\
claimed to be pregnant with Venis' baby, he dumped her...\\broke up by July, when Jacqueline Moore \\
became frustrated with Runnels' infatuation with Meat.
\end{tabular}&
\begin{tabular}{c} 
\includegraphics[width=15mm,height=15mm]{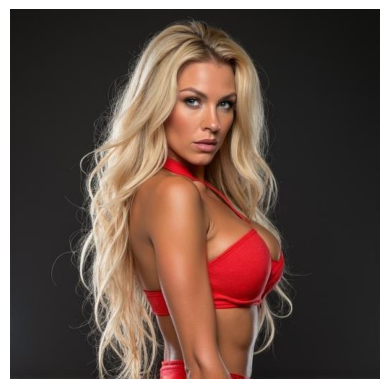}
\end{tabular}&
\begin{tabular}{c} 
\includegraphics[width=15mm,height=15mm]{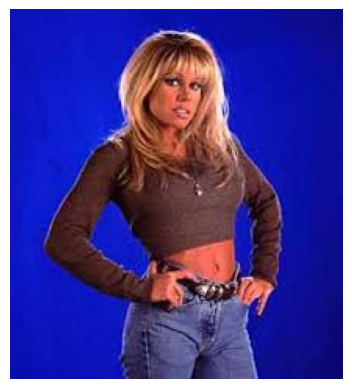}
\end{tabular}&
\begin{tabular}{l} 
Who did this person claim to be \\
\\pregnant with in 1998?
\end{tabular}\\
\midrule

pharmaceutical & 
\begin{tabular}{c} 
\includegraphics[width=15mm,height=15mm]{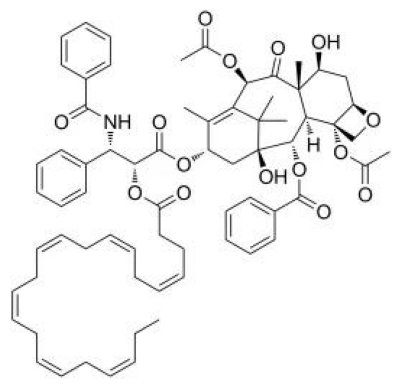}
\end{tabular}&
\begin{tabular}{l} 
DHA-paclitaxel (or Taxoprexin) is an investigational \\
drug (from Protarga Inc) made by linking paclitaxel to\\
docosahexaenoic acid (DHA), a fatty acid that is easily\\
...may be able to treat more types of cancer than Taxol\\
has been able to treat.
\end{tabular}&
\begin{tabular}{c} 
\includegraphics[width=15mm,height=15mm]{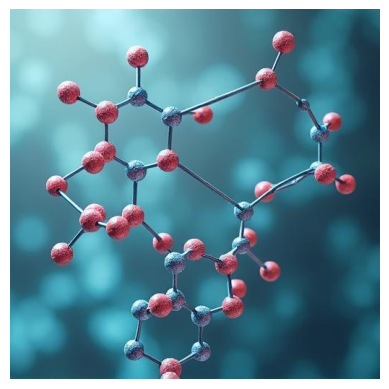}
\end{tabular}&
\begin{tabular}{c} 
\includegraphics[width=15mm,height=15mm]{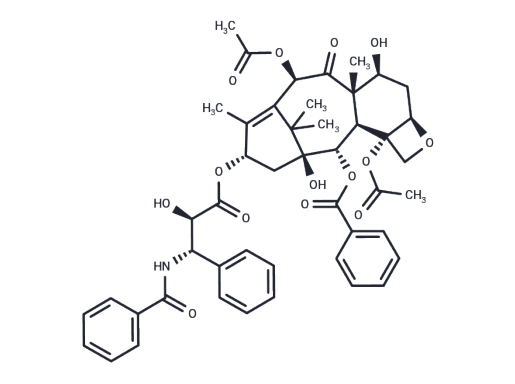}
\end{tabular}&
\begin{tabular}{l} 
What is the advantage of \\
this drug over paclitaxel?
\end{tabular}\\
\midrule

plant & 
\begin{tabular}{c} 
\includegraphics[width=15mm,height=15mm]{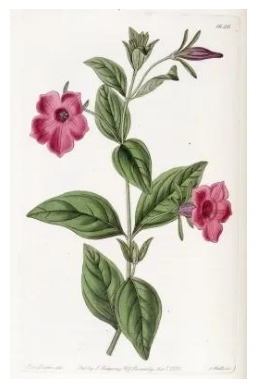}
\end{tabular}&
\begin{tabular}{l} 
The species was first described as Salpiglossis\\ integrifolia by William Jackson Hooker in 1831. \\
It was transferred to the genus Petunia as P. \\
integrifolia by Hans Schinz and Albert Thellung...\\
ranges, with P. inflata growing in more northern areas.
\end{tabular}&
\begin{tabular}{c} 
\includegraphics[width=15mm,height=15mm]{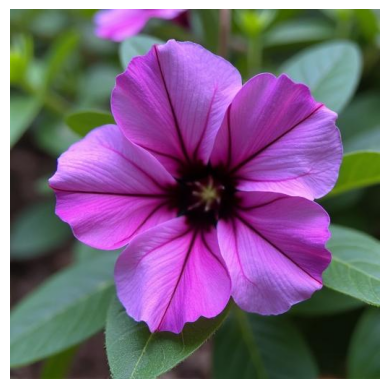}
\end{tabular}&
\begin{tabular}{c} 
\includegraphics[width=15mm,height=15mm]{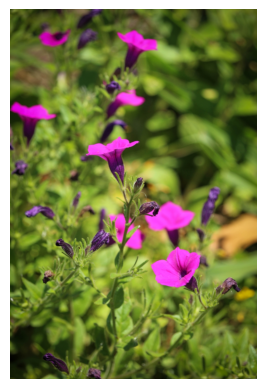}
\end{tabular}&
\begin{tabular}{l} 
What was the original genus of \\
this plant?
\end{tabular}\\
\midrule

sport & 
\begin{tabular}{c} 
\includegraphics[width=15mm,height=15mm]{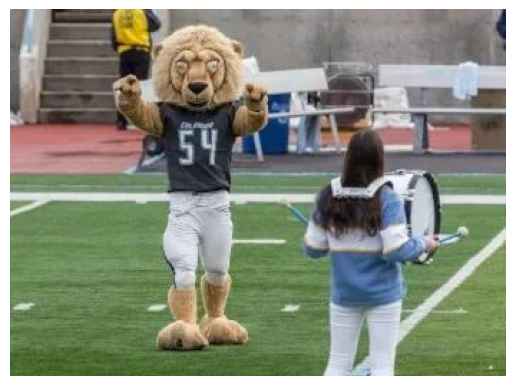}
\end{tabular}&
\begin{tabular}{l} 
The Columbia University Lions are the collective athletic\\ teams and their members from Columbia University, an Ivy\\
League institution in New York City, United States. The \\
current director of athletics is Peter Pilling.
\end{tabular}&
\begin{tabular}{c} 
\includegraphics[width=15mm,height=15mm]{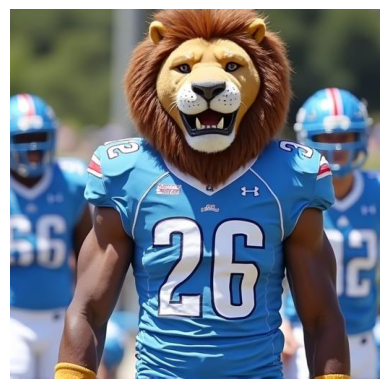}
\end{tabular}&
\begin{tabular}{c} 
\includegraphics[width=15mm,height=15mm]{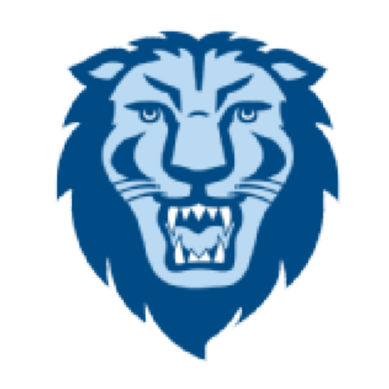}
\end{tabular}&
\begin{tabular}{l} 
What is the name of the athletic \\
teams from this university?
\end{tabular}\\
\midrule

vehicle & 
\begin{tabular}{c} 
\includegraphics[width=15mm,height=15mm]{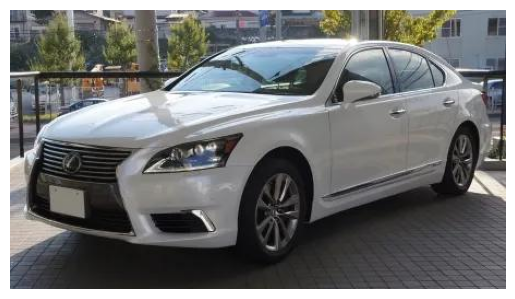}
\end{tabular}&
\begin{tabular}{l} 
A specialized Lexus LS 460 is used in a warehouse-sized\\ driving simulator at Toyota's Higashifuji Technical\\
Center in Shizuoka, Japan. This vehicle is mounted\\
... automotive safety features in a secure environment.
\end{tabular}&
\begin{tabular}{c} 
\includegraphics[width=15mm,height=15mm]{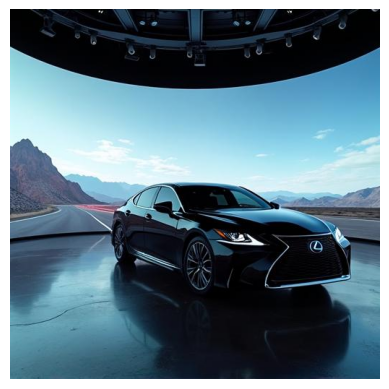}
\end{tabular}&
\begin{tabular}{c} 
\includegraphics[width=15mm,height=15mm]{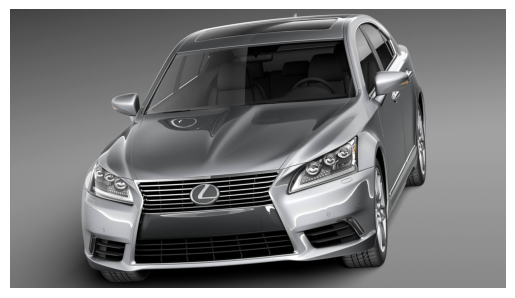}
\end{tabular}&
\begin{tabular}{l} 
What is the purpose of this driving simulator\\
at Toyota's Higashifuji Technical Center?
\end{tabular}\\

\bottomrule
\end{tabular}}
\caption{Examples of synthetic data. Due to the limitations of the table, we have cropped the displayed text.}
\label{tab:synthetic_data_example}
\end{table*}

\section{Limitations}
In this work, we present a benchmark for training and testing Universal Multimodal Retrieval (UMR). To better accomplish this task, we explore strategies for adapting Multimodal Large Language Models (MLLMs) into UMR models, presenting GME, a powerful embedding model capable of retrieving candidates across different modalities. However, this work has its limitations, which are outlined below:

\paragraph{1. Single Image Limit}
In MLLMs, one image is converted into a very large number of visual tokens. In Qwen2-VL, we limit the number of visual tokens to 1024. Due to model training efficiency and a lack of relevant data, our queries and candidates in UMRB only retain a single image. Thus, performance on interleaved data (where multiple images and texts are mixed together) cannot be assessed.

\paragraph{2. Single Language Limit}
Although the backbone of our model, Qwen2-VL, supports multiple languages, we only utilized a single language, English, during the training and testing processes of our GME. Consequently, performance in other languages could not be evaluated.

\end{document}